\documentclass{BaylorThesis_ma} 

\usepackage{lscape}

\usepackage{lipsum}

\usepackage{amsfonts, amssymb, amsmath, latexsym, amsthm}

\usepackage{url}

\usepackage{subcaption}

\usepackage[T1]{fontenc}
\usepackage[utf8]{inputenc}

\usepackage[all]{nowidow}

\usepackage{ragged2e}
\usepackage[apaciteclassic]{apacite} 
\usepackage{float}

\usepackage{graphicx} 

\usepackage{rotating} 

\usepackage{booktabs,threeparttable,threeparttablex}
\usepackage{makecell}

\usepackage{tikz}
\usetikzlibrary{calc,arrows,positioning,shapes,patterns}

\usepackage{longtable} 
\usepackage{array}
\newcolumntype{C}[1]{>{\centering\let\newline\\\arraybackslash\hspace{0pt}}p{#1}}
\newcolumntype{L}[1]{>{\raggedright\let\newline\\\arraybackslash\hspace{0pt}}p{#1}}

\usepackage{float}

\usepackage{graphicx} 

\usepackage{rotating} 

\usepackage{booktabs,threeparttable} 

\usepackage{tikz}
\usetikzlibrary{calc,arrows,positioning,shapes,patterns}

\usepackage{longtable} 
\usepackage{array}
\newcolumntype{C}[1]{>{\centering\let\newline\\\arraybackslash\hspace{0pt}}p{#1}}
\newcolumntype{L}[1]{>{\raggedright\let\newline\\\arraybackslash\hspace{0pt}}p{#1}}

\usepackage{listings}
\sectionnumbering{\secnumfalse} 

\usepackage{lmodern}

\usepackage[inline]{enumitem}

\title{Combatting Human Trafficking in the Cyberspace: A Natural Language Processing-Based Methodology to Analyze the Language in Online Advertisements}
\author{Alejandro Rodriguez Perez}
\degrees{B.S.} 

\mentor{Pablo Rivas, Ph.D.}
\reader{Chen Zhao, Ph.D.}
\readerThree{Liang Dong, Ph.D.}
\readerFour{D. Smith, Ph.D.} 
\readerFive{E. Smith, Ph.D.}

\confDate{October 2023} 

\makeCopyrightPage

\graduateDean{J. Larry Lyon, Ph.D.}
\deptChair{Pablo Rivas, Ph.D.}

\abstract{
This project tackles the pressing issue of human trafficking in online C2C marketplaces through advanced Natural Language Processing (NLP) techniques. We introduce a novel methodology for generating pseudo-labeled datasets with minimal supervision, serving as a rich resource for training state-of-the-art NLP models. Focusing on tasks like Human Trafficking Risk Prediction (HTRP) and Organized Activity Detection (OAD), we employ cutting-edge Transformer models for analysis. A key contribution is the implementation of an interpretability framework using Integrated Gradients, providing explainable insights crucial for law enforcement. This work not only fills a critical gap in the literature but also offers a scalable, machine learning-driven approach to combat human exploitation online. It serves as a foundation for future research and practical applications, emphasizing the role of machine learning in addressing complex social issues.
}

\abbreviation{

\singlespacing
\setlength{\extrarowheight}{0.5\baselineskip}
\begin{tabular}{L{.2\textwidth}L{.75\textwidth}}
HT & Human Trafficking \\
NLP & Natural Language Processing \\
OEA & Online Escort Advertisement \\
SLR & 	Systematic Literature Review \\
CNN & Convolution Neural Network \\
RNN & Recurrent Neural Network \\
IE & Information Extraction \\
HTRP & Human Trafficking Risk Prediction \\
OAD & Organized Activity Detection \\
HTIA &  Human Trafficking Indicators Analysis \\
ML & Machine Learning \\
PCA & Principal Components Analysis \\
LDA & Latent Dirichlet Allocation \\
TF-IDF & Term Frequency-Inverse Document Frequency \\
NER & Named Entity Recognition \\
BERT & Bidirectional Encoder Representations from
Transformers \\
GPT & Generative Pre-trained Transformers\\
ALBERT & A Lite BERT\\
RoBERTa & Robustly optimized BERT approach\\
XLNet & eXtreme Multi-Label Net\\
MLM & Masked Language Modeling \\
NSP & Next Sentence Prediction \\
IG & Integrated Gradients \\
DF & Deliver Fund \\

\end{tabular}
}

\acknowledgements{
First and foremost, I would like to express my deepest gratitude to Dr. Pablo Rivas at Baylor University. As my advisor for this project, Dr. Rivas has guided me through the complexities of my research. His leadership in the NSF grant that inspired this work has been invaluable.

I am also immensely grateful to Dr. Gisela Bichler at California State University San Bernardino, Dr. Laurie Giddens at the University of North Texas, and Dr. Stacie Petter at Wake Forest University. Their mentorship, guidance, and provision of data, references, and valuable insights as subject matter experts have been crucial. Their collaboration in interpreting some of the results has enriched this work immeasurably.

Special thanks go to Mr. Michael Fullilove at DeliverFund for partnering with our team of investigators and providing part of the data we needed for training our models. Your contribution has been pivotal to the success of this project.

I want to extend my appreciation to Dr. Tomas Cerny at the University of Arizona for his guidance as a software engineering expert and researcher. Your expertise has been a great asset to this research.

I am also indebted to Dr. Javier Turek at Intel Labs for his valuable insights as a machine learning subject matter expert, particularly in the field of NLP. Our fruitful discussions and feedback have been incredibly enlightening.
    
I must also acknowledge the hard work and dedication of PhD students Ernesto Quevedo and Korn Sooksatra. Their commitment to helping with experimental tasks and engaging in research discussions, as well as running additional experiments for verification, has been commendable.

Last but not least, I would like to thank Maisha Rashid Binte, Jorge Yero, and Ana Paula Arguelles for their contributions to this research. Their additional validation on specific hypothesis tests and contributions to research review and analysis have been invaluable.

This work is supported by the National Science Foundation under Grant 2210091. The views expressed herein are solely those of the author and do not necessarily reflect those of the National Science Foundation.

}

\newcounter{rowno}

\begin{document}
\pagenumbering{arabic}

\setlength{\abovedisplayskip}{3pt}
\setlength{\belowdisplayskip}{3pt}

\chapter{Introduction}

\section{Consumer-to-consumer Marketplaces}

Consumer-to-consumer (C2C) marketplaces are online platforms where individuals can buy and sell goods or services directly to and from each other. These marketplaces have gained significant popularity in recent years due to the ease of online transactions and the potential for finding unique items or services~\shortcite{dan2014consumer}. C2C marketplaces offer advantages such as direct interaction between buyers and sellers, often competitive prices, and the potential to find unique or specialized items~\shortcite{wei2019trust}. However, experts advise users to exercise caution when making transactions on these platforms and be aware of safety and fraud prevention measures to ensure a secure and positive experience~\shortcite{dan2014consumer}.

Despite their obvious advantages, some platforms are prone to illicit activity, like the selling of stolen car parts and human trafficking~\shortcite{giddens2023navigating}. The benefits that legitimate consumers and sellers enjoy by using these platforms are also appreciated by cyber-criminals who wish to sell their illegally acquired goods or offer unlawful services~\shortcite{latonero2011human,aniello2018selling}. 

\section{The Problem of Human Trafficking}

The United Nations Office on Drugs and Crime~(UNODC) defines human trafficking as \textit{(\dots) the recruitment, transportation, transfer, harbouring or receipt of people through force, fraud or deception, with the aim of exploiting them for profit. Men, women and children of all ages and from all backgrounds can become victims of this crime, which occurs in every region of the world. The traffickers often use violence or fraudulent employment agencies and fake promises of education and job opportunities to trick and coerce their victims}~\shortcite{unodc2023human}.

The number of victims of human trafficking is estimated in the order of millions worldwide~\shortcite{latonero2011human}, and it generates multi-billion profits for the perpetrators~\shortcite{zhu2019detecting}. The most recent Global Report on Trafficking in Persons~(GLOBTiP) reveals that the majority~(74\%) of the reported trafficked victims are due to sexual exploitation~\shortcite{unodc2022glotip}, making sex trafficking the main modern form of slavery. These figures come around despite the multiple international instruments that exist to fight human trafficking and the adherence of 87\% of UN members to all of them~\shortcite{unodc2022glotip}.

Prior to the widespread adoption of the Internet, pimps used street advertising to promote their victims. This practice offered law enforcement a tangible location to confront the issue and locate victims. Conversely, the virtual environment offers several advantages to traffickers advertising their illegal enterprise online. It allows them to expand their reach and client base significantly while remaining anonymous. Furthermore, it is easier for them to move victims across states to avoid detection and still be able to conduct business securely~\shortcite{kennedy2012predictive}.

The GLOBTiP also shared findings like a reduction in 11\% of the number of detected victims of human trafficking in the world and that trafficking for sexual exploitation was 24\% less 
detected during the pandemic~\shortcite{unodc2022glotip}. Despite being seemingly favorable, these metrics are not necessarily inviting to optimism. A potential explanation is that human trafficking, like many other activities, has, at least partially, moved from physical to online environments successfully, where the issue is harder to trace. All the more reasons to thoroughly investigate criminal activity occurring in the cyber-space.

Fighting sex trafficking is, unfortunately, a time-consuming and expensive process, and in recent decades, the challenge has escalated with the transition to online classified advertisements~\shortcite{zhu2019detecting}. Reportedly, around 150,000 escort advertisements flood the internet daily, and law enforcement agencies do not have the resources to process them exhaustively~\shortcite{giddens2023navigating}. In this context, the research community, through individual contributions as well as government-funded programs in the U.S., has contributed with methods and tools that allow semi-automated processing of the high volume of advertisements with successful results~\shortcite{kejriwal2017flagit,vajiac2022trafficvis,vajiac2023deltashield}.

\section{Natural Language Processing}

Natural Language Processing~(NLP) is a subfield of artificial intelligence~(AI) that focuses on the interaction between computers and human languages. Its primary aim is to enable computers to understand, interpret, and generate human language in a valuable and meaningful way. Beginning with early rule-based systems and statistical models, NLP has evolved significantly with the advancements in machine learning and deep learning techniques. Traditional approaches, relying on rule-based methods and statistical models, have gradually given way to more sophisticated algorithms, enabling tasks such as sentiment analysis, language translation, and speech recognition. Recent developments in NLP have been particularly marked by the emergence of large language models, such as GPT-3~\shortcite{brown2020language}, which leverage deep learning to comprehend and generate human-like text, revolutionizing various applications in text generation, summarization, and even conversational AI. These models represent a significant breakthrough in NLP, significantly enhancing the capabilities of machines to understand and generate human language.

\begin{figure}
    \centering
    \includegraphics[width=\textwidth]{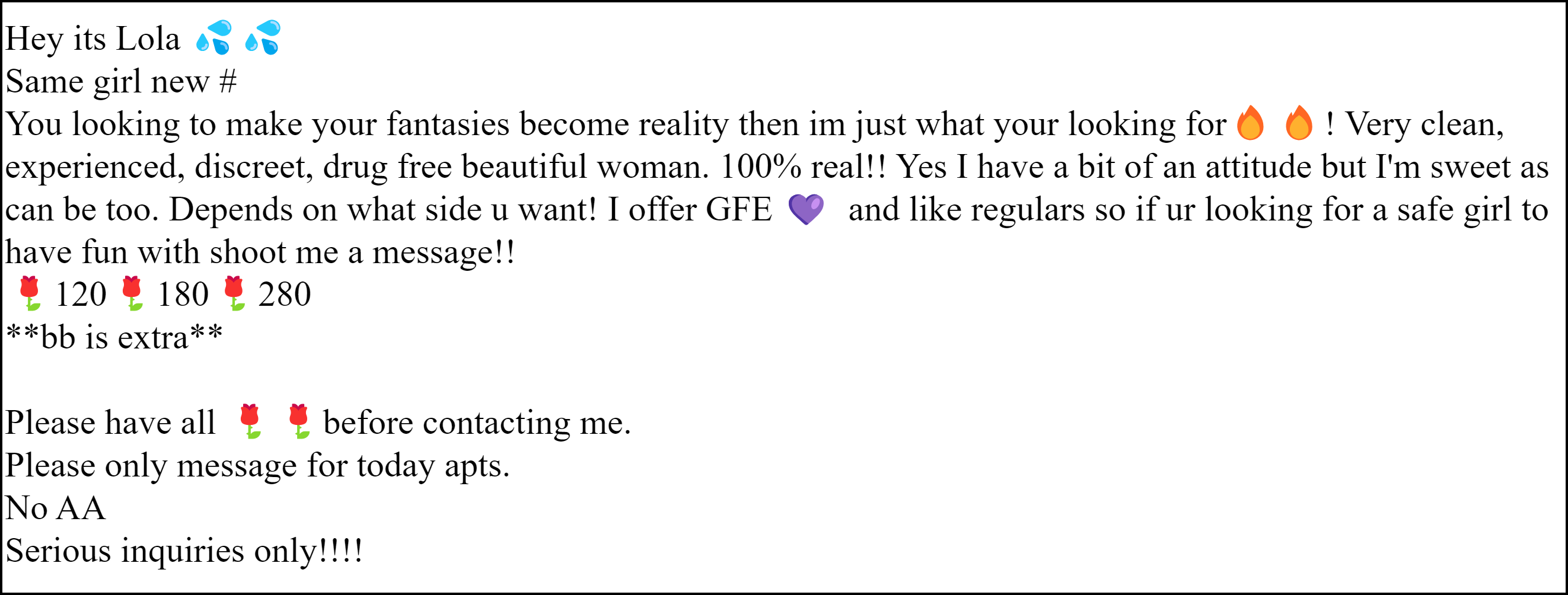}
    \caption{Example of an online escort advertisement.}
    \label{fig:post_example}
\end{figure}

Since a significant share of the information found in an escort advertisement comes from text fields like title and description, NLP naturally plays an important role in efforts to automate the inspection and analysis of information found in online escort advertisements~(OEAs). Nevertheless, dealing with OEAs in the context of human trafficking faces significant challenges. First, the presence of sex trafficking is relatively scarce and hidden among other legal activities. A study suggests that the volume of sex trafficking cases amounts to only 5.5\% of the total number of advertisements posted~\shortcite{dubrawski2015leveraging}. Second, the language in the advertisements is very noisy. It contains misspellings, jargon, emojis, other non-ASCII characters, and grammar mistakes. Figure \ref{fig:post_example} illustrates this phenomenon. In the figure, the name appearing is made up. Observe the presence of emojis, misspellings, and acronyms.

The noisy language is not the only problem. Additionally, the use of the language is adversarial since authors intentionally try to obfuscate and hide their real intentions to avoid being caught~\shortcite{zhu2019detecting}. Finally, curating and labeling a dataset for use by the research community is an expensive process that would involve law enforcement and academic experts.

\section{Research Objective}

The reasons just presented motivate us to define the following research objective. We aim to:

\textbf{Developing a methodology that allows us to interpret the language in advertisements from C2C marketplaces in order to find new potential indicators of human trafficking activity or validate existing ones.
}

In order to achieve that goal, our work was divided into several broad tasks, outlined below.

\begin{enumerate}
    \item Study the existing literature that addresses the problem of human trafficking from a computational perspective, specifically, methods working with escort websites.
    \item Based upon the findings in the literature, define an extensible and robust methodology to develop pseudo-labeled datasets with minimal human supervision in tasks related to human trafficking.
    \item Train and evaluate state-of-the-art NLP models on the datasets defined.
    \item Interpret models' predictions using a suitable, well-defined, and quantifiable method such that explainable insights can be extracted.
\end{enumerate}

The remainder of the thesis explains our research and the steps to address each task. Particularly, it is organized as follows. In Chapter \ref{chpt:slr}, we describe a Systematic Literature Review~(SLR) we conducted on relevant literature. The methods utilized for the study are explained, and the results are presented and interpreted. An important gap that is described is the lack of curated, publicly available data. Consequently, the findings of that study were extrapolated to Chapter \ref{chpt:data}, where we elaborate on a methodology we designed to obtain and process data from online escort websites. The processed data is converted into a pseudo-labeled dataset with multiple tasks related to human trafficking, namely, Human Trafficking Risk Prediction~(HTRP) and Organized Activity Detection~(OAD). Next, in Chapter \ref{chpt:attr}, we describe how we trained and evaluated state-of-the-art language models to learn the HTRP and OAD tasks. Furthermore, we inspect the fit models' predictions by means of an axiomatic gradient-based interpretability method for differentiable functions called Integrated Gradients. The results of that analysis are presented and discussed thoroughly. Finally, Chapter \ref{chpt:conclusion} draws the conclusions of our work, reviews our contributions, and provides further research alternatives.

\section{Summary of Contributions}

We identified the following main contributions of our research:

\begin{enumerate}
    \item A comprehensive SLR of the literature addressing the problem of human trafficking from a computational perspective.
    \item A flexible and robust methodology to develop datasets related to human trafficking activity based on heuristics, and thus, with minimal supervision.
    \item Part of the dataset generation method requires an information extraction system. Our contribution in this area is an evaluation of state-of-the-art Transformer models to solve the Named Entity Recognition problem in the adversarial context of online escort ads text.
    \item A methodology to analyze the predictions of a trained neural network with the purpose of understanding and finding new insights in the language utilized by offenders.
\end{enumerate}

%
\chapter{Methods for Systematic Analysis of Online Escort Advertisements to Combat Human Trafficking: a Rapid Literature Review}
\label{chpt:slr}

\section{Introduction}

This project's scope concerns human trafficking activity in online platforms, particularly in consumer-to-consumer~(C2C) websites. The research community has devoted significant efforts to studying this and other closely related topics. A survey of the existing literature would allow for identifying existing methods and discerning gaps and areas to contribute.

This chapter presents a study of the literature related to human trafficking detection in online escort advertisements~(OEAs). Our study follows the principles of a Systematic Literature Review~(SLR) as it was carefully planned, research goals and questions were defined, relevant literature was identified and inspected for relevancy in a reproducible manner, the data extraction process was well-specified, and the results were synthesized and utilized to answer the research questions. The limitations in the search scope make this study fall under the category of a Rapid Review.

The remainder of the chapter is organized as follows. Immediately following this introduction, a section discusses related work that we encountered. Next, we describe the methods we used for our review, which include the research questions we posed, the inclusion and exclusion criteria used to select relevant literature, the search process we followed, the data extraction form we applied to each paper, and a description of how the method we used to synthesize the data we collected for each paper. After that, we present the results of our study organized by research questions. Finally, the conclusion notes for the chapter are presented.

\section{Related Work}

We identified papers that review the existing literature and report the state of the art, computational methods, open venues, etc., in the field of human trafficking detection. Those studies are not subject to our systematic review. However, this section summarizes their methods and contributions and points out how ours fills some existing gaps.


In their work, \shortciteA{razi2021human} sought to identify computational approaches for detecting online sexual risks. The study encompassed various types of risks, including sexual violence, harassment, grooming, and sex trafficking. It emphasized the importance of realistic datasets and high-quality annotations that consider the perspectives of those affected by the risks. The articles were qualitatively coded based on factors such as data validity, algorithmic approaches, model quality metrics, and real-world deployment.

\shortciteA{dimas2022operations} investigated and categorized the available literature on Operations Research and Analytics oriented to fight human trafficking. They also suggested future research directions in this area, addressing a knowledge gap in the existing literature. The study concentrated on the four fundamental anti-human-trafficking principles: prevention, protection, prosecution, and partnership.

\shortciteA{xian2022human} adopted a systematic approach to reviewing existing literature with the goal of exploring how analytics techniques had been used to address the problem of human trafficking. The paper defines a taxonomy of methods divided into two major groups: Machine Learning Analysis and Statistical Analysis, and groups reviewed documents accordingly. Their findings include a prevalent use of classification algorithms and data sampling methods in the existing literature and the absence of methods addressing the aspect of forced labor in human trafficking.

Table \ref{tab:surveys} summarizes information related to the literature review studies mentioned before, whereas Table \ref{tab:rqs} lists the research questions addressed in each article. From \ref{tab:surveys}, we observe that all three are relatively recent works, literature from before 2007 is not considered, and IEEE Xplore, Springer Link, and ACM Digital Library are indexers preferred in more than one study. Interestingly, in table \ref{tab:rqs}, note that  \citeA{xian2022human} did not formally state their research questions.

\begin{table}[]
    \centering
    \caption{Related literature review studies.}
    \begin{tabular}{|p{.2\textwidth}|p{0.2\textwidth}|p{.3\textwidth}|p{.15\textwidth}|}
        \hline
         \textbf{Study} & \textbf{Publications Years} & \textbf{Indexers} & \textbf{Primary Studies} \\
         \hline
         \citeA{razi2021human} & 2007 - 2020 & 
         \begin{itemize}[leftmargin=*,nolistsep]
             \item IEEE Xplore
             \item ACM DL
             \item Science Direct
             \item Springer Link
             \item ACL Anthology
         \end{itemize} & 73 \\
         \hline
         \citeA{dimas2022operations} & 2010 - 2021 & \begin{itemize}[leftmargin=*,nolistsep]
             \item Scopus
             \item Web of Science
             \item Google Scholar
         \end{itemize} & 142\\
         \hline
         \citeA{xian2022human} & 2011 - 2020 & 
         \begin{itemize}[leftmargin=*,nolistsep]
             \item ArXiv
             \item Science Direct
             \item ACM DL
             \item IEEE Xplore
             \item Springer Link
         \end{itemize} & 32\\
         \hline
    \end{tabular}
    
    \label{tab:surveys}
\end{table}

\begin{table}[]
    \centering
    \caption{Research questions of related literature review studies.}
    \begin{tabular}{|p{.2\textwidth}|p{.75\textwidth}|}
        \hline
         \textbf{Study} & \textbf{Research Questions} \\
         \hline
         \citeA{razi2021human}& 
            \begin{enumerate}[leftmargin=*,nolistsep]
                \item Are the datasets ecologically valid for detecting the targeted risk for the desired user population?
                \item Are the algorithmic models grounded in human theory, understanding, and knowledge?
                \item How well do the algorithms perform, both computationally and in meeting end users’ needs?
                \item What system artifacts were developed, and what were the outcomes when deployed in real-world settings?
            \end{enumerate}\\

            \hline
        \citeA{dimas2022operations} & 
            \begin{enumerate}[leftmargin=*,nolistsep]
                \item What aspects of human trafficking are being studied by OR and Analytics researchers?
                \item What OR and Analytics methods are being applied in the anti-HT domain?
                \item What are the existing research gaps associated with 1 and 2?
            \end{enumerate}
             \\
             \hline
         \citeA{xian2022human}& Not specified \\
         \hline
    \end{tabular}
    
    \label{tab:rqs}
\end{table}

\section{Methods}

The purpose of this review is to \textbf{appraise and synthesize research evidence of how OEAs are investigated, mined, and analyzed by means of computational methods to detect and combat human trafficking}. To attain that goal, we devised a method that fit our personnel, budget, and time constraints but also fell close to a strict and well-defined methodology for an SLR.

\subsection{Research Questions}

As for any SLR, research questions define the essence of the process and guide the scope and results that yield from it. As a result of previous screening of related work and endorsed by Information Systems and Criminology experts, we elaborated the following four research questions to orient our study:

\begin{enumerate}
    \item What problems and systematic methods to solve them exist that contribute to fighting human trafficking activity operated through OEAs?
    \item Which indicators of human trafficking found in OEAs text are used to predict human trafficking?
    \item  What feature of escort advertisements can be leveraged to build connections between different ads and identify a network of providers?
    \item What are the characteristics of the escort advertisement data used in the relevant literature?
\end{enumerate}

The relevance and scope behind each research question will become completely clear in the \textit{Results} Section.

\subsection{Inclusion and Exclusion Criteria}

The inclusion and exclusion criteria are important to be set in order to filter out irrelevant documents and concentrate the study solely on the essential contributions. We defined the following set of inclusion and exclusion criteria.

We aimed at \textit{including}:

\begin{enumerate}
    \item documents that use systematic computational methods for the collection and/or analysis of data found in OEA websites with applications in human trafficking risk prediction or organized activity detection; and
    \item papers that, despite not providing systematic methods for automated processing of ads, describe and highlight the importance of human trafficking indicators found in OEAs.
\end{enumerate}

We decided to \textit{exclude} the following:

\begin{enumerate}
    \item thesis, when peer-reviewed publications do not cite these or whenever their contributions have been made available through subsequent peer-reviewed publications;
    \item duplicate records;
    \item papers that are not peer-reviewed;
    \item papers not written in English;
    \item papers not using systematic computational methods;
    \item papers where the methods are not evaluated explicitly in the scope of human trafficking, even when those may apply to the human trafficking domain; and
    \item studies that do not focus on OEAs as their main source of information, e.g., social media and review websites)
\end{enumerate}

\subsection{Search Process}

Our study did not start from scratch. Prior work by collaborators in our project included an SLR on Natural Language Processing~(NLP) methods for detecting human trafficking in OEAs. That study yielded 21 papers identified as relevant. Before the last filter was applied, there were another 29 identified as irrelevant for that study; however, some were relevant for our study. Appendix \ref{ap:seed_papers} provides the detailed methods employed in the prior study of our collaborators.

Given the close relationship between the two studies, we decided to utilize the outcome of the former as the seed to incorporate more studies. Our initial list of 30 papers is the result of applying our inclusion and exclusion criteria to the list of 50 papers mentioned above. We extended that list by means of snowballing.

Snowballing is an artifact in literature review work aimed at completeness. It is the iterative process of identifying additional relevant sources by examining the citations and references in previously found literature. Given our constraints and the characteristics of the field, snowballing seemed a more suitable way to extend the existing study to the updated purposes than it was to re-design the search protocol and conduct a time-consuming evaluation phase over an extensive collection of papers.

We used both forward snowballing and backward snowballing. Each reference in the written document is inspected for backward snowballing, whereas forward snowballing was conducted using the Google Scholar citations tool. The latter constitutes a limitation of this study since the scope of papers observed this way is constrained by the indexing capability of Google. The title, abstract, and sometimes the full text are processed for each new entry, and the inclusion and exclusion criteria explained before are applied to evaluate its addition to the study. The snowballing process was conducted between May and June of 2023.

As a result, 29 new relevant works were identified, including those we regarded as related works: 11 from forward snowballing, 9 from backward snowballing, and 9 could have resulted from either of those, depending on the order in which the initial list was processed. After a full read, the list of 59 papers was left in 43. Out of those, 3 are regarded as relevant related work, and the other 40 were used as the final list of papers and those in which the remaining of the study was conducted.

For completeness, Table \ref{tab:fulllist} provides the final list of primary studies utilized in the remainder of the protocol execution.

\subsection{Data Extraction}

The coding scheme of the study is defined to capture the essential information relevant to answering the research questions.

For RQ1, we capture the definition of the problem(s) at hand and extract a summary of the method(s) used. For its relevance in this and other areas, in case there is a mention of Machine Learning~(ML) as part of the methods, we specifically code it, along with a description of features utilized by the ML approach. Finally, we collect the reported results of the proposal.

We base our analysis of RQ2 on the concept of human trafficking indicators. Our coding is thus comprised of definitions, ways in which those are (or could be) extracted, as well as flagging each paper as referencing using an indicator or not.

The codification of RQ3 is similar to RQ2, but instead of extracting indicators of human trafficking, features and information that allow building relations between advertisements are identified.

Finally, for RQ4, we considered extracting the source website(s) of the data, the dates covered by the ads, their location, the information collected from the metadata, the number of ads, and whether some human-supervised labeling had been given to the data.

Table \ref{tab:coding_fields} presents a summary of the data extraction fields for each research question.

\begin{table}[H]
    \centering
    \caption{Data extraction fields corresponding to each research question.}
    \begin{tabular}{|p{0.25\textwidth}|p{0.65\textwidth}|}
        \hline
        \textbf{Research Question} & \textbf{Coding Scheme} \\
        \hline
         RQ1 & Problem, Methods, Algorithms, ML Features, Results \\
         RQ2 & Binary coding of indicators, Definitions \\
         RQ3 & Binary coding of connectors, Definitions \\
         RQ4 & Source website(s), Date range, Location(s), Publicly available,	Metadata, Size\\
         \hline
    \end{tabular}
    
    \label{tab:coding_fields}
\end{table}

The results were recorded in a spreadsheet. To facilitate further analyses and synthesis, the data was normalized and consolidated into a completely structured document in JSON format.\footnote{Both the original spreadsheet and the consolidated JSON file are available at \url{https://github.com/Baylor-AI/SaTC}.}

\subsection{Data Synthesis}

We analyze the information collected in our study, both qualitatively and quantitatively. The results are synthesized in narrative, tabular, and graphical formats, as discussed next.

For RQ1, we describe our taxonomy of relevant problems existing in the literature reviewed and how they interconnect to contribute to the general goal of detecting and combating human trafficking. Additionally, we briefly narrate the methods employed for each problem. We also count the number of works belonging to several different categories defined based on the problem solved and the methods used in order to observe the prevalence of certain problems or methods. Finally, we attempt to consolidate the state-of-the-art in each problem, acknowledging the limitations derived from the lack of well-established public benchmarks.

In the case of RQ2, we focus our attention on identifying what expert knowledge can be directly applied to systematically aid the detection of human trafficking and organized activity in online platforms. We capture that knowledge in the form of a taxonomy of indicators. Each indicator defines a concrete piece of interpretable knowledge from law enforcement that can help detect human trafficking or organized activity contributing to it. We synthesize the application of such indicators and the way they are extracted in the literature by counting the number of papers that apply each.

Finally, in RQ3, we quantify how much data has reportedly been collected and what information is available. We showcase, by means of count histograms, the relative attention that has been paid to different websites and locations. Finally, we sketch an analysis of the availability of the data that has been collected and the importance of common benchmarks.


\section{Results}\label{sec:slr_results}

This section presents the results of our study. First, general analyses are presented. Then, we address each research question separately.

\subsection{RQ1: What problems and systematic methods to solve them exist that contribute to fighting human trafficking activity in online escort advertisements~(OEAs)?}\label{subsec:rq1}

We start our discussion by describing a taxonomy of problems we found in the literature relevant to detecting human trafficking in OEAs. Throughout this reading, we use consistent terminology to refer to each problem, but readers can find them named differently in the literature. When adequate, we mention different aliases employed for the same problem.

After a careful revision of the problems posed by the several documents inspected, we decided to categorize recurrent similar problems into three main categories: (a) Human Trafficking Risk Prediction~(HTRP); (b) Organized Activity Detection~(OAD); and (c) Human Trafficking Indicators Analysis~(HTIA).

The main focus of our work is the problem of HTRP. This problem aims to identify instances of human trafficking derived from the analysis of OEAs. Throughout the literature inspected, it adopts several forms, depending on the number of classes considered (binary vs. multi-class) and whether the classification targets an individual advertisement or a group of those.

Secondly, the OAD category is targeted toward finding structure and traces of organized activity based on the information found in OEAs. Most commonly, the goal is to find clusters of related ads and then treat them as a single organization to conduct further analyses, including detecting whether that organization is likely performing human trafficking activities.

Finally, the HTIA category encompasses documents that study the presence of indicators of human trafficking in OEAs, despite not doing so by means of automated computational methods. We deemed these to be relevant for two main reasons. First, in most cases, their analyses can be extrapolated and automated. Secondly, they allow us to enrich a taxonomy of indicators of human trafficking that can potentially be found in OEAs.

\begin{figure}
    \centering
    \includegraphics[width=\textwidth]{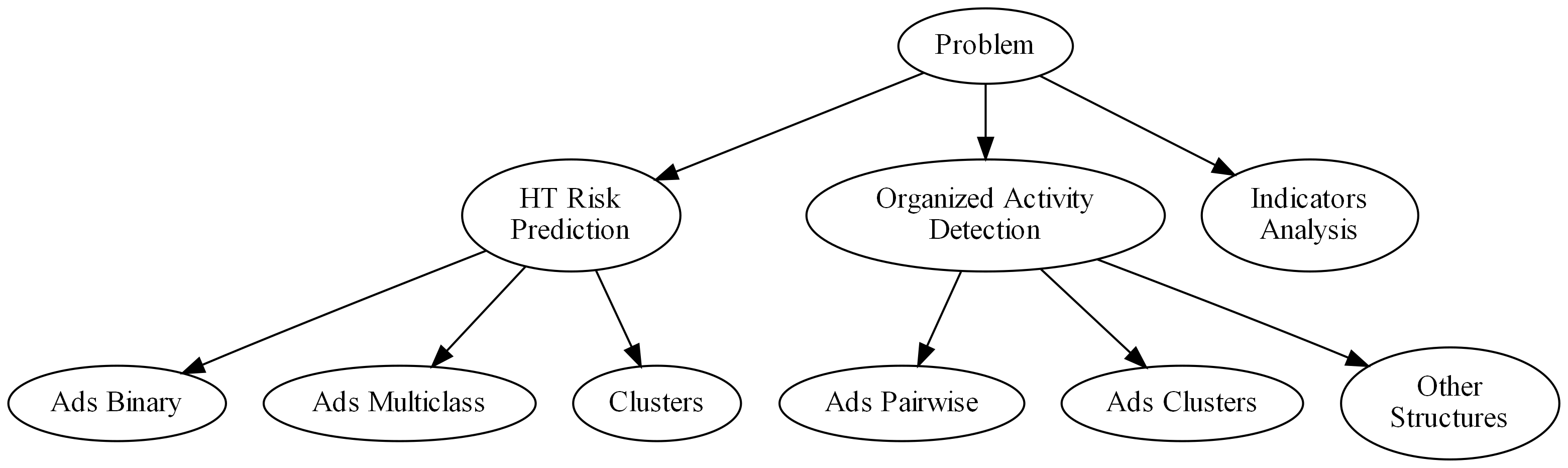}
    \caption{Taxonomy of problems the study focuses on.}
    \label{fig:problems_tree}
\end{figure}

Figure \ref{fig:problems_tree} illustrates the taxonomy of problems analyzed and their sub-problems. Regarding the sub-problems of HTRP, Ads Binary is the variant where the classification targets are binary labels, e.g., HT versus non-HT; Ads Multiclass is when there are more than two classes, e.g., several degrees of human trafficking risk; and the Clusters category encompasses setups that classify batches or clusters instead of individual ads. On the other hand, the Ads Pairwise sub-problem is when discovering organized activity occurs by finding pairwise connections between ads; Ads Cluster is when it happens by clustering groups of related ads together; and the Other Structures class labels to problems that find an organization by relating other entities, not advertisements, e.g., locations, accounts, and phone numbers.

\begin{figure}
    \centering
    \includegraphics[width=0.65\textwidth]{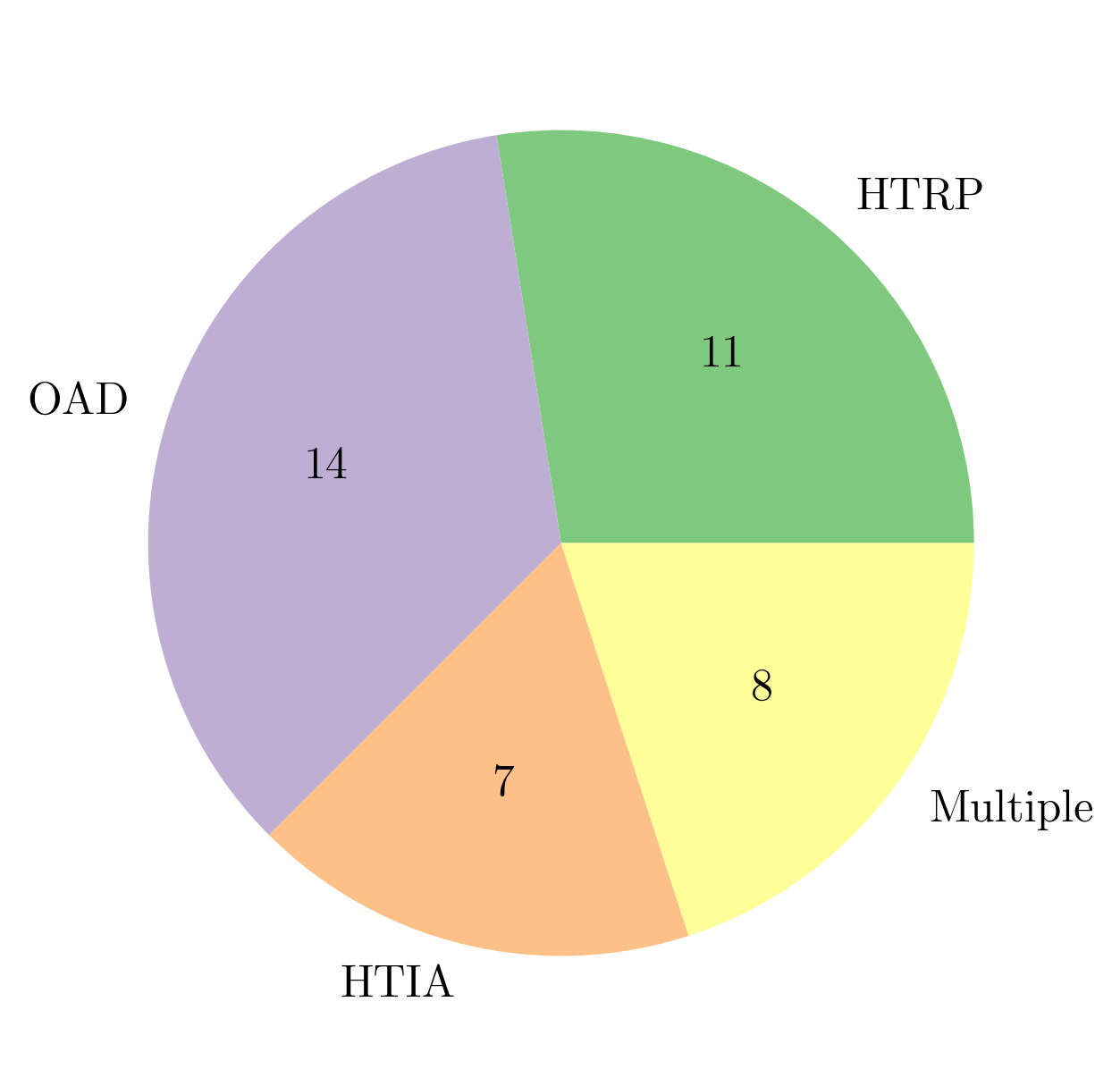}
    \caption{Distribution of the primary studies according to the problems addressed.}
    \label{fig:problems_pie}
\end{figure}

Figure \ref{fig:problems_pie} shows a distribution of the papers addressing each problem. Notice the prevalence of OAD detection in the collection reviewed and the significant number of papers that address several problems. Figure \ref{fig}

\begin{figure}
    \centering
    \includegraphics[width=\textwidth]{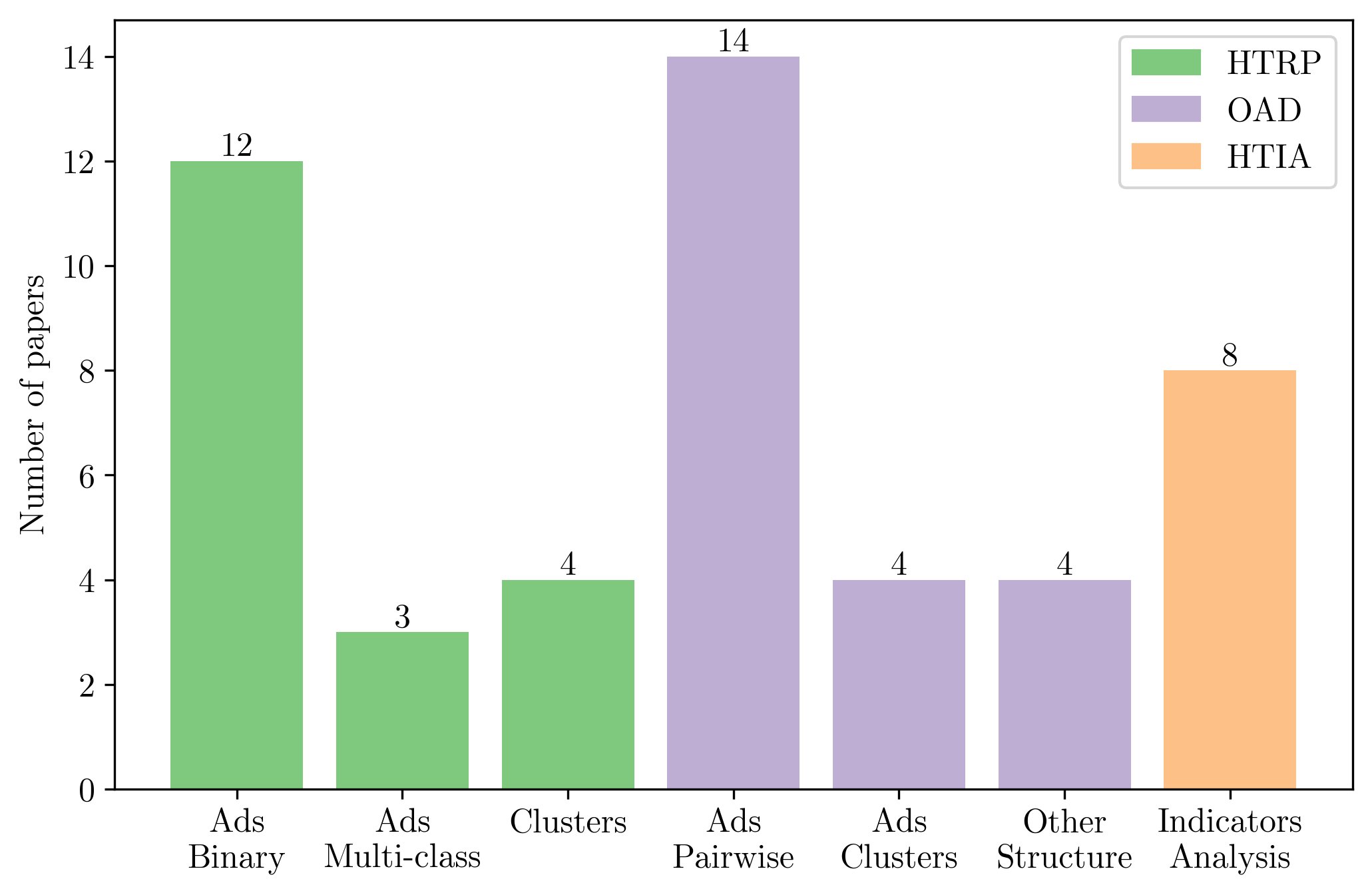}
    \caption{Distribution of the primary studies according to the sub-problems addressed.}
    \label{fig:problems_subproblems}
\end{figure}

Additionally, Figure \ref{fig:problems_subproblems}  shows the prevalence of each sub-problem considered. Observe that most work is devoted to the Ads Binary category in the case of HTRP and Ads Pairwise one in the case of OAD.

\subsubsection{Methods}

\begin{figure}
    \centering
    \begin{subfigure}[b]{.49\textwidth}
        \includegraphics[width=\textwidth]{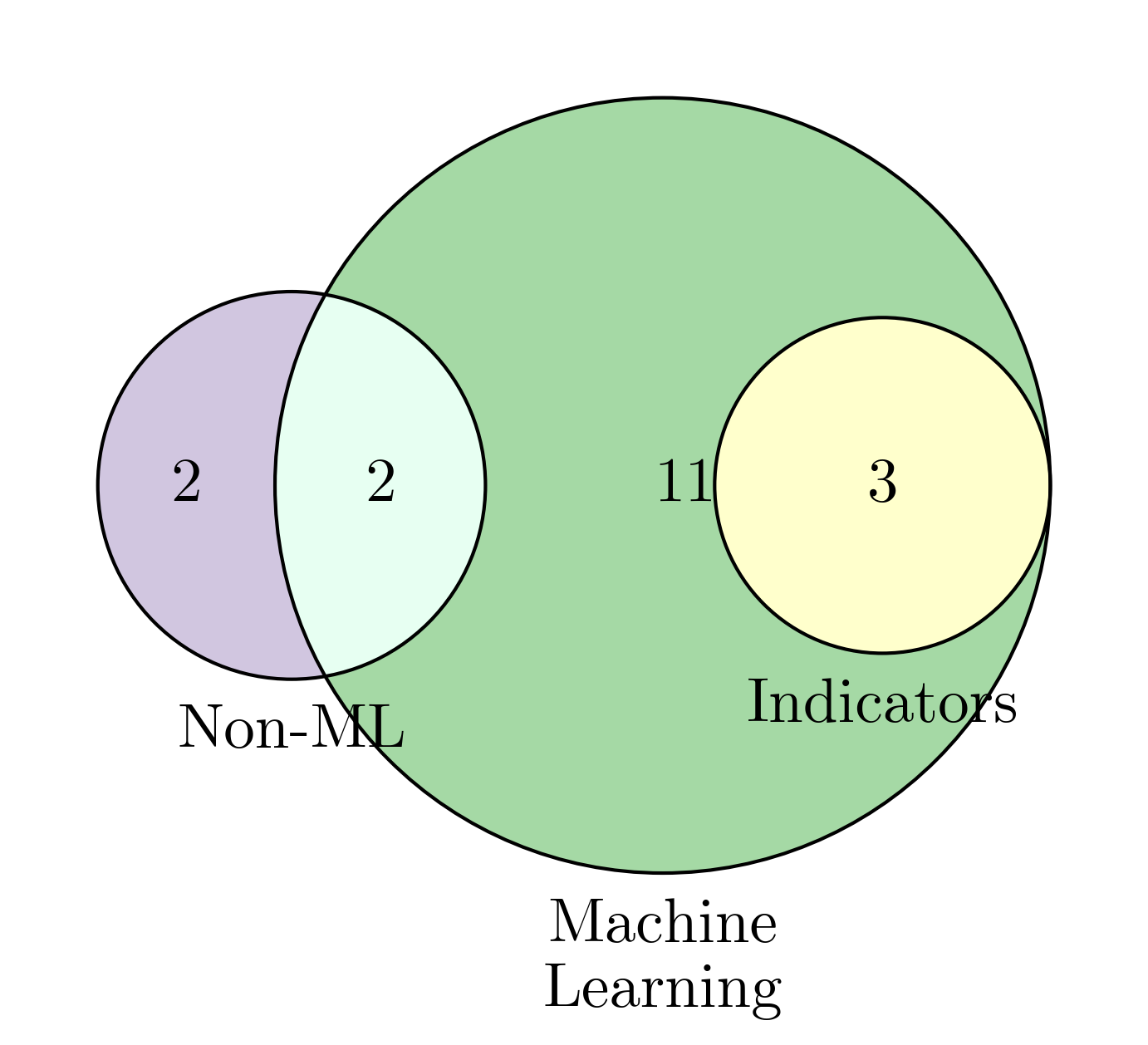}
        \caption{}
        \label{fig:methods_ht}
    \end{subfigure}
    \hfill
    \begin{subfigure}[b]{.49\textwidth}
        \includegraphics[width=\textwidth]{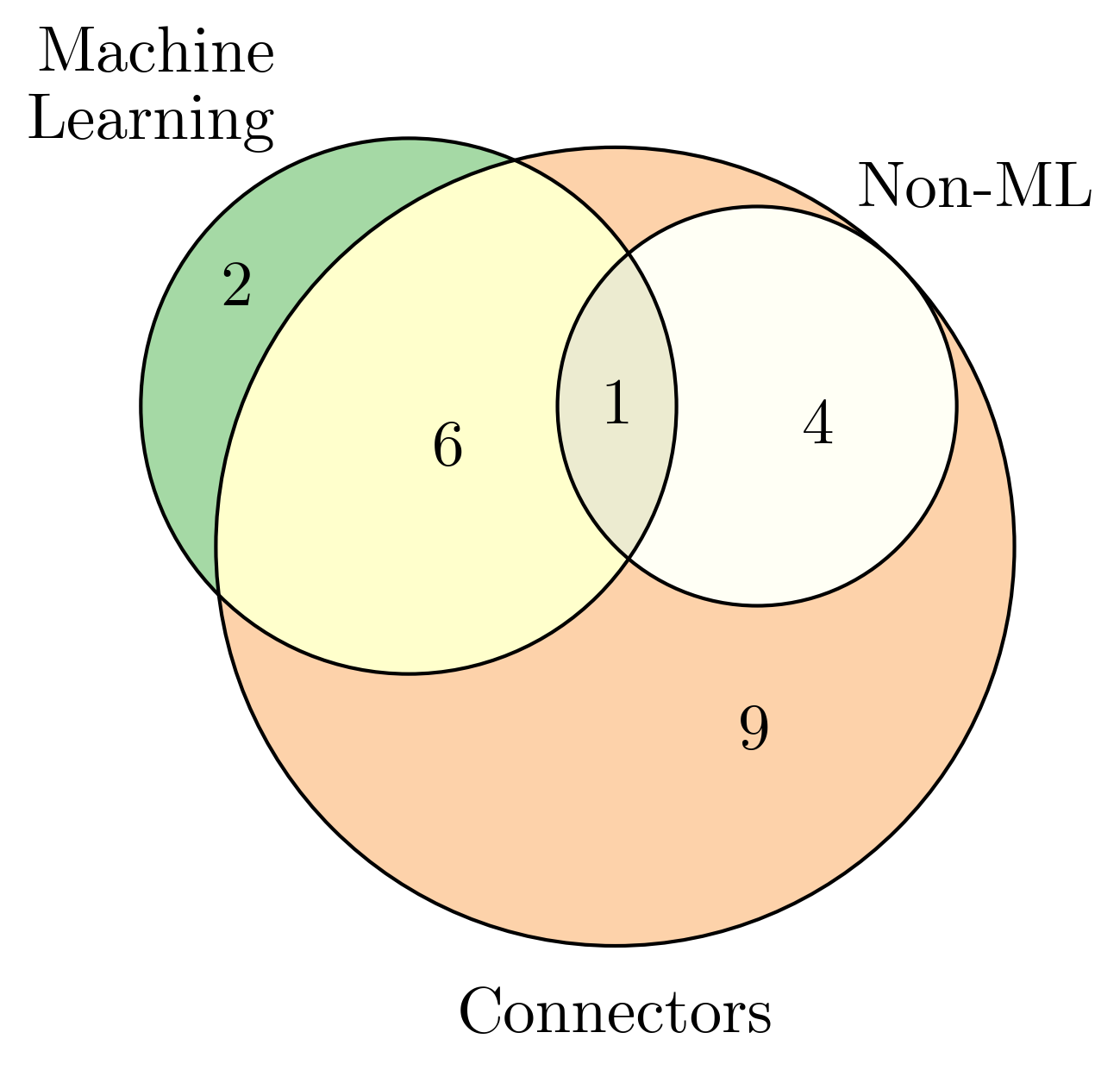}
        \caption{}
        \label{fig:methods_oa}
    \end{subfigure}
    \caption{Main methods for solving the (a) Human Trafficking Risk Prediction and (b) Organized Activity Detection problems.}
    \label{fig:methods_venn}
\end{figure}

Next, we elaborate on a variety of methods to solve both problems, namely, HTRP and OAD. Figure \ref{fig:methods_venn} highlights where papers fall as far as their methods are concerned when classified in one of three categories. Notice in Figure \ref{fig:methods_ht} we group the methods into three sub-groups: Machine Learning, Non-Machine-Learning, and Indicators; whereas in \ref{fig:methods_oa}, the first two are the same, but the third category is labeled Connectors. The Indicators category refers to classifying or labeling the data using a piece of law-enforcement intelligence, widely called indicators or proxies to human trafficking. We expand on this topic in the next section. On the other hand, the Connectors category refers to classifying or labeling the data using evidence found in the OEAs that is believed to establish that two ads come from the same source or organization. For example, when ads share a phone number, that is strong evidence that those are connected due to the operational nature of phone numbers~\shortcite{ibanez2014detection}.

\begin{figure}
    \centering
    \begin{subfigure}[b]{\textwidth}
        \centering
        \includegraphics[width=\textwidth]{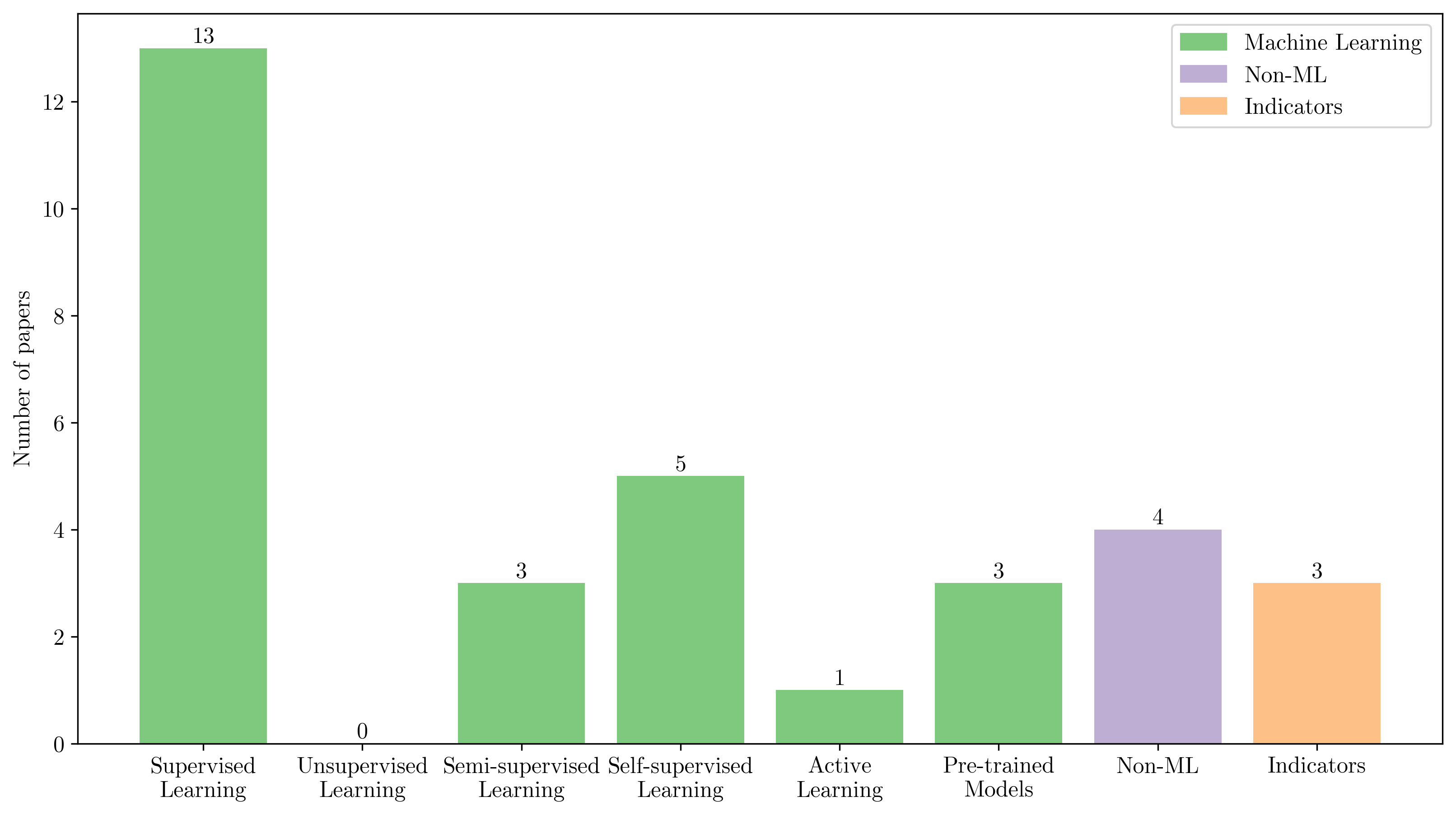}
        \caption{}
        \label{fig:mlmethods_ht}
    \end{subfigure}
    \begin{subfigure}[b]{\textwidth}
        \centering
        \includegraphics[width=\textwidth]{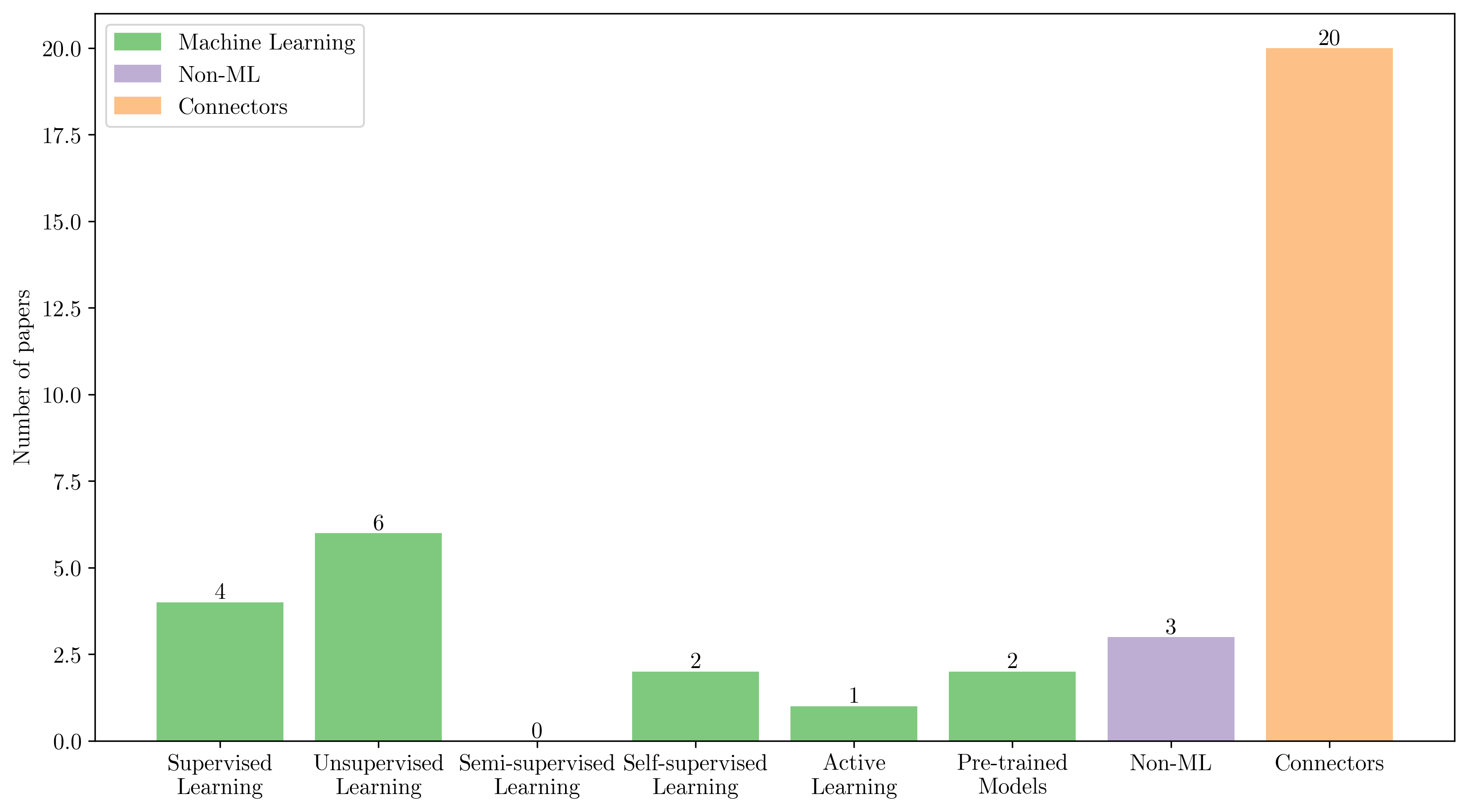}
        \caption{}
        \label{fig:mlmethods_oa}
    \end{subfigure}
    \caption{Machine Learning methods addressing the (a) Human Trafficking Risk Prediction and (b) Organized Activity Detection problems.}
    \label{fig:mlmethods}
\end{figure}

The better part of the studies analyzed leverage Machine Learning methods in their solutions. Figure \ref{fig:mlmethods} elaborates on the previous topic by expanding the Machine Learning category into smaller subcategories we discuss.

From the figure, we see a prevalence of supervised learning methods overall. In the supervised learning setting, the learner receives $N$ labeled samples, this is $\mathcal{S}=\{(x_1,y_1), \dots, (x_N,y_N)\}$. It is assumed there is a function that maps from $x_i$ to the corresponding $y_i$, possibly with some unknown stochastic error. The learning goal is to find a function~(a hypothesis) that approximates that mapping~\shortcite{mohri2018foundations}.

Framing HTRP as a supervised learning problem requires having information on whether an ad, or groups of those, are suspected instances of human trafficking activity. Papers reviewed have resorted to different strategies to label their data like
\begin{enumerate*}[label=(\roman*)]
  \item domain knowledge from experts~\shortcite{tong2017combating, nagpal2017entity} and
  \item law enforcement knowledge of known human trafficking cases or suspicious activity~\shortcite{dubrawski2015leveraging,hundman2018always, mensikova2018ensemble,esfahani2019context}.
\end{enumerate*}.

Conversely, for OAD, a supervised learning strategy requires having ground truth information of an underlying structure or relation between groups of ads. In this case, two strategies are worth highlighting:
\begin{enumerate*}[label=(\roman*)]
  \item resourcing to known or suspected trafficking organizations with knowledge coming from law enforcement~\shortcite{zhu2019detecting} and
  \item using connectors as a proxy for pairwise relation ground truth~\shortcite{zhu2019detecting,kulshrestha2021detection,rodriguez2021identifying}  
\end{enumerate*}. In this context, some connectors are often referred to as ``hard'' identifiers.

Several titles highlight the inconvenience of having limited labeled data or none at all~\shortcite{kejriwal2017flagit, rodriguez2021identifying}. Obtaining data on known human trafficking activity in OEAs is, at best, an expensive and complicated process~\shortcite{tong2017combating} for several reasons that include: the relatively low prevalence of human trafficking cases across sexual services platforms, the restrictions for accessing law enforcement information, and the subjectivity of labeling suspicious ads without concrete evidence of trafficking activity. Roughly the same applied to detecting organized activity, except for the advantage of having hard identifiers to connect ads. Abiding by these limitations, investigators have resorted to alternative methods with weaker supervision requirements like unsupervised, semi-supervised, and active learning.

In unsupervised learning, a collection of unlabeled samples $\mathcal{U}=\{x_1, \dots, x_N\}$ is given, and based on the structure of the data, a learner attempts to make predictions on unseen data. The semi-supervised setting allows labeled and unlabeled examples, usually a limited amount of labeled data. A learner needs to learn how to leverage the unlabeled data to improve its predictions on the labeled one. Finally, in the active learning setting, it is assumed that the learner can query an oracle for labeled examples, but it is constrained in how many queries it can make. Therefore, it must intelligently select which queries to make to maximize performance.

As observed in Figure \ref{fig:mlmethods}, unsupervised learning is only used in research that addresses detecting organized activity. In this case, clustering is predominantly leveraged as a solution to grouping ads that are similar in some latent space of features~\shortcite{hundman2018always,li2018detection,liu2019coupled,kulshrestha2021detection,rodriguez2021identifying}.

Conversely, semi-supervised learning appeared as a solution to circumvent the lack of labeled data for detecting human trafficking risk. \shortciteA{kejriwal2017flagit} train classifiers to predict the existence of certain indicators based on a small labeled dataset. Then, they sampled unlabeled ads from the extremes of the probability distribution and labeled those accordingly.

Additionally, \shortciteA{alvari2016nonparametric} use label propagation algorithms, namely, Label Propagation~\shortcite{zhur2002learning} and Label Spreading~\shortcite{zhou2003learning}. Finally, \shortciteA{alvari2017semi} propose $S^3V\text{-}R$, a novel semi-supervised learning method that extends a Laplacian SVM~(Support Vector Machine) by incorporating additional information from the feature space as a regularization term.

\shortciteA{ramchandani2021unmasking} use pool-based active learning to learn to distinguish recruitment versus sales advertisements, but instead of targeting at actively labeling posts that would minimize model uncertainty, they incorporate the likelihood of identifying new network connections so that the aggregate training set is not biased toward specific locations. On the other hand, \shortciteA{rabbany2018active} present an interactive method to extract connections of related ads connected to a seed. Importantly, the success of their method relies on the multi-modality of the scoring function to infer new connected ads and re-weighting the importance of each piece of evidence as the learning progresses based upon the feedback received from the oracle. 

One last Machine Learning paradigm we would like to mention is that of self-supervised learning. Self-supervised learning looks a lot similar to supervised learning, except that labels are not obtained through human supervision. Instead, the nature of the problems allows for a systematic definition of a learning target. A common application of this paradigm, and widely used throughout the literature reviewed, is the creation of word embeddings. Word embedding is a technique to represent text units, usually words, as vectors in a continuous space such that the geometry of that space has a correlation with the meaning of the words. We observed some such methods employed~\shortcite{tong2017combating,li2018detection,wang2019sex,esfahani2019context,ramchandani2021unmasking,rodriguez2021identifying} to featurize the text in the advertisements, for example, GloVe~\shortcite{pennington2014glove}, word2vec~\shortcite{mikolov2013efficient}, and FastText~\shortcite{joulin2017bag}.

\begin{figure}
    \centering
    \includegraphics[width=.9\textwidth]{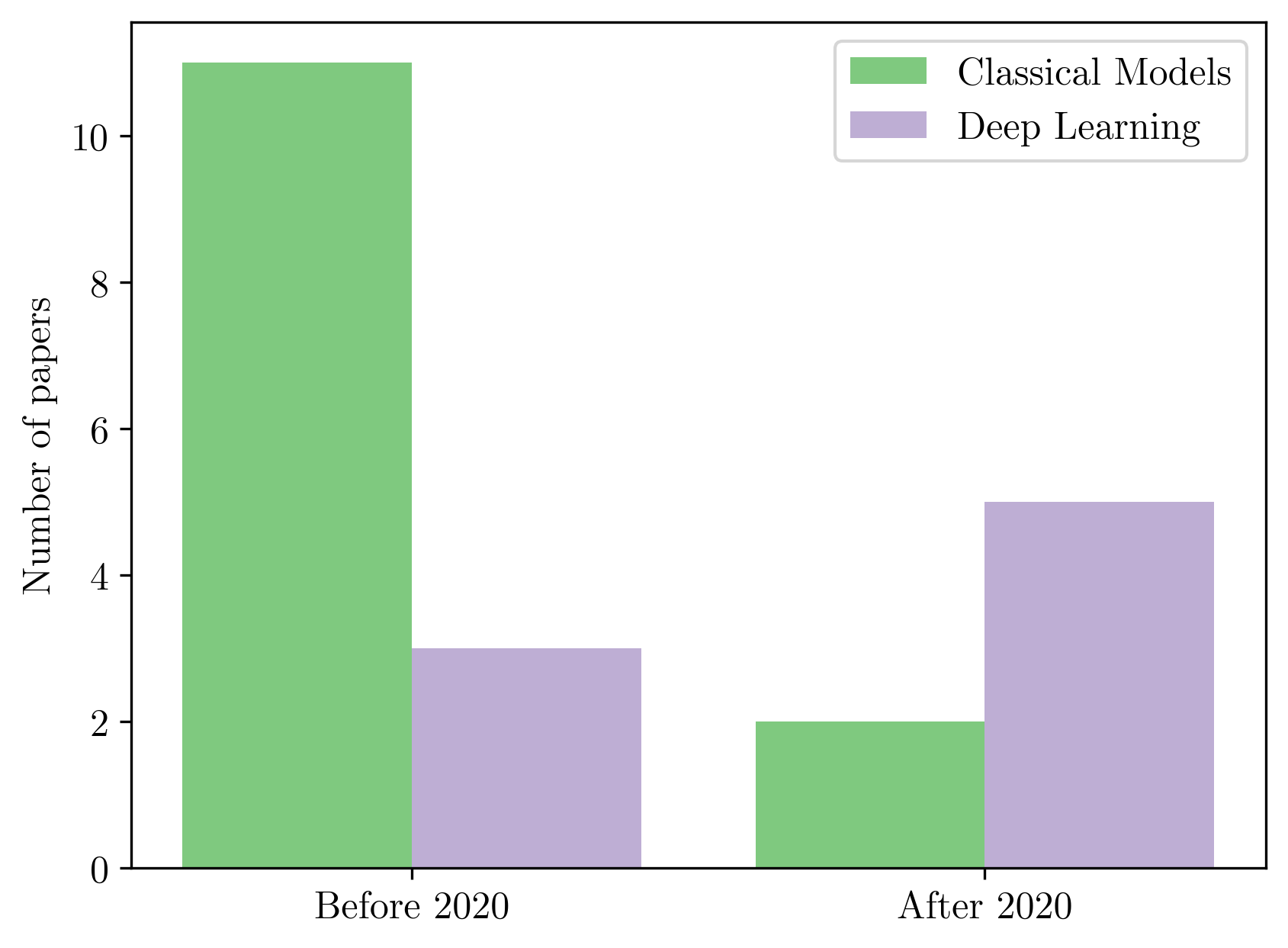}
    \caption{Deep Learning versus classical ML models trend.}
    \label{fig:problems_trend}
\end{figure}

Another aspect worth pointing out is the trend that favors Deep Learning solutions over classical simpler models in most areas where Machine Learning is applied. As we observed, the human trafficking domain does not escape this tendency, as showcased by Figure \ref{fig:problems_trend}.

\begin{figure}
    \centering
    \includegraphics[width=\textwidth]{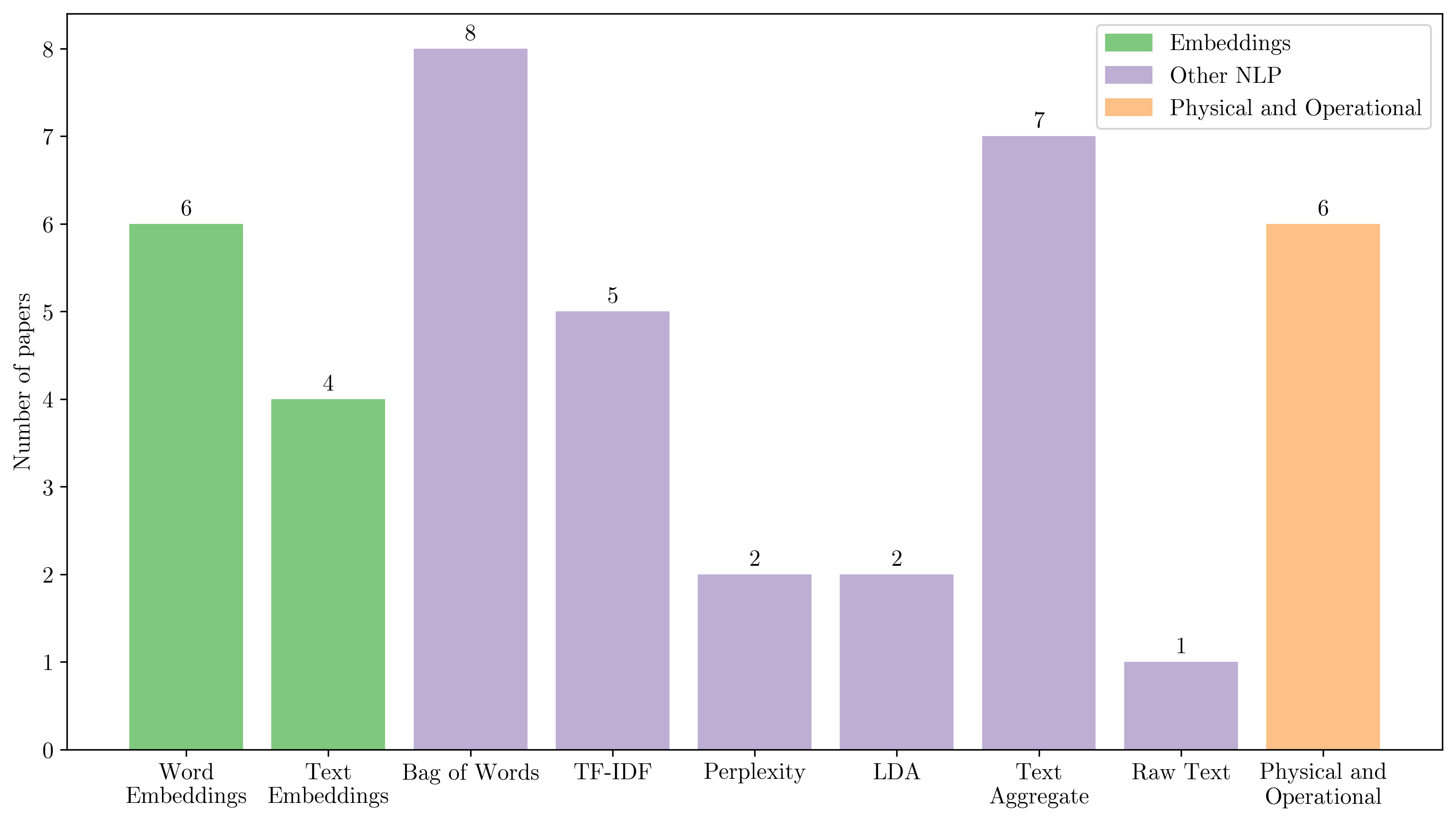}
    \caption{Features employed by the primary studies to feed Machine Learning models.}
    \label{fig:features}
\end{figure}

To close the topic of Machine Learning, let us refer to ways investigators have decided to represent their inputs numerically, namely, the features they have used to encode OEAs. Figure \ref{fig:features} shows a barplot with aggregated statistics of different categories of features as used by the papers studied. Notably, we consider three major categories:
\begin{enumerate*}
    \item embeddings,
    \item other NLP-related features, and
    \item domain-specific physical/operational features
\end{enumerate*}.

Within the embedding category, we distinguish word embeddings and text embeddings. As mentioned earlier, word embeddings encode words in a latent vectorial space, and the whole text is thus represented as a sequence of such vectors. Conversely, text embeddings encode a whole piece of text as a single vector.

In the Other NLP category, notice the presence of various subcategories. We called bag of words any featurization strategy in which a word, character, $n$-gram, etc., corresponds to a feature within the feature vector and considers some basic statistic about that word/character/$n$-gram, such as how many times it occurs within the text. For example, \shortciteA{dubrawski2015leveraging} extract counts of word occurrences and use the top 300 components of Principal Component Analysis~(PCA) as their text representation. \shortciteA{portnoff2017backpage}, in turn, utilize a subset of the WritePrint~\shortcite{zheng2006framework} feature set that includes counts of words, characters, and punctuation, among others. Lastly, \shortciteA{zhu2019detecting} computes various numeric values, including counts, on selected keywords deemed relevant to the domain.

Other NLP techniques include the Term Frequency - Inverse Document Frequency~(TF-IDF) method~\shortcite{manning2009introduction}, perplexity scores~\shortcite{manning1999foundations}, and Latent Dirichlet Allocation~(LDA)~\shortcite{blei2003latent}. Also, it is common to find numerical values being computed for the whole document other than bags of words, like average perplexity, length, etc. Finally, \shortciteA{summers2023multi} propose an end-to-end model, that is, they utilize the raw text to feed their neural network.

Physical and operational characteristics utilized as features included, but were not limited to, body measures, age, location information, and date/time information.

For completeness, a comprehensive list of identified Machine Learning methods, algorithms, and features used per paper can be found in the tables \ref{tab:ml_method}, \ref{tab:ml_algo}, and \ref{tab:ml_feat}, respectively.

Importantly, some work has been done outside the scope of Machine Learning. For example, in the case of HTRP, \shortciteA{vajiac2023deltashield} propose that, after their careful clustering of ads into different groups, any instance that belongs to a group should be suspected of human trafficking since their clustering shows evidence of organized activity. Additionally, \shortciteA{mensikova2018ensemble} propose using an ensemble of sentiment analysis models along with some heuristics related to negation and certain flagged locations to categorize an ad as risky. Also, in the context of HTRP, other investigators~\cite{zhu2019detecting,kulshrestha2021detection} have proposed ways to, after performing some form of clustering, come up with a classification of the degree of concern of the cluster altogether.

On the other hand, Non-ML methods to aid Organized Activity Detection include community detection algorithms over graphs~\cite{li2018detection}, a random walk-based connected components algorithm~\cite{kejriwal2022knowledge}, and a combination of sequence alignment and text compression techniques to create fine-grained clusters based on text similarity~\cite{vajiac2023deltashield}.

As illustrated in Figure \ref{fig:methods_ht}, we marked three papers as using indicators for HTRP. \shortciteA{kejriwal2017flagit} train models that learn to predict the presence of several indicators and consequently flag advertisements as suspicious of being involved in human trafficking activity. Moreover, \shortciteA{ramchandani2021unmasking} proposes identifying the supply chain of human trafficking activity in OEAs. They first identify which ads constitute recruiting and which advertise a service offered. Then, after constructing a network of advertisements, any group of ads that contains both recruiting and sale activity is suspected to be conducting some form of illegal activity associated with human trafficking. We categorize these heuristics as an indicator. Finally, \shortciteA{alvari2016nonparametric} use indicators to filter out less suspicious ads before passing the data to experts for labeling.

As mentioned earlier, detecting organized activity~(or relationships between OEAs that could indicate such), is sometimes framed as a pairwise similarity problem, where a pair of ads is classified as either related or not. This problem finds a lot of different names in the literature: the match function in an Entity Resolution algorithm~\shortcite{dubrawski2015leveraging}, Authorship Classification~\shortcite{portnoff2017backpage}, Record De-duplication~\shortcite{keskin2021cracking}, Alias Detection~\shortcite{wang2012data}, among others. In this context, the use of connectors is commonplace, either as ground truth for supervised machine learning algorithms~\shortcite{dubrawski2015leveraging,portnoff2017backpage,nagpal2017entity,rodriguez2021identifying}; or as the proposed solution to construct a network for further analyses~\shortcite{wang2012data,silva2014data,szekely2015building,dubrawski2015leveraging,hundman2018always,zhu2019detecting,kejriwal2019network,kejriwal2020network,keskin2021cracking,ramchandani2021unmasking,vajiac2022trafficvis,kejriwal2022knowledge,vajiac2023deltashield}.

\subsubsection{Results}

This section will expand on the results reported in the literature reviewed. Unfortunately, a major gap identified throughout our study is the lack of benchmarking datasets. There is but one dataset consistently used in multiple studies, and it is the only one on which we can report comparative results. The dataset is aimed at the HTRP task and was coined Trafficking-10k~\shortcite{tong2017combating}. This dataset consists of over 10,000 OEAs from a former website called Backapge, annotated by experts according to how suspicious an ad appears to be of human trafficking in one of seven classes: \textit{Certainly no}, \textit{Likely no}, \textit{Weakly
no}, \textit{Unsure}, \textit{Weakly yes}, \textit{Likely yes}, and \textit{Certainly yes}. This scale can and has been collapsed into a binary classification problem, usually by mapping the first three classes to the negative label, the last three to the positive label, and the Unsure class ignored. The dataset classes are not balanced, and the negative classes amount to roughly twice the ads in the positive classes.

Table \ref{tab:t10k_multi} shows the reported results of different studies in the multi-class classification version of the problem. Conversely, \ref{tab:t10k_bin} serves the same purpose for the binary version.

\begin{table}
    \centering
    \caption{Results in the Trafficking-10k dataset in the multi-class version.}
    \begin{tabular}{|c|c|c|c|}
         \hline
         \textbf{Paper} & \textbf{Accuracy} & \textbf{\makecell{Weighted\\Accuracy}} & \textbf{$\mathbf{F_1}$ Score} \\
         \hline
         \shortciteA{tong2017combating} & 0.8 & 0.753  & 0.665 \\
         \shortciteA{wang2019sex} & 0.818 & 0.772  & - \\ 
         \shortciteA{wiriyakun2021feature} & 0.77 & - & -\\
         \hline
    \end{tabular}
    
    \label{tab:t10k_multi}
\end{table}

\begin{table}
    \centering
    \caption{Results in the Trafficking-10k dataset in the binary version.}
    \begin{tabular}{|c|c|c|}
         \hline
         \textbf{Paper} & \textbf{Accuracy} & \textbf{$\mathbf{F_1}$ Score} \\
         \hline
         \shortciteA{zhu2019identification} & - & 0.696\\
         \shortciteA{wiriyakun2022extracting} & - & 0.648 \\
         \shortciteA{summers2023multi} & 0.82 & 0.69 \\
         \shortciteA{vajiac2023deltashield} & - & 0.635 \\
         \hline
    \end{tabular}
    
    \label{tab:t10k_bin}
\end{table}

From the table, we see that the state-of-the-art performance in the multi-class setting is achieved by \shortciteA{wang2019sex}. The authors employ word embeddings pre-trained in a corpus of Backpage advertisements and a deep neural network architecture for classification that consists of a Gated-Feedback Recurrent Neural Network~\shortcite{chung2015gated}. The authors utilize an ordinal regression to model the problem, and the last layer in their architecture is tailored correspondingly.

On the other hand, \shortciteA{zhu2019identification} hold the state of the art as far as the $F_1$ score is concerned in the binary setting. They developed a language model in which some phrases are regarded as a single token. Two strategies for feature selection are presented. First, those words or phrases that occur most frequently as determined by a threshold are selected. In the second variant, an L1-penalized SVM is trained to select the most important features and discard the sparse ones. In either case, they leverage average perplexity and TF-IDF as the numerical values for each feature and use a logistic regression model for classification.

\subsection{RQ2: Which indicators of human trafficking found in online escort ads text are used to predict human trafficking?}\label{subsec:rq2}

Several institutions have developed guidelines that describe the nature and characteristics of human trafficking~\shortcite{unodc2012human,jrsa2022law,office2023modern}. Through collaboration with law enforcement experts, researchers have also defined and operationalized indicators that could suggest the presence of human trafficking in OEAs. These indicators are useful not only for their ability as predictors of human trafficking but also for the interpretability of the results. As has been pointed out by research~\shortcite{hundman2018always}, not only do law enforcement agents seek a system that can automate the detection and facilitate their work, but they also necessitate strong evidence of the claims the system makes so that they can further investigate and gather concrete proof.

As Figure \ref{fig:problems_subproblems} shows, there is a group of papers we include in our research that specifically explains some of these indicators. This section expands on the indicators found mostly in this piece of the literature.

\subsubsection{Multiple victims} Human trafficking is, by definition, an organized activity. Any evidence found that multiple victims are being promoted by the same organization strongly suggests that trafficking is occurring. We found this indicator is most often detected by (i) spotting first-person plural language in individual ads or simply multiple provider names mentioned~\shortcite{kennedy2012predictive,ibanez2014detection,ibanez2016detecting,alvari2016nonparametric,hultgren2016using,alvari2017semi,giommoni2021identifying,l2021identifying}, by (ii) finding multiple hard identifiers~(phone numbers, names, emails) in groups of related ads~\shortcite{hultgren2015exploratory,ibanez2016detecting,giommoni2021identifying,l2021identifying}, and by (iii) using a glossary of predefined terms or keywords~\shortcite{kejriwal2017flagit,hultgren2016using}.

\subsubsection{Business} It is known that massage parlors and night clubs are sometimes a cover for illicit trafficking activity~\shortcite{oregon2019nationwide}. That is why researchers have determined that whenever a business activity like this is advertised on escort websites, it could bear the risk of human trafficking. Concretely, this indicator takes one of two forms: either (i) the full address of the business is advertised~\shortcite{ibanez2016detecting,alvari2017semi} or (ii) a reference to an external business is given is provided~\shortcite{kennedy2012predictive,alvari2016nonparametric}. We need to make a special mention of the work by \shortciteA{ramchandani2021unmasking} in which they propose to target groups of ads where there was evidence of both selling and recruiting activity. This particular intuition was also labeled under this category.

\subsubsection{Shared Management} Existing evidence that sex workers are controlled or managed by others is considered a huge red flag of human trafficking~\shortcite{unodc2012human}. We observed two ways of operationalizing this indicator. First, (i) using third-person language in the ad description signals that the service is being posted by a person different than the one delivering the service~\shortcite{kennedy2012predictive,ibanez2016detecting,alvari2016nonparametric,alvari2017semi,l2021identifying}. Second, (ii) common markers appearing in different ads, potentially advertising different victims, signal that the same person or group might be controlling others~\shortcite{kennedy2012predictive,hultgren2016using,l2021identifying}. For example, when the same phone number appears in multiple ads.

Notice the first three categories described all point out the organizational nature of the human trafficking activity. The readers might find themselves exchanging some of the meaning. Remember that the categories we present are just how we decided to categorize them for better understanding and not a tight boundary on the definitions.

\subsubsection{Mobility of victims} The sex market involves satisfying clients with varied and fresh options. Also, pimps need to keep their activities transient to avoid being discovered~\shortciteA{whitney2017km}. Therefore, the scenario of victims moving across wide areas and/or only staying in a single place for a short period of time constitutes an important indicator of human trafficking. Identifying the presence of this indicator is most commonly approached in two ways: (i) looking for a specific set of keywords indicative of transient language~(e.g. ``new in town'', ``just arrived'', ``limited time'', etc)~\shortcite{ibanez2014detection,hultgren2016using,kejriwal2017flagit,whitney2018dont,l2021identifying}; and by (ii) finding evidence that a provider has availability in multiple locations; for example, the area code of a phone number is different from the advertised location~\shortcite{ibanez2014detection,hultgren2016using,ibanez2016detecting,hultgren2018knowledge}, similar or connected ads are across different spread-out locations~\shortcite{kennedy2012predictive,ibanez2016detecting,ibanez2016circuits}, and a provider deliberately advertises they are willing to move to many locations~\shortcite{giommoni2021identifying,l2021identifying}.

\subsubsection{Risky activity} Sex workers willing to provide dangerous or violent services or simply cannot refuse to do so are at a higher risk of being exploited~\shortcite{unodc2012human}. This activity is also referred to in the literature as unconventional sex. Our findings indicate that this indicator is detected through keywords that advertise those dangerous services or other phrases that imply the victim is willing to let the customer do as they please, like ``open-minded'', ``fetish'', etc~\shortcite{kejriwal2017flagit,hultgren2018knowledge,giommoni2021identifying,l2021identifying}.

\subsubsection{Restricted movement} Sex workers with restricted freedom of movement may indicate that they are being controlled by others and, consequently, being trafficked. Only so much can be done with advertisement information regarding this issue. However, as little as can be done, it is common for studies to search for specific keywords that may indicate a victim's movement restrictions like ``in-call only'', ``no outcalls'', etc., as an indicator of human trafficking~\shortcite{ibanez2014detection,hultgren2016using,ibanez2016detecting, whitney2018dont,hultgren2018knowledge,giommoni2021identifying,l2021identifying}.

\subsubsection{Underage victims} The sexual exploitation of minors is illegal and considered human trafficking. Researchers have resorted to looking for keywords that allude to the youth of the victims~\shortcite{kennedy2012predictive,alvari2016nonparametric,alvari2017semi,hultgren2016using,hultgren2018knowledge,whitney2018dont,giommoni2021identifying,l2021identifying}. Some work has also leveraged age prediction of the victims based on the body measures advertised~\shortcite{kennedy2012predictive,alvari2016nonparametric,alvari2017semi,giommoni2021identifying,l2021identifying}.

\subsubsection{Ethnicity/nationality} It has been shown that clients of sexual services sometimes show preferences for certain races or national origins of the providers. In an attempt to tailor the description to a target group, providers may post ads where their ethnicity and nationality are disclosed~\shortcite{whitney2017km,jrsa2022law}. It was originally proposed by \shortciteA{ibanez2014detection}, without a thorough explanation, to use the presence of ethnicity or nationality as an indicator of sex trafficking activity. Nevertheless, this approach has been replicated~\shortcite{hultgren2016using,ibanez2016detecting,alvari2016nonparametric,alvari2017semi,whitney2018dont, hultgren2018knowledge,l2021identifying}.

\begin{figure}
    \centering
    \includegraphics[width=\textwidth]{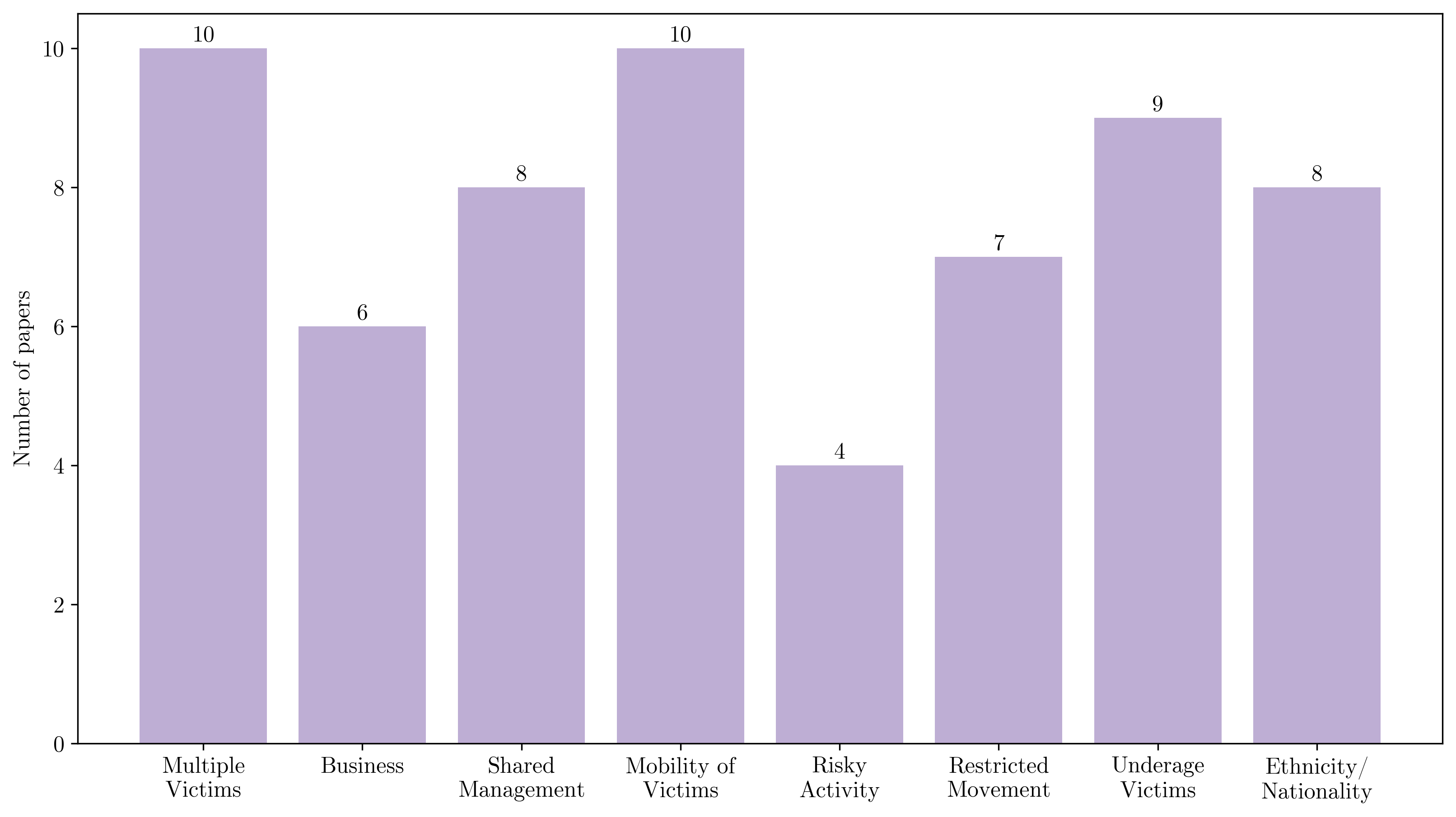}
    \caption{Distribution of indicators of human trafficking mentions across the primary studies.}
    \label{fig:indicators}
\end{figure}

Figure \ref{fig:indicators} shows the prevalence of some of the most important indicators found in the literature. As observed, the distribution is spread out, and the most used indicators are Multiple Victims, Mobility of Victims, and Underage Victims.

\subsubsection{Others} The eight indicators mentioned before are not the only ones referenced in the literature. Others include, for example, advertising services with a low price~\shortcite{l2021identifying,giommoni2021identifying} and also showing inconsistent stories~\shortcite{ibanez2014detection,l2021identifying}. Other research points out that simply advertising sexual services for sale indicates human trafficking~\shortcite{whitney2018dont}. Whereas \shortciteA{kejriwal2017flagit} state that references to incall or outcall services, regardless, is also a red flag. Poor grammar was suggested as an indicator by \shortciteA{l2021identifying} and \shortciteA{mensikova2018ensemble} showed that the ads' text sentiment is correlated in some capacity with human trafficking risk. Finally, \shortciteA{hultgren2016using} suggested the presence of disguised phone numbers, listing many services, low-quality images, and distress detected in subjects as a result of processing the ad images are all potential clues of human trafficking.

For completeness, Table \ref{tab:indicators} contains a comprehensive list of indicators found in each document referencing at least one.

\subsection{RQ3: What feature of escort advertisements can be leveraged to build connections between different ads and identify a network of providers?}

Human trafficking is an organized activity. Consequently, criminals operate in groups and create organizations to support their activities. Evidence of organized sexual services is an indicator itself that trafficking could be taking place. Moreover, discovering an underlying network of providers possibly engaging in illegal activity could yield to detecting patterns and trends beyond the limited information provided by an individual post, like movement trends, organizational structure, organization scope, and operations size, thus contributing to facilitating the job of law enforcement forces to combat human trafficking.

We argued in the sub-section dedicated to RQ1 that pairwise relation prediction was a typical method used to find network structures in collections of ads. Moreover, we discovered that it was commonplace to use connectors as a heuristic or proxy for the existence of those relations. This section looks closely at these connectors and their use in the literature we inspected.

\begin{figure}
    \centering
    \includegraphics[width=\textwidth]{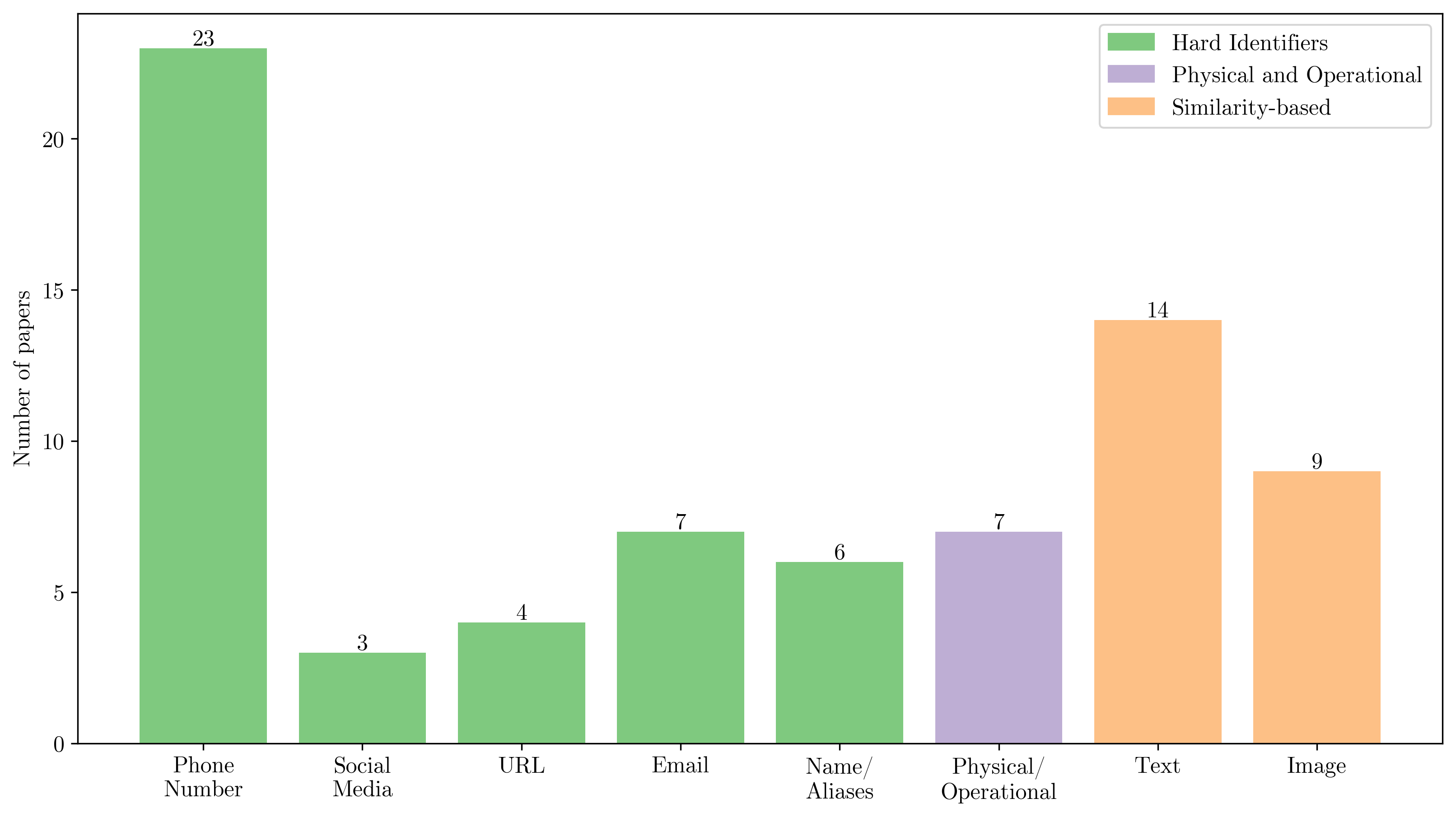}
    \caption{Distribution of connectors mentions across the primary studies.}
    \label{fig:connectors}
\end{figure}

First, observe in Figure \ref{fig:connectors} the number of papers that utilize the most common connectors identified. The first (green) category groups the so-called hard identifiers. Hard identifiers are elements that strongly link the referenced items because of how tightly those are bound to someone's identity: phone numbers, social media handles, personal or business website URLs, email accounts, and names or aliases, among others not included in the plot. We observed that phone numbers are preferred for two main reasons. First, phone numbers are a common way of communication among criminals and between clients and service providers~\shortcite{ibanez2014detection,keskin2021cracking}. Second, law enforcement has information on suspicious activities tied to phone numbers~\shortcite{dubrawski2015leveraging,hundman2018always}. However, working with phone numbers~(and hard identifiers in general) has reported limitations. For instance, those are not always available in the ad description or are difficult to extract because they appear intentionally disguised or in code language~\shortcite{kennedy2012predictive,wang2012data,hundman2018always}. Moreover, phone numbers are not static: people may change phone numbers, which could introduce bias to the data~\shortcite{portnoff2017backpage,keskin2021cracking}.

Physical and operational characteristics can also be leveraged to connect entities. Since each physical or operational feature is not very discriminative by itself, approaches leveraging this method usually do so in a multifactorial manner, requiring several features to match between two ads to confidently claim they are connected~\shortcite{dubrawski2015leveraging,nagpal2017entity}.

Finally, the last category groups papers claiming that similarities in OEAs' textual descriptions or images can connect and associate them to the same source. Text similarity approaches we observed include exact text matching~\shortcite{keskin2021cracking}, longest common sub-string~\shortcite{nagpal2017entity}, and closeness in some latent vector space~\shortcite{silva2014data,giommoni2021identifying}. Notably, in our data synthesis, this category could also refer to papers that use textual information as input features of Machine Learning models to predict the relatedness of advertisements~\shortcite{dubrawski2015leveraging,portnoff2017backpage,li2018detection}.

On the other hand, images are also leveraged, either by determining if the exact same image is shared across ads~\shortcite{wang2012data}, computing image similarity~\shortcite{silva2014data,szekely2015building,keskin2021cracking}, or extracting features from the images~\shortcite{hundman2018always}.

It is important to notice that, although vastly predominant, connections are not always constructed between advertisement instances. For instance, \shortciteA{kejriwal2020network} construct what they define as Activity Networks, where user accounts are nodes in the graph and the edges are different kinds of real-life activity connecting those accounts. Aiming as well at detecting organized activity, \shortciteA{liu2019coupled} propose the Coupled Clustering of Time-series and their underlying Network problem. The paper assumes entities are associated with time series behaviors and weighted connections between them. The goal is to construct both temporally and structurally coherent clusters. In their use case for OEAs, the entities, i.e., the nodes in the graph, are phone numbers. Finally, other research has highlighted the importance of finding relationships between different locations to identify circuits of illegal activity~\shortcite{ibanez2014detection,ibanez2016detecting}.

For completeness, Table \ref{tab:connectors} contains a comprehensive list of connectors found in each document referencing at least one.

\subsection{RQ4: What are the characteristics of the escort advertisement data used in the relevant literature?}

This section explores the nature of the data collected and used to develop and train the models studied. As described in the objectives of this study, the main data source in the documents reviewed is online escort websites.

Plenty of research has highlighted how the advent of technology facilitates human trafficking~\shortcite{ibanez2014detection,latonero2011human}. In particular, online classifieds websites play an important role in promoting prostitution and sex services in general~\shortcite{latonero2011human,kennedy2012predictive}. On the other hand, the fact that this information is publicly available makes it easier for researchers and law enforcement agencies to study the patterns that arise and develop strategies to counter illegal activities. Every primary study cited in this chapter is evidence of that.

Online escort advertisement data is difficult to deal with. The language in ads is noisy, with misspellings, intentional obfuscation, and strange symbols~\shortcite{zhu2019detecting}. The information provided could also be misguiding. It is commonplace, for instance, to announce a fake age, fake personal descriptions, and aliases instead of names, all with the intention of misleading law enforcement and sounding more appealing to clients~\shortcite{kennedy2012predictive}.

There is a long list of online classified websites that offer escort services. Perhaps the most notable one is the case of Backpage. Backpage was an online classifieds website that operated globally and advertised several categories of products, including adult services. It became the most important website for adult services~\shortcite{kennedy2012predictive,ibanez2014detection} before it was shut down in 2018 by the US federal agencies, who also confiscated their servers and data~\shortcite{kejriwal2020network}.

\begin{figure}
    \centering
    \includegraphics[width=\textwidth]{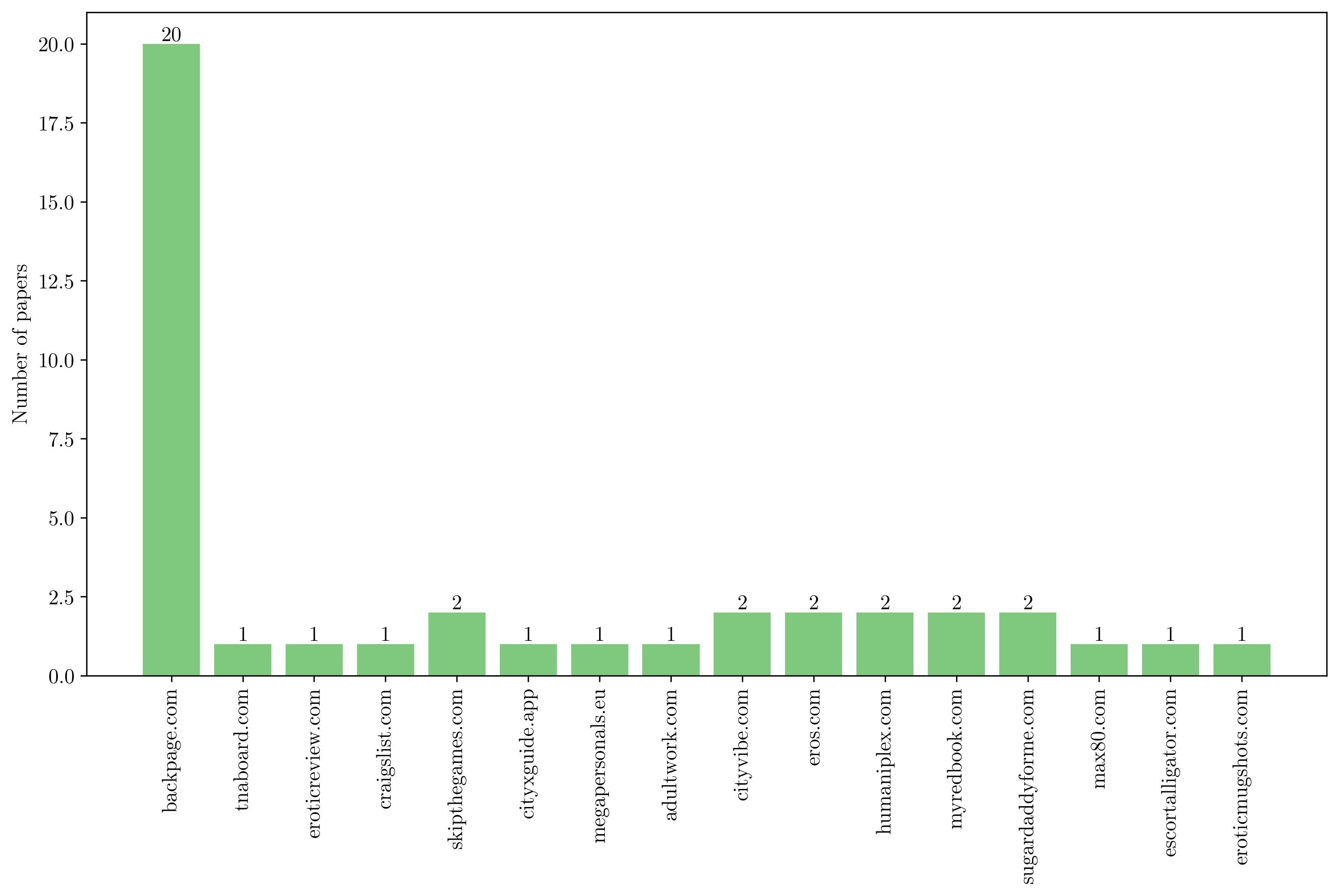}
    \caption{Escort advertisement websites distribution across the primary studies.}
    \label{fig:websites}
\end{figure}

As a measure of the degree of importance attributed to Backpage, note in Figure \ref{fig:websites} different classifieds websites used to retrieve data for the primary studies inspected in this systematic review and the corresponding number of papers that reference each.

After Backapage was shut down, the users and corresponding traffic moved to other websites that have consequently become more popular. But even under those circumstances, researchers have continued to use data from Backpage obtained through law enforcement agencies~\shortcite{kejriwal2020network}.

According to our study, around $141,139,081$ individual advertisements have been scraped from the web. In aggregate, the reported scraped or publication dates in that set cover all months, with no gaps, from January 2008 to February 2020 and the March-June period of 2021. Most of the data collected is sourced in the US. Other important sources include the UK~\shortcite{kejriwal2020network}, and several have used worldwide data~\shortcite{portnoff2017backpage,giommoni2021identifying,ramchandani2021unmasking}. Apart from the textual description of the ad, which we assume is always extracted, other common metadata fields scraped include the ad's title, location, date and time information, age, phone numbers, ad's URL, and images.

Despite its vast and comprehensive nature, little of this data has been made publicly available due to its sensitivity. This is particularly problematic in a context where Machine Learning could prove very useful because most modern Machine Learning methods are data-hungry. The absence of common benchmarks also makes it complicated to compare different approaches.

One outstanding exception to that rule is the Trafficking-10k dataset~\shortcite{tong2017combating}. Trafficking-10k is a dataset annotated for the detection of human trafficking, which includes more than 10,000 trafficking ads labeled with likelihoods of having been posted by traffickers. Even though the source is not mentioned in the original paper, it is known that the ads come from Backpage~\shortcite{zhu2019detecting}. It was annotated by three experts with experience in the human trafficking domain. The dataset consists of text and images, and the ads were sampled randomly from a large set of escort ads for annotation. The text in the dataset's instances is the plain text that results from removing HTML tags from the scraped source. The images are RGB and were resized to 224 $\times$ 224 pixels. The dataset is diverse and covers various advertisements across the United States and Canada. The sub-section on RQ1 provided a comparative summary of the results of different models evaluated against this dataset.

Despite its popularity, some issues have been reported. For example, it has been mentioned that exact duplicate ads had conflicting annotations~\shortcite{zhu2019detecting,zhu2019identification}. Consequently, some authors have left the Unsure class out of their analysis~\shortcite{zhu2019detecting,zhu2019identification,wiriyakun2022extracting}. As part of our research goals, we were interested in working with the Trafficking-10k dataset and did the due diligence to obtain it. However, we were not able to get access to it.

\section{Conclusion}

This chapter presented a Systematic Literature Review of existing literature on human trafficking analysis via computational methods in online escort websites. We followed a systematic procedure to investigate existing research, filter relevant documents, extract useful information to answer the research questions, and analyze the data collected.

In our study, we identified two main problems that are addressed in this context: Human Trafficking Risk Prediction~(HTRP) and Organized Activity Detection~(OAD). Importantly, many papers use Machine Learning methods to address the task at hand. In this area, we observed that supervised learning prevails in proposed solutions for HTRP, whereas unsupervised learning techniques abound in papers addressing OAD. Importantly, despite a trend to move towards deep learning-based approaches, there is still recent work dispensing with it in a domain where the text and image processing tasks are primordial and could, consequently, benefit a lot from state-of-the-art models in those areas.

Another important set of documents was deemed relevant because they explained heuristics used as proxies to identify human trafficking activity. These proxies, called indicators, are very relevant since they provide domain experts with an interpretable framework to derive predictions. In most cases, highly cited indicators are related to the operational and organized nature of the human trafficking activity and are spotted via regular expression matching or keywords from a curated list.

Finally, we inspected the characteristics of the data described in the literature. We found that authors needed to collect data from publicly available sources in most cases because high-quality data is lacking or not publicly available due to its sensitivity. The absence of publicly available datasets opens an important gap in the field and negatively impacts its advance.
%

\chapter{Creation of a Dataset for Human Trafficking Risk Prediction and Organized Activity Detection in Online Escort Advertisements}\label{chpt:data}

\section{Introduction}

The study in Chapter \ref{chpt:slr} allowed us to conclude, among other things, that data in the human trafficking~(HT) domain is not easy to obtain and curate. Labeling instances of human trafficking requires the collaboration of academic experts, law enforcement, and, sometimes, victims of sex exploitation. Such efforts are not easy to coordinate. By proposing heuristics as proxies to human trafficking activity and semi-automated labeling processes, researchers have been able to circumvent these difficulties. Moreover, it was often the case that even raw data was not readily available, and considerable effort was invested in obtaining it from publicly available sources with limited success~\shortcite{dubrawski2015leveraging, portnoff2017backpage, hundman2018always}.

Language plays an essential role in illicit online activity, including sex trafficking. Code communications allow criminals to hide their personal information and real intentions, a jargon is built around sexual services that is well understood by clients and providers, and similarities in the language can be used to identify underlying connections between potential offenders. However, the lack of up-to-date data makes it complicated to study the subject because this language evolves and adapts to social and cultural changes~\shortcite{dubrawski2015leveraging}. Additionally, traffickers may modify their posts as a mechanism to avoid detection~\shortcite{latonero2011human}. Therefore, a model could easily be rendered obsolete if it was developed under assumptions rooted in outdated research.

These reasons motivate us to design a better process. Concretely, this chapter elaborates on a robust methodology we developed to collect, process, and label data coming from online escort advertisements. Namely, we describe the nuances of a scraping process, how to modify raw data adequately, and how to generate a final dataset based on a pseudo-labeling process. Importantly, part of the method includes performing Named Entity Recognition~(NER) to extract relevant information from the raw text of the ads. Consequently, we devote significant space to explaining the methods we relied on and our results.

\section{Methodology}

The ultimate goal of the research we discuss here is to obtain two labeled datasets for the tasks of Human Trafficking Risk Prediction~(HTRP) and Organized Activity Detection~(OAD), the binary and ads pairwise variants, respectively~(as defined in Chapter \ref{chpt:slr}). Importantly, our methodology aims to require minimal human supervision throughout the entire process.

\begin{figure}
    \centering
    \includegraphics[width=\textwidth]{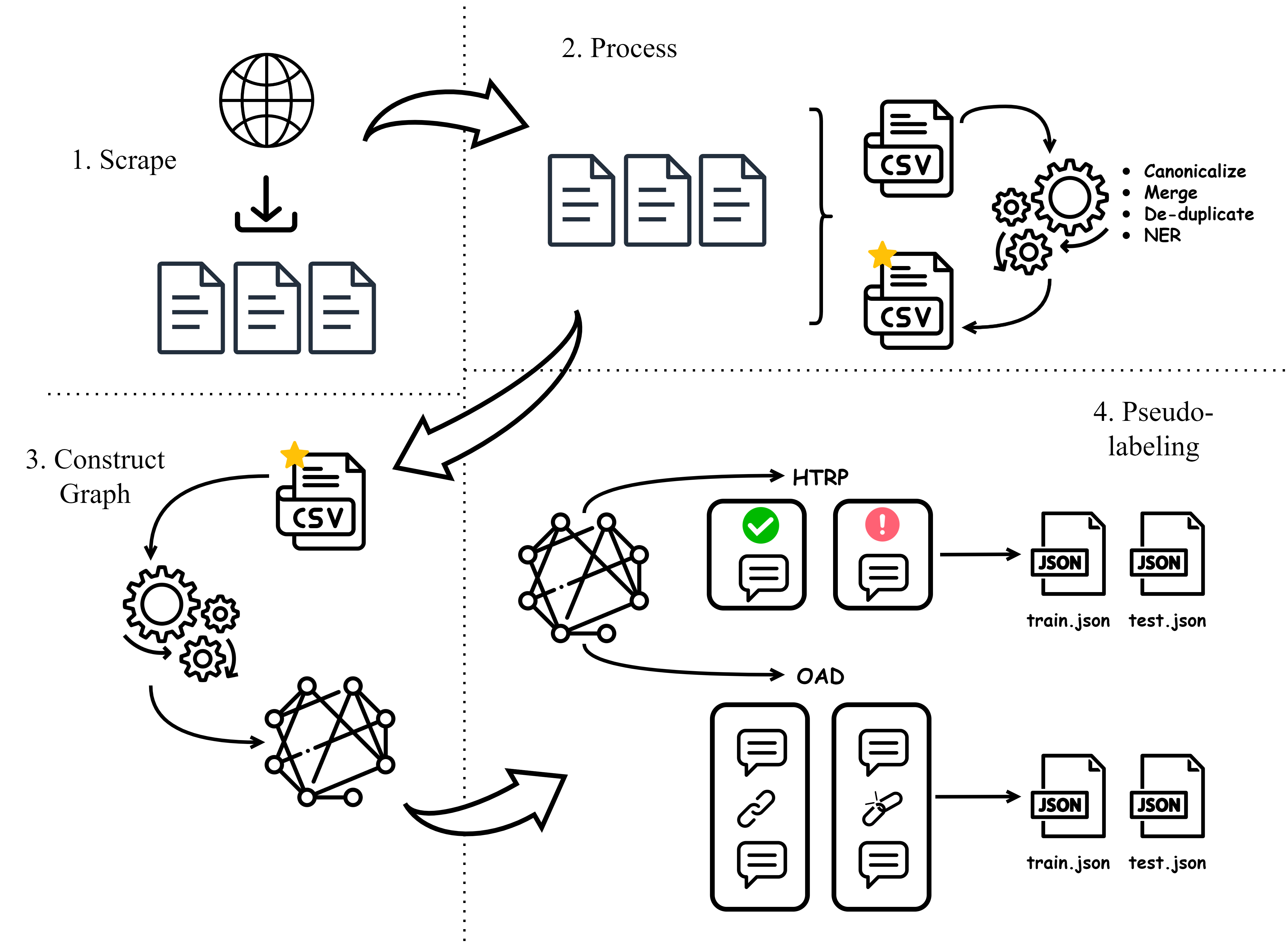}
    \caption{Methodology to generate a pseudo-labeled dataset in HTRP and OAD tasks.}
    \label{fig:datamethod}
\end{figure}

To that end, we proceeded as depicted in Figure \ref{fig:datamethod}. First, raw data is scraped from online escort website(s). The outcome is a collection of documents, one per ad, including several ad metadata. Then, that collection is processed by normalizing the fields to be used in later stages, consolidating different data sources, aggregating important information of exact textual duplicates, and mining information not found in the metadata. Third, a graph of ads is constructed using shared connectors as edges. The pairwise connections serve as labels for the ads-pairwise OAD task. Then, each connected component in the graph is applied a heuristic scoring to determine if it is significantly exhibiting features of human trafficking activity within its contents. Finally, ads belonging to components identified as HT constitute positive examples in the HTRP task.

\subsection{Data Scraping}

There exist plenty of websites advertising sexual services, many of which may be hosting sex trafficking activity in the background. In principle, our proposal is to collect as much data as possible, possibly from several websites. Commonly, there are several fields available for direct extraction via HTML parsing or requests to the website's public API. Among those, the following ones are indispensable to our method:

\subsubsection{Post ID} A unique identifier for the post is relevant, among other things, to track re-posting activity. But, at the very least, it serves organizational purposes in later processing steps. A unique identifier is commonly found in the advertisement's URL.

\subsubsection{Post description body} Advertisements come with a body where the author describes the services offered and other important details associated with them. Importantly, the text must be saved by preserving unicode characters like emojis since those have been proven useful for HT activity detection (citation).

\subsubsection{Location} Usually, OEAs are subdivided, either logically or by sub-domains and sub-paths, into geographical areas. For example, skipthegames.com has sub-paths for several greater urban areas, and a search query allows you to filter by several areas nearby that greater area or city. In other cases, structured fields allow the poster to specify location information. In any case, available location information is important to obtain.

\subsubsection{Contact information} If available in specific fields within the web page source HTML, contact information like phone numbers or emails is extracted.

\subsubsection{Posting Date/Time} It is standard for OEA websites to identify how old a post is. This allows us to estimate the posting date. If this were not possible, a good estimate for the posting date is the scraping date since classified websites often make their posts expire within a short period.

\subsubsection{Images} Images play an important role in analyzing the activity in OEAs websites. In our method they serve as a way to connect related posts. In the web page source, there are links that point to the images. In order to save space, we found it useful to identify an image by a deterministic hash of its contents, and such an image is only stored if it was not saved before.

Specifically, we scraped the Skip the Games~(STG) website. STG's traffic has greatly increased after the fall of Backpage since previous Backpage users needed a new platform. As mentioned before, STG organizes its contents by large populated areas identified by the name of the main city in the area. Other smaller cites are covered but in the form of query strings. For example, the URL:
\begin{center}
\texttt{https://skipthegames.com/posts/austin/?area[]=GRK}    
\end{center}
lists the posts in the city of Killeen, TX. Similar results are obtained if one substitutes ``austin'' for identifiers of other large Texas cities, e.g., ``dallas'', ``waco''. The advertisements are organized chronologically and grouped into 10 sub-pages within the search results section.

Out scraping program is written in the Python programming language. It works by navigating to every link in the gallery of results. Once inside a post, the HTML of the post webpage is parsed, and several components are identified and extracted from the webpage body: ad title, posting date, and location. The location field is a free-text input, so it is not very reliable as a well-defined geotag for the ad. We use Selenium\footnote{\url{https://selenium-python.readthedocs.io/index.html}} for automated browser control and the Beautiful Soup\footnote{\url{https://pypi.org/project/beautifulsoup4/}} library to parse HTML content. The software also downloads the images posted by the author. However, following advice of the team of senior investigators, we decided not to use the raw content of the images in any part of our analysis. Instead, we rely on a deterministic hash of an image's content simply as a method to connect related advertisements.

\begin{figure}[h!]
    \centering
    \includegraphics[width=1\linewidth]{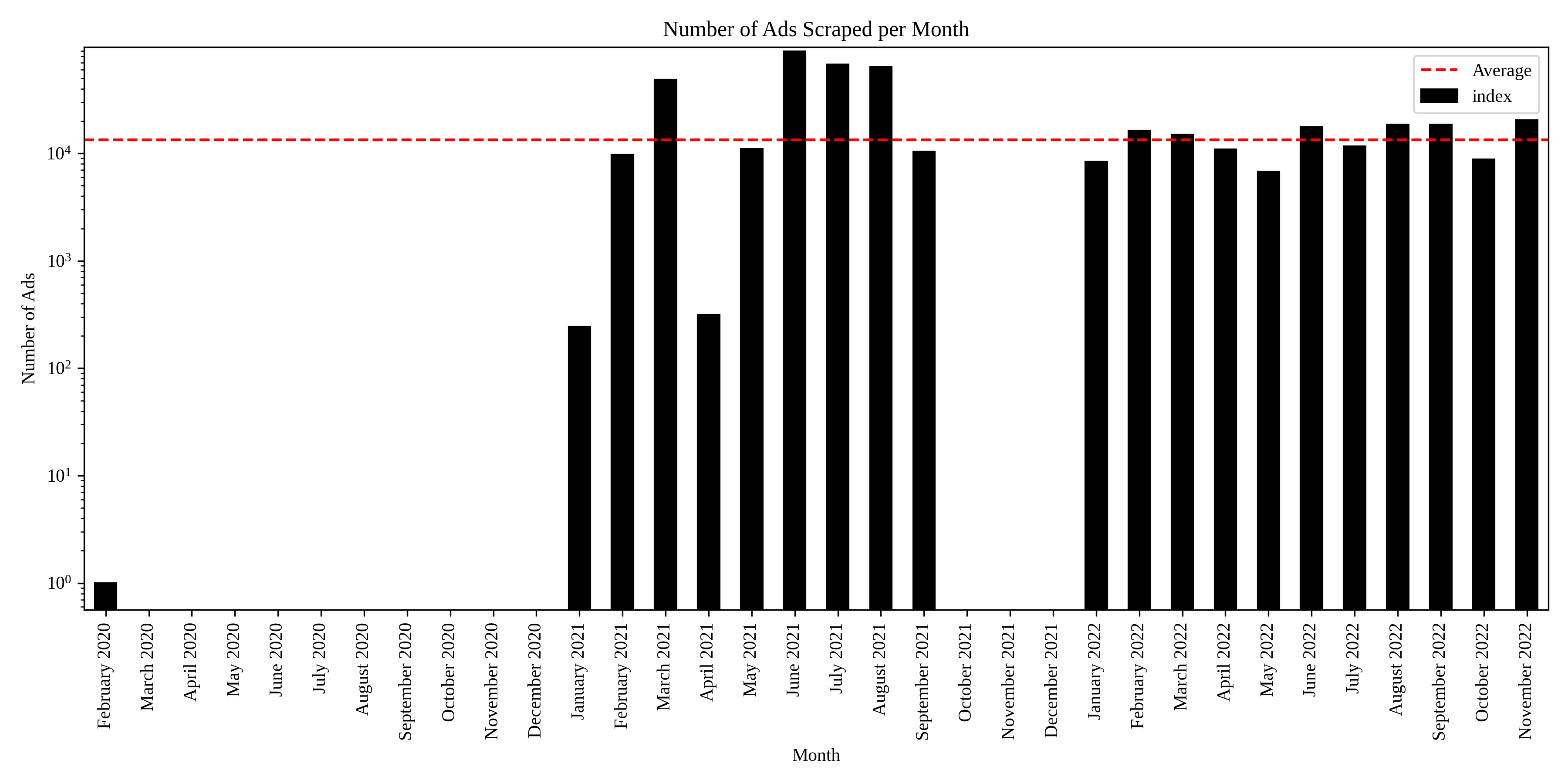}
    \caption{Distribution of scraped advertisements by days. The y-axis is in a logarithmic scale.}
    \label{fig:scrape}
\end{figure}

The scraped process ran from January 2022 to April 2023. A total of 827,606 advertisements were downloaded. Figure \ref{fig:scrape} shows the distribution of the number of ads scraped in the time frame. Observe in the figure the gap between October and December 2021, which separates the data obtained from our collaborators in Deliver Fund and our own scraped data.

\subsection{Data Processing}

Depending on how the scraping process was done, the raw data may need to be cleaned and organized. For example, if raw HTML was downloaded, one may want to remove HTML tags. In data processing pipelines for language processing, it is usual to find steps to standardize the data, e.g., removing unknown words and symbols, lemmatization, and removing stop words. However, in this context, those procedures can make vital information disappear~\shortcite{zhu2019detecting,wiriyakun2022extracting}. For example, gender and number, emojis, and code words are essential in the context of sex trafficking and might get lost on the way if we use classic processing methods.

Another processing step is to unify all the collected data, e.g., data from different websites, and canonicalize it into a single representation. Fields representing the same information from different sources need to be unified. However, it is crucial to ensure the compatibility of types and reformatting to a common format for future processing. Also, it is helpful to add a field to identify the provenance of a record.

In our particular case, we acquired data from STG through our paid collaborators, a non-profit intelligence organization called Deliver Fund~(DF)\footnote{\url{https://deliverfund.org/}} dedicated to equipping law enforcement with tools and training to fight human trafficking. That data amounts to 4,225,643 advertisements scraped from January 2021 to September 2021. DF collected different fields, but there were common ones like posting date, location information, title, and description body. Importantly, that data was missing the hashes of the pictures in the posts. Our collection is missing the URL region in an initial subset of the data because it was not collected until the scraper was updated to do so. Collectively, the two datasets add up to 5,053,249 posts.

Considering the advertisement's description strings, we computed an approximate rate of 90\% exact textual duplicates. This is a problem since we anticipate this data to be fed to NLP models, which may be affected by such a bias. To prevent such unwanted bias, duplicate records need to be identified and collapsed. An exact match in the advertisements' bodies does not necessarily imply those two records are identical. That's why other relevant fields corresponding to duplicate records are aggregated. For example, the posting dates are preserved for all duplicates.

After de-duplication and dropping records that did not have a post description text, our collection is reduced to 515,865 unique ads, considering the description text as the basis of uniqueness.

\subsection{Named Entity Recognition}

As part of our data processing pipeline, extracting information not always present in the metadata is necessary, such as personal contact information and service location(s). The purpose of extracting such info is two-fold:
\begin{enumerate*}[label=(\roman*)]
    \item to serve as connectors for underlying structure discovery and
    \item to be utilized as input to the heuristics we propose for pseudo-labeling ads
\end{enumerate*}.

Even though NER is a well-studied problem and existing models achieve high performance on standard benchmarks, applying them in this context is infeasible for two main reasons. First, the entity types those models are prepared to extract do not necessarily match the ones we are interested in. Second, those models are not robust against the adversarial nature of the text in OEAs.

Some work has been done on information extraction in this kind of adversarial context. \shortciteA{kejriwal2017information} developed a lightweight Information Extraction~(IE) system consisting of two components: a high recall recognizer for the target entities and a classifier for refining the annotations. Their system relies upon a few manually annotated examples for each entity type. On the other hand, \shortciteA{chambers2019character} focused specifically on the problem of phone numbers. These need to be paid special attention since advertisement writers intend to corrupt the textual representation of the number to avoid detection by automated software. For example, the number (254) 123-4567 could be presented as ``2five4onetwothree4567''. The authors created an adversarial dataset of phone numbers and developed RNN-based models to solve the
problem. Interestingly, under the assumption that the task is visually easier for humans, for we are able to interpret visual similarities, they leveraged CNN-based models to process unseen unicode characters. Additionally, \shortciteA{kapoor2017using} use an integer programming formulation to solve the Geo-tagging problem of advertisements. Geo-tagging refers to assigning markers of identifiable locations in the world to a specific entity, in this case a post. Finally, \shortciteA{li2022extracting} propose a combination of rule-based and dictionary-based extraction techniques with a contextualized language model.

\subsubsection{Dataset for NER task} We resourced to supervised learning to address the information extraction problem. To that end, our team of research collaborators labeled a dataset of advertisements using a tool called Doccano,\footnote{\url{https://doccano.github.io/doccano}} an open-source data labeling tool. In Doccano, users can label datasets for multiple tasks, including text classification, image classification, image segmentation, and the one that serves our purposes, sequence labeling. Doccano can be set-up in different ways. We deployed it on an Amazon Web Services' t2-small\footnote{\url{https://aws.amazon.com/ec2/instance-types}} server. In total 1,810 posts were annotated by highlighting the entities and their types. The dataset was split approximately in an 85/15 ratio, for train and test, respectively. Table \ref{tab:ner_ann} shows the counts per entity type class and split of the dataset.

\begin{table}[H]
    \centering
    \caption{Counts of entities grouped per type and split of the NER dataset.}
    \begin{tabular}{|c|c|c|}
         \hline
         & \multicolumn{2}{c|}{\textbf{Number of instances}} \\
         \hline
         \textbf{Entity type} & \textbf{Training set} & \textbf{Test set} \\
         \hline
         Phone Number & 1088 & 182\\
         Name/Nickname & 604 & 105 \\
         Location & 474 & 80\\
         Onlyfans & 144 & 24 \\
         Snapchat & 110 & 22\\
         Username (Other) & 85 & 17\\
         Instagram & 68 & 15\\
         Twitter & 40 & 7 \\
         Email & 18 & 3\\
         Cashapp & 13 & 2\\
         Pornhub & 5 & 1\\
         Venmo & 2 & -\\
         Payment (Other) & 1 & - \\
         \hline
         \textbf{Total} & \textbf{2652} & \textbf{467}\\
         \hline
    \end{tabular}
    
    \label{tab:ner_ann}
\end{table}

 In our split procedure, we intentionally tried to ensure undersampled classes were represented in both sets unless impossible. For example, there is only one post with a sample from the \textit{Payment (Other)} class. In that case, we assigned it to the training set. In the case of \textit{Venmo}, it must have been the case that the same post that has \textit{Payment (Other)} also had an instance of \textit{Venmo}, in which case the remaining instance ended up in the training set as well. Also, notice the remarkable imbalance of the classes in the dataset. These reasons motivated us to consider a different set of classes. Namely, the classes \textit{Username (Other)}, \textit{Cashapp}, \textit{Pornhub}, \textit{Venmo}, and \textit{Payment (Other)}, were all consolidated into a single class \textit{Username (Other)}. Table \ref{tab:ner_ann2} shows the distribution of the classes after the merge.

 \begin{table}[H]
    \centering
    \caption{Counts of entities grouped per type and split of the NER dataset after combining some classes.}
    \begin{tabular}{|c|c|c|}
         \hline
         & \multicolumn{2}{c|}{\textbf{Number of instances}} \\
         \hline
         \textbf{Entity type} & \textbf{Training set} & \textbf{Test set} \\
         \hline
         Phone Number & 1088 & 182\\
         Name/Nickname & 604 & 105 \\
         Location & 474 & 80\\
         Onlyfans & 144 & 24 \\
         Snapchat & 110 & 22\\
         Username (Other) & 106 & 20\\
         Instagram & 68 & 15\\
         Twitter & 40 & 7 \\
         Email & 18 & 3\\
         \hline
         \textbf{Total} & \textbf{2652} & \textbf{467}\\
         \hline
    \end{tabular}
    
    \label{tab:ner_ann2}
\end{table}

\subsubsection{Modeling NER} Several Transformer-based models were leveraged for our NER task. The next paragraph briefly describes each one.

\textbf{BERT}, or ``Bidirectional Encoder Representations from Transformers'', is a state-of-the-art natural language processing model that learns contextual word representations by training on massive amounts of text data. It is pre-trained using a masked language modeling~(MLM) objective, where it predicts missing words in sentences, allowing it to capture bidirectional context and produce embeddings that enhance language understanding in downstream applications. In addition to the masked language modeling objective, BERT employs the next sentence prediction~(NSP) task during its pre-training. This task involves predicting whether a sentence will likely follow another sentence in a given text~\shortcite{devlin2019bert}.

\textbf{ALBERT}, or ``A Lite BERT'', is a variation of the BERT model. It aims to reduce the computational resources required for pre-training while maintaining or even improving the model's performance. ALBERT achieves this by implementing parameter-sharing strategies and factorization techniques, resulting in a more efficient and compact model that's suitable for large-scale natural language processing tasks, making it a resource-efficient alternative to traditional BERT models~\shortcite{lan2020albert}.

\textbf{BigBird} is a novel transformer architecture designed to handle long sequences efficiently by introducing a global attention mechanism. Unlike traditional transformers, which employ self-attention across all tokens in a sequence, BigBird uses global attention to connect tokens at larger intervals, reducing computational complexity for lengthy inputs. This approach makes BigBird well-suited for tasks requiring the processing of massive text data, such as document summarization and language modeling~\shortcite{zaheer2020big}.
    
\textbf{CANINE}, or the ``Character-Augmented Neural Information Encoding'' model, is a Transformer-based encoder designed explicitly for processing character sequences without the need for tokenization. It excels in multilingual contexts and offers superior performance compared to similar models. CANINE simplifies the workflow for practitioners by addressing many common engineering challenges in natural language processing tasks~\shortcite{clark2022canine}.
    
\textbf{GPT-2}, short for ``Generative Pre-trained Transformer 2'', is a state-of-the-art natural language processing model developed by OpenAI. It's renowned for its ability to generate coherent and contextually relevant text across various topics. GPT-2 is pre-trained on a massive corpus of text data, enabling it to understand and generate human-like language, making it a valuable tool in various applications, from content generation to language translation~\shortciteA{radford2019language}.

\textbf{Longformer} is a specialized Transformer-based model designed to handle long documents and sequences more efficiently than traditional models. It utilizes a combination of global and local attention mechanisms to capture broad context and fine-grained details while significantly reducing computational complexity. Longformer is particularly beneficial for tasks involving lengthy text, such as document summarization, question-answering on lengthy articles, and other natural language processing tasks requiring extended document understanding~\shortcite{beltagy2020longformer}.

\textbf{RoBERTa}, which stands for ``Robustly optimized BERT approach'', is an improved recipe for training BERT models that involves simple modifications such as training the model longer, with bigger batches, over more data, removing the next sentence prediction objective, training on longer sequences, and dynamically changing the masking pattern applied to the training data~\shortciteA{zhuang2021robustly}.

\textbf{XLNet}, or ``eXtreme Multi-Label Net'', combines the best features of autoregressive and autoencoding language models. It leverages a permutation-based training objective, where it predicts words in a sentence while considering all possible permutations of word orders, allowing it to capture bidirectional context similar to BERT. This approach has led to impressive results on a wide range of NLP tasks and benchmarks, making XLNet a significant advancement in the field of language understanding~\shortcite{yang2019xlnet}.

To compare all Transformer models effectively, we conducted a 10-fold cross-validation process. All models were trained during 15 epochs, with batch size of 1 using the AdamW optimizer~\shortcite{loshchilov2017decoupled}. The training routines and the models and tokenizers implementations are from the Transformers Python library~\shortcite{wolfetal2020transformers}. Models were trained on a machine with a Tesla V100-PCIE-16GB GPU. Table \ref{tab:ner_models} summarizes the models used. From the table, we see the number of parameters and the Huggingface's hub name of the model\footnote{\url{https://huggingface.co/models}}. The table shows that the number of parameters of all the models is within the same order of magnitude, except for the ALBERT model, which is a more lightweight version of BERT.

\begin{table}[]
    \centering
    \caption{Summary of the models used in the NER task.}
    \begin{tabular}{|>{\centering\arraybackslash}p{0.13\columnwidth}|c|>{\centering\arraybackslash}p{0.25\columnwidth}|>{\centering\arraybackslash}p{0.15\columnwidth}|}
        \hline
         \textbf{Model} & \textbf{Transformers Hub name} & \textbf{Reference} & \textbf{Parameters} \\
         \hline
         ALBERT & albert-base-v2 & \shortciteA{lan2020albert} & 11,094,530 \\
         BERT & bert-base-cased & \shortciteA{devlin2019bert} & 107,721,218 \\
         BigBird & google/bigbird-roberta-base & \shortciteA{zaheer2020big} & 127,470,338 \\
         CANINE & google/canine-c & \shortciteA{clark2022canine} & 132,084,482 \\
         GPT-2 & gpt2 & \shortciteA{radford2019language} & 124,441,346 \\
         Longformer & allenai/longformer-base-4096 & \shortciteA{beltagy2020longformer} & 148,070,402 \\
         RoBERTa & roberta-base & \shortciteA{zhuang2021robustly} & 124,056,578 \\
         XLNet & xlnet-base-cased & \shortciteA{yang2019xlnet} & 116,719,874 \\
         \hline
    \end{tabular}
    
    \label{tab:ner_models}
\end{table}

\subsubsection{Tokenizers} We believe the characteristics of the tokenizer employed by each model to be essential in measuring their fit for this context. In the language models' world, the fact that word-based tokenization is not satisfactory enough has become apparent for several reasons. A typical word-based tokenizer would split a sentence based on a set of rules that might include separating by space or punctuation. For example, the sentence \textit{The kid plays with his dog.} is split into the tokens $[\text{\textit{The}, \textit{kid}, \textit{plays}, \textit{with}, \textit{his}, \textit{dog}, \textit{.}}]$. This has two issues. First, not all languages' words are space-separated~(take Japanese and Korean, for instance). Second, with a large enough corpus, this process could result in a massive vocabulary because it accounts for misspellings, gender and number modifications, verbal tense modifications, etc. Additionally, this procedure makes it hard to deal with out-of-vocabulary~(OOV) words effectively, and attempting to do so may result in an even more extensive vocabulary.

Some alternatives to word-based tokenization have become predominant, namely, sub-word-based and character-based tokenization. The main idea behind sub-word-based tokenization is allowing tokens to be below the level of words, i.e., fragments of words or even single characters. This allows the codification of strange words as combinations of sub-word tokens~\shortcite{bostrom2020byte}. Going back to the previous example, and assuming the word \textit{plays} is not part of the vocabulary of the tokenizer, yet the fragments \textit{play} and \textit{s} are, the sentence could be split into the tokens $[\text{\textit{The}, \textit{kid}, \textit{play}, \textit{s}, \textit{with}, \textit{his}, \textit{dog}, \textit{.}}]$

Sub-word-based tokenization algorithms are described in terms of two sub-tasks: training and encoding. Training the tokenizer takes a corpus of text and produces a collection of tokens called vocabulary at the bare minimum, and possibly other items. Encoding is the process by which a trained tokenizer splits a piece of text into tokens in its vocabulary. 

Some sub-word-based tokenization methods stand out, such as Byte-Pair Encoding, Word Piece, and Unigram. In Byte-Pair Encoding~(BPE), the training algorithm starts with a collection of all allowed characters and merges them pairwise to create new tokens. To encode a string, it assumes the string has been pre-tokenized, i.e., divided into words. The training process not only computed the vocabulary but also preserved the order in which the merges occurred. Then, encoding a word only requires reproducing the same merges as during training~\shortcite{sennrich2016neural}.

Word Piece tokenization is similar to BPE in the sense that training starts with a collection of characters and proceeds by merging them. The difference is the merging criteria. Whereas BPE only considers the frequency of the resulting pair, Word Piece considers the pairs that maximize the likelihood of the training data when added to the vocabulary. This is the same as finding a pair so that its frequency count divided by the product of the frequency counts of its parts is maximum among all pairs. Encoding a word in Word Piece is done by iteratively finding the longest substring that matches a token in the vocabulary. If, at some point, no remaining substring exists in the vocabulary, special tokens are employed to represent them~\shortcite{schuster2012japanese}.

Finally, Unigram's training process is the converse of BPE and Word Piece. It starts with a vast vocabulary and trims it each iteration until a desired size is reached. Given the current vocabulary, a unigram language model~\footnote{A unigram language model is a model that assigns each token a probability, e.g., by dividing the frequency count of the token by the number of tokens in the training corpus.}, and a loss function defined in terms of the previous, the algorithm discards a certain percentage of the tokens that cause the loss to increase the least when removed from the vocabulary. The process ends with a list of tokens and a probability assigned to each of them. This allows to have a joint probability distribution for each possible tokenization of a word, and it returns the most likely one~(or a randomized one, depending on the use case)~\shortcite{kudo2018subword}.

On top of the sub-word tokenization resides the concept of Sentence Piece tokenization. All methods mentioned before required a pre-tokenization step. Sentence Piece waives that requirement, and whitespaces are considered a regular symbol within the vocabulary. Tokenization can then occur with any of the methods mentioned before. This has advantages like lossless tokenization and end-to-end sub-word segmentation~\shortcite{kudo2018sentencepiece}. In the Transformers Python library, Sentence Piece is always used in conjunction with Unigram tokenization~\shortcite{huggingface2023docs}.

Despite the great success of sub-word-based tokenization, recent work has highlighted limitations~\shortcite{klein2020getting}. Some research has attempted to deviate from that paradigm by proposing more radical approaches in which authors skip tokenization altogether and consider representing the input as a sequence of individual characters~\shortcite{clark2022canine}, bytes~\shortcite{xue2022byt5}, or even learning tokenization as part of the network ~\shortcite{tay2022charformer}.

See Table \ref{tab:tokenizers} for a comparison among the tokenizers employed in the models we used. Notably, the character-based tokenizer of CANINE, although flexible~(no unknows), has a large vocabulary. Conversely, BPE achieves the same with a much smaller number of tokens. Additionally, some models can handle simultaneously very long windows of tokens; hence, no sentences in our dataset are truncated after tokenization.

\begin{table}[]
    \centering
    \caption{Summary of the properties of the tokenizers of each of the models used.}
    \begin{tabular}{|>{\centering\arraybackslash}p{0.13\textwidth}|c|>{\centering\arraybackslash}p{0.12\textwidth}|>{\centering\arraybackslash}p{0.11\textwidth}|>{\centering\arraybackslash}p{0.15\textwidth}|>{\centering\arraybackslash}p{0.15\textwidth}|}
    \hline
        \textbf{Model} & \textbf{Tokenizer} & \textbf{Vocab. size} & \textbf{Max. context} & \textbf{Truncated Sentences} & \textbf{Average Unknowns} \\
        \hline
        ALBERT & Sent. Piece & 30,000 & 512 & 19 & 85.54\\
        BERT & Word Piece & 28,996 & 512 & 31 & 91.87\\
        BigBird & Sent. Piece & 50,358 & 4096 & 0 & 156.43\\
        CANINE & Characters & 1,114,112 & 2,048 & 17 & 0\\
        GPT-2 & BPE & 50,257 & 1,024 & 11 & 0\\
        Longformer & BPE & 50,265 & 4,096 & 0 & 0\\
        RoBERTa & BPE & 50,265 & 512 & 51 & 0\\
        XLNet & Sent. Piece & 32,000 & $\infty$ & 0 & 71.84\\
        \hline
    \end{tabular}
    
    \label{tab:tokenizers}
\end{table}

It is important to emphasize that pre-trained models are tightly bound to the tokenizer used during pre-training. Consequently, despite considering some tokenizer more advantageous in our context, it is not feasible to plug in a tokenizer of our choice for fine-tuning in the NER task.

\subsubsection{Results} The NER task is usually defined as a token classification problem in which each token from the input sequence is assigned a class that contains information about the entity type the token belongs to and the location of that token within the entity. For example, say there is an entity \textit{Alejandro} of type \textit{Name}. Also, assume the tokenizer splits the word \textit{Alejandro} into the tokens $[\text{\textit{Ale}, \textit{jandro}}]$. Then, those tokens are labeled \textit{B-NAME} and \textit{I-NAME}, respectively. The $B$ and $I$ prefixes indicate that the tokens are in the beginning and inside the entity, respectively, while the suffix \textit{NAME} designates the entity type. Then, given a tokenized input sentence, once a model predicts a class for each token, a decoding procedure takes the sequence of predictions and produces a list of the entities identified.

To compare different models, we evaluate them in the decoded NER output. That is, we do not report scores in the token classification task but in the final entities predicted after the model output is decoded into a collection of entities. The decoding implementation comes from the Transformers library pipeline utility, and we set the aggregation strategy to simple, which empirically produced better results. Additionally, the decoding process returns a score in the range $[0,1]$ for each extracted entity that characterizes the model's confidence in that particular entity. For improved precision, we only considered entities with a prediction score greater than $0.9$.

For evaluation, we compute a modified $F_1$ score that accounts for which correct, partial, missing, incorrect, and spurious matches~\shortcite{piad2020overview}. Each entity produced by the overall pipeline comprises a text span~(start and end indexes) and an entity type. And, for each sample piece of text, a set of gold entities exist that establish the ground truth for that input string. Given these, we can define each type of match.

Correct matches are those in which a predicted entity matches both in the text span and in the entity type with one in the golden set. Incorrect matches are those that exactly match their text span but have a different entity type. Partial matches count predicted entities that partially overlap with an entity in the golden set of the same entity type. Missing matches are entities in the golden set that were not predicted. Finally, spurious matches are any model prediction that was not in the golden set. For fair scoring, any entity in the predicted or golden set is counted only once, and matches are computed in the order correct, incorrect, partial, missing, and spurious.

Given these counts, we compute precision~(Prec) and recall~(Rec) as follows:
\begin{equation}
    \text{Prec}=\frac{C+\alpha P}{C+I+P+M}
\end{equation}
\begin{equation}
    \text{Rec}=\frac{C+\alpha P}{C+I+P+S},
\end{equation}
where $C,I,P,M,\text{ and }S$ represent correct, incorrect, partial, missing, and spurious matches, respectively. The coefficient $0\leq\alpha< 1$ controls how much weight is given to a partial match. In our experiments, we set $\alpha=0.5$. The $F_1$ score is then computed as usual:
\begin{equation}
    F_1=2\cdot\frac{\text{Prec}\cdot \text{Rec}}{\text{Prec}+\text{Rec}}\text{ .}
\end{equation}

Table \ref{tab:ner_results} shows each model's precision, recall, and $F_1$ score during training and validation. Each item is obtained by averaging the results from the 10 folds for the best-performing checkpoint in each fold according to that fold's evaluation set.

\begin{table}[]
    \centering
    \caption{Average results of the 10-fold cross-validation process. Best entries are highlighted in bold. The standard deviation across folds is shown in parentheses.}
    \begin{subtable}{.9\columnwidth}
       \centering
       \begin{tabular}{|c|c|c|c|}
        \hline
         \textbf{Model} &  \textbf{Precision} & \textbf{Recall} & \textbf{$\mathbf{F_1}$ score}\\
         \hline
            ALBERT & 0.98~(0.006) & 0.93~(0.009) & 0.954~(0.007) \\
            BERT & 0.932~(0.002) & 0.866~(0.007) & 0.898~(0.003) \\
            Big Bird & 0.758~(0.006) & 0.701~(0.037) & 0.728~(0.023) \\
            Canine & 0.985~(0.004) & 0.917~(0.029) & 0.95~(0.018) \\
            GPT-2 & 0.736~(0.005) & 0.7~(0.017) & 0.718~(0.011) \\
            Longformer & 0.979~(0.007) & 0.956~(0.011) & 0.967~(0.009) \\   
            RoBERTa & 0.984~(0.004) & 0.937~(0.004) & 0.96~(0.004) \\
            XLNet & 0.986~(0.003) & 0.951~(0.003) & 0.969~(0.002) \\
         \hline
    \end{tabular}
    \caption{Training}
    \label{tab:ner_results_train}
   \end{subtable}
   \begin{subtable}{.9\columnwidth}
       \centering
       \begin{tabular}{|c|c|c|c|}
        \hline
         \textbf{Model} &  \textbf{Precision} & \textbf{Recall} & \textbf{$\mathbf{F_1}$ score}\\
         \hline
            ALBERT & 0.859~(0.031) & 0.799~(0.028) & 0.827~(0.022) \\
            BERT & 0.825~(0.031) & 0.705~(0.046) & 0.759~(0.031) \\
            Big Bird & 0.684~(0.029) & 0.582~(0.032) & 0.629~(0.029) \\
            Canine & 0.852~(0.044) & 0.743~(0.055) & 0.792~(0.035) \\
            GPT-2 & 0.62~(0.033) & 0.57~(0.023) & 0.593~(0.017) \\
            Longformer & 0.866~(0.024) & 0.857~(0.034) & 0.862~(0.024) \\
            RoBERTa & 0.867~(0.027) & 0.847~(0.041) & 0.856~(0.022) \\
            XLNet & 0.86~(0.02) & 0.841~(0.034) & 0.85~(0.022) \\
         \hline
    \end{tabular}
    \caption{Validation}
    \label{tab:ner_results_val}
   \end{subtable}

      \label{tab:ner_results}
\end{table}

From the table we can see that Longformer and XLNet significantly outperform other baselines, with Longformer being slightly better. Note in Table \ref{tab:tokenizers} how Longformer's and XLNet's tokenizers exhibit some desired properties, like having a context long enough not to need to truncate any sentence in the dataset. Moreover, Longformer's tokenizer did not produce any unknown tokens. Along with RoBERTa's and GPT-2's, Longformer's tokenizer uses byte-level BPE, which starts with an initial vocabulary of all possible bytes. Each individual byte is preserved after training, hence the lack of unknowns.

In Table \ref{tab:ner_results}, we can also observe that the CANINE model is significantly worse than others despite using Unicode-based character tokenization, hence avoiding any unknowns. Notice, however, that encoding the entire Unicode alphabet results in a very large vocabulary. We sense this must contribute significantly to its failure in our context. Additionally, notice that GPT-2's uses a tokenization process similar to RoBERTa's and needs to truncate even fewer sentences during training due to larger context, yet it performs significantly worse. Note also that GPT-2 is an auto-regressive transformer, i.e., it only considers previous inputs to predict the subsequent output. We consider this a limitation in our context because NER is a highly bi-directional task in which evidence found later in the input could contribute to classifying the current token.

After the validation process described in the subsection \textit{Modeling NER}, we selected the Longformer architecture and retrained a model with the full training and validation data. The resulting model was evaluated in a held-out testing set. The overall results in terms of precision, recall, and $F_1$ score are shown in Table \ref{tab:ner_results_test}. The performance was also measured in the individual classes. The overall performance is reported in the table as the micro average of the per-class results.

\begin{table}[]
    \centering
    \caption{Testing overall and per-class performance of the Longformer model trained will full training and validation data.}
    \begin{tabular}{|c|c|c|c|}
         \hline
         \textbf{Class} & \textbf{Precision} & \textbf{Recall} & \textbf{$\mathbf{F_1}$ score} \\
         \hline
            Phone Number & 0.941 & 0.973 & 0.957 \\
            Name/Nickname & 0.775 & 0.738 & 0.756 \\
            Location & 0.75 & 0.708 & 0.728 \\
            Onlyfans & 0.85 & 0.708 & 0.773 \\
            Snapchat & 0.762 & 0.727 & 0.744 \\
            Username (Other) & 0.429 & 0.3 & 0.353 \\
            Instagram & 0.667 & 0.667 & 0.667 \\
            Twitter & 1.0 & 1.0 & 1.0 \\
            Email & 0.4 & 0.667 & 0.5 \\
            \hline
            \textbf{Overall (Micro)} & \textbf{0.827} & \textbf{0.804} & \textbf{0.815} \\
        \hline
    \end{tabular}

    \label{tab:ner_results_test}
\end{table}

From Table \ref{tab:ner_results_test}, notice the overall performance is below the estimation in the validation set. Also, the classes \textit{Username (Other)}, \textit{Instagram}, and \textit{Email} perform really poorly. Conversely, the model performs above average when predicting the classes \textit{Phone Number} and \textit{Twitter}. Note additionally how the overall precision value is above the recall value for the selected threshold, which is a desired property in our context in order to create meaningful connections between ads.

\begin{figure}
    \centering
    \includegraphics[width=\textwidth]{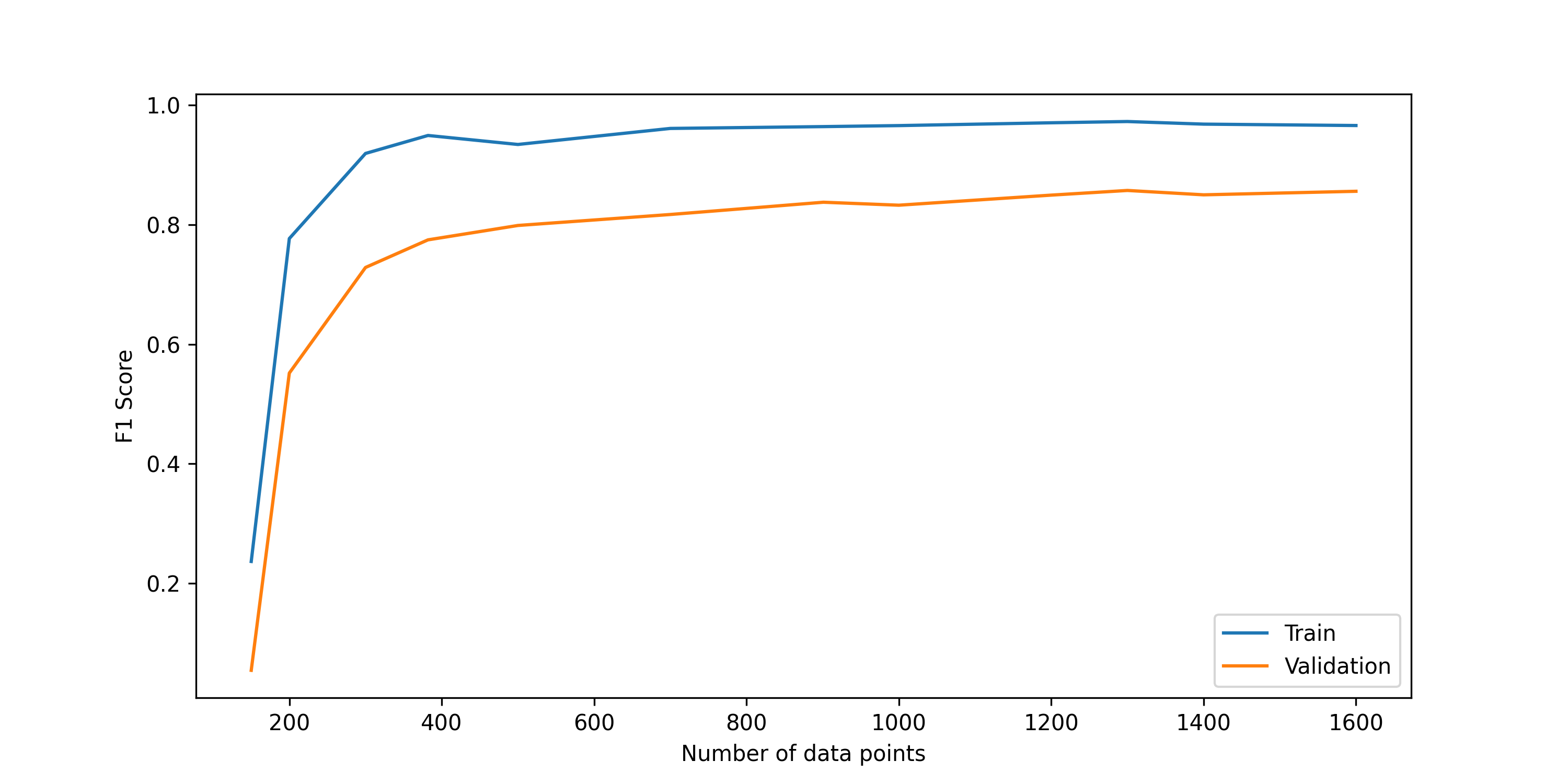}
    \caption{Learning curves of a fine-tuned Longformer model. A data point in this plot coincides with a post of the dataset, not a named entity.}
    \label{fig:learning_curve}
\end{figure}

Finally, observe in Figure \ref{fig:learning_curve} the learning curves computed for the Longformer model. The flat behavior of the curve by the end suggests that no significant performance improvement would be achieved by increasing the number of data points in our training dataset.

\subsubsection{Post-processing} The NER extraction pipeline, although powerful, is not enough for our purposes as it is only able, to a good extent, to identify fragments of text that constitute a relevant entity. In order to compare two of those identified fragments for equality, they must first be normalized to a canonical representation, and those that cannot be canonicalized must be discarded. For all entities, we strip whitespace and lowercase. In the specific case of phone numbers, only US phone numbers are allowed. Additionally, spelled-out digits are converted to their digit representation. For example, the string ``2five4onetwothree4567'' would be transformed into ``2541234567''. Finally, all the phone numbers are mapped to the E.164 format~\footnote{\url{https://www.itu.int/rec/T-REC-E.164/}}~(e.g. +10123456789). For emails, we dispense with the NER pipeline and use a regular expression to detect them instead, which yields a much better performance~(0.83 $F_1$ score).

\begin{figure}
    \centering
    \includegraphics[width=.95\textwidth]{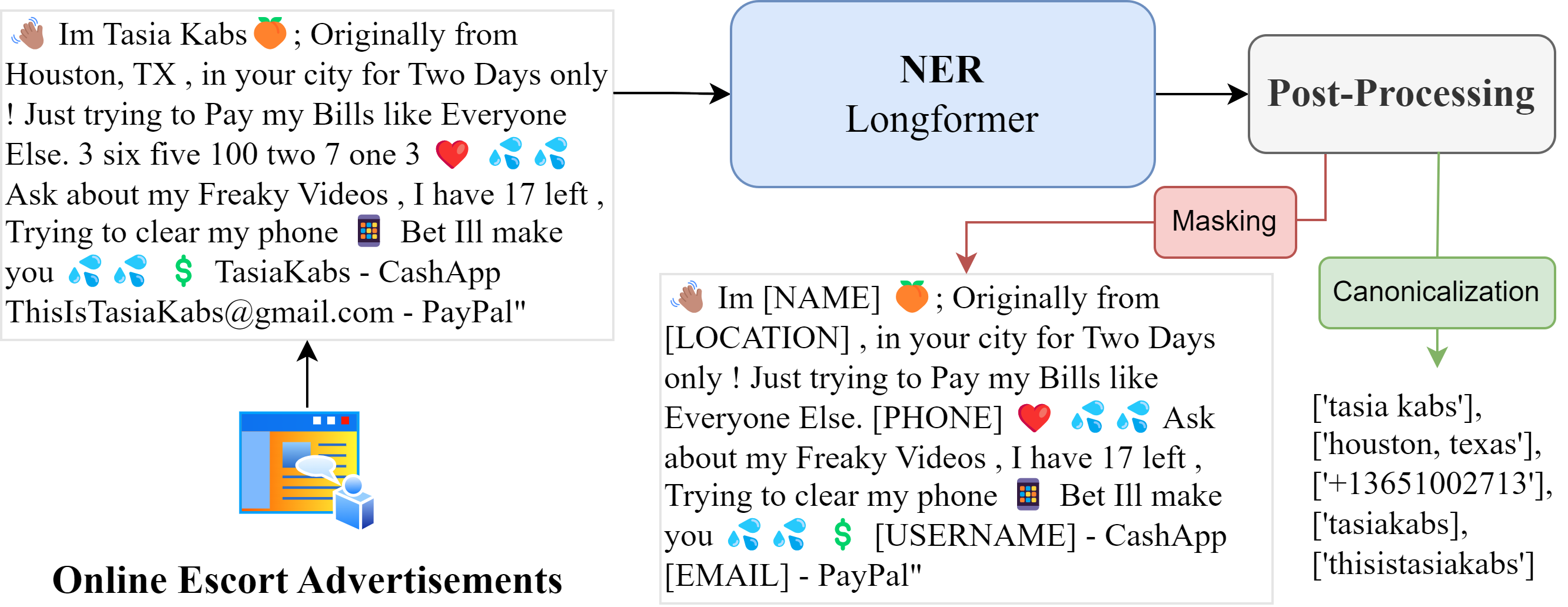}
    \caption{Processing an advertisement description text with the NER pipeline.}
    \label{fig:ner_overall}
\end{figure}

Apart from storing the information of the extracted entities in the record of each OEA, the entities serve the purpose of anonymizing the post's description text. We do so by masking occurrences of the extracted unnormalized fragments. The masked elements are replaced with special tokens. This process prevents the models from focusing their attention on specific values that may prevail in the training dataset and instead forces them to understand the input text in the absence of identifiable information. Figure \ref{fig:ner_overall} summarizes the NER processing of an ad's description text and illustrates the outcome.

This concludes the explanation of the data processing steps we conduct. Next, we proceed to elaborate on how the processed data is used to create a labeled dataset.

\subsection{Relatedness Graph}

The first step to produce the pseudo-labeled tasks is to construct what we call a Relatedness Graph. Each disambiguated post\footnote{Disambiguated here means that the post description text is unique, as described in the de-duplication step.} corresponds to a node in the graph. An edge connects two posts if they share common images or hard identifiers. In our case, those hard identifiers are:

\begin{itemize}
    \item phone number,
    \item email, and
    \item social media handle
\end{itemize}.

Our selection criteria were grounded on the literature reviewed in Chapter \ref{chpt:slr}. Recalling our findings, phone numbers were identified as a crucial connector between related advertisements. Email accounts were also widely used, and it is relatively easy to extract them using a regular expression, as long as they are not somehow coded. Despite not many papers citing social media handles as hard identifiers, we consider them paramount. We attribute their relatively low presence in the literature to the difficulty of extracting them from the text.

On the other hand, we decided to dispense with other potential hard identifiers like names that were not used because of the low performance of the NER system on those entity types. A text similarity-based method was also impractical in our case since we aim to train models that predict connections based on the input texts' shared features; thus, we discard texts that are too similar from the dataset.

As social media handles, we utilized the usernames of Onlyfans, Snapchat, and Twitter platforms. Instagram was ignored due to low performance in the NER task. By the same token, we did not use the Username (Other) class.

The Relatedness Graph has some interesting characteristics we explore. First, it is a very sparse graph, with 515,865 nodes and 2,206,198 edges~(out of roughly $10^{11}$ possible). It is separated into multiple connected components of diverse sizes. The distribution of the sizes in the components is extremely skewed, as shown in Table \ref{tab:cc_size}, which means the majority of the ads bear no relation to others in terms of our heuristics.

\begin{table}[]
    \centering
    \caption{Size of the connected components in the Relatedness Graph.}
    \begin{tabular}{|c|c|}
        \hline
         \textbf{Size range} &  \textbf{Components} \\
         \hline
            1 node & 184877\\
            2-10 nodes & 51117\\
            10-100 nodes & 5928\\
            100-1000 nodes & 80\\
            1000+ nodes & 1\\
            \hline
            \textbf{Total} & \textbf{287,192} \\ 
        \hline
    \end{tabular}
    
    \label{tab:cc_size}
\end{table}

\begin{figure}
    \centering
    \includegraphics[width=\textwidth]{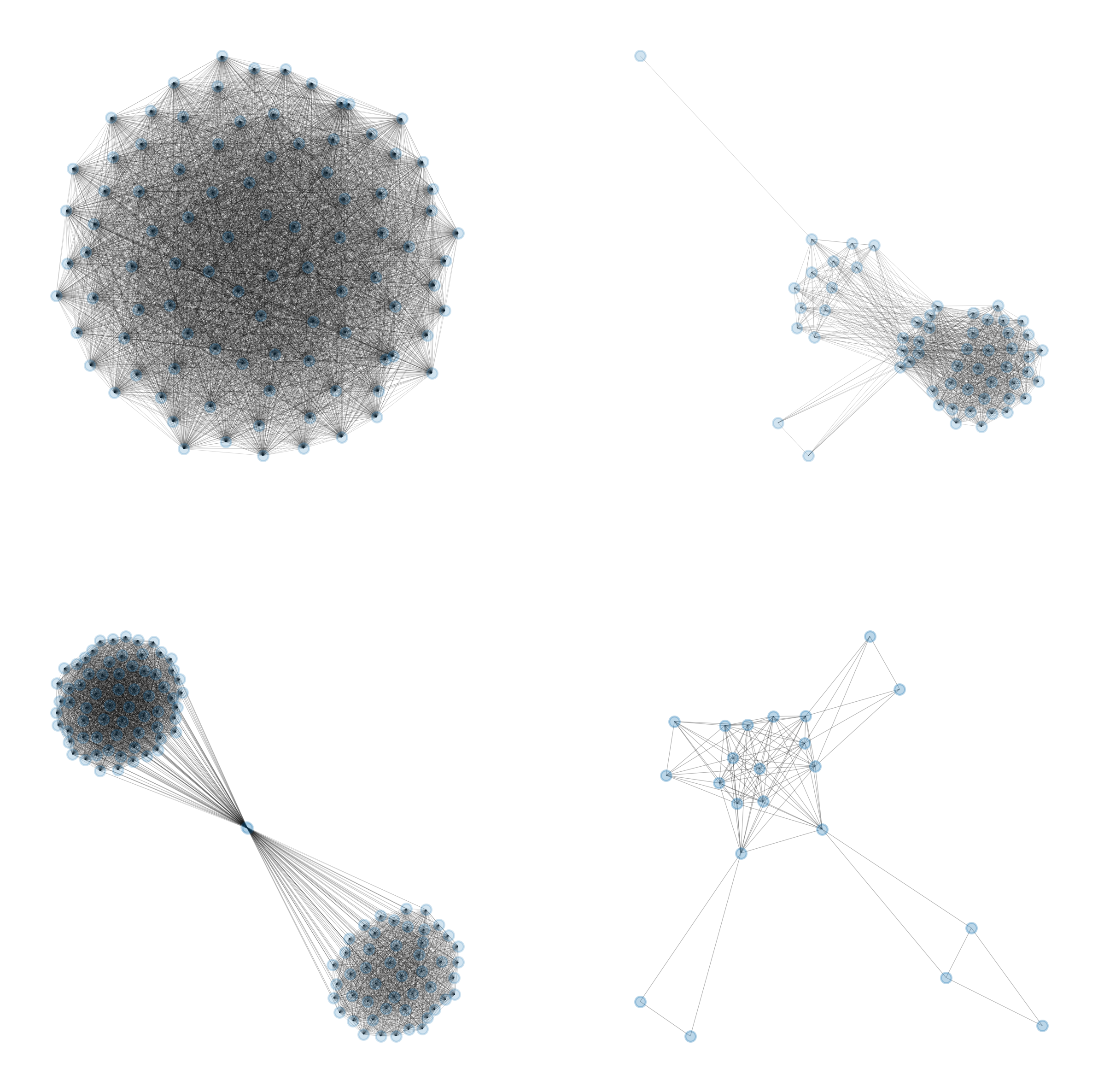}
    \caption{Sample of connected components of the Relatedness Graph.}
    \label{fig:cc_sample}
\end{figure}

As observed, most of the graph is made up of isolated nodes. Interestingly, 7,751 components have nodes with mixed provenance, i.e., from our collaborators in DeliverFund and our own scraped data. The behavior of the components varies. Figure \ref{fig:cc_sample} presents a sample of the structure in some of them. There, we observe fully connected ~(upper left), non-fully connected~(upper right), funnel-like~(lower left), and star-like~(lower right) structures.

\begin{figure}[H]
    \centering
    \begin{subfigure}[b]{.49\textwidth}
        \includegraphics[width=\textwidth]{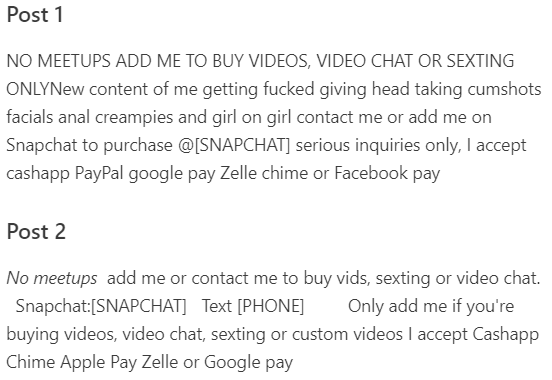}
        \caption{}
        \label{fig:cc_power_text}
    \end{subfigure}
    \hfill
    \begin{subfigure}[b]{.49\textwidth}
        \includegraphics[width=\textwidth]{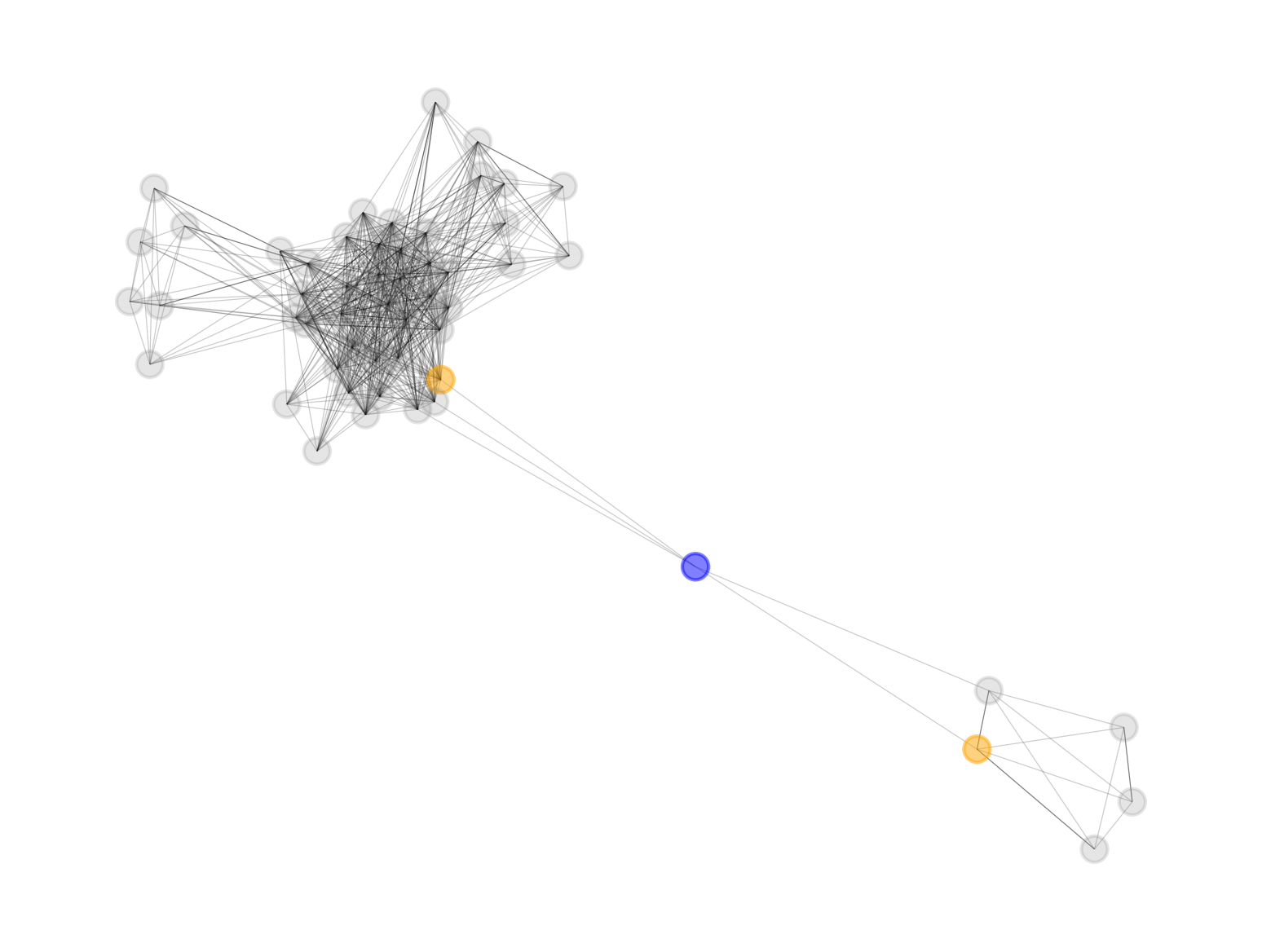}
        \caption{}
        \label{fig:cc_power_graph}
    \end{subfigure}
    \caption{Ads connected indirectly in a connected component of the Relatedness Graph. (a) Description text of the highlighted posts. (b) Connected component graph. The referred posts are highlighted in red.}
    \label{fig:cc_power}
\end{figure}

To understand the power of the Relatedness Graph, observe the connected component depicted in Figure \ref{fig:cc_power}. As noted, the posts highlighted in orange do not share an edge. However, they are connected indirectly through the blue node, which allows grouping all the identifiers recognized in either post~(two different Snapchat accounts and a phone number) and associating them with the same underlying activity, criminal or not.

\subsection{Pseudo-labeling}

The final artifacts of the dataset are the pseudo-labels for the tasks:

\begin{itemize}
    \item Organized Activity Detection~(OAD)
    \item Human Trafficking Risk Prediction~(HTRP)
\end{itemize}

We designed both OAD and HTRP as binary classification tasks. In both cases, the Relatedness Graph is first separated into connected components. Those components are randomly split into training and test sets by trying to get 80\% of advertisements to go into the training set and 20\% into the test set. The splitting process is not straightforward because the distribution of the components' sizes is not uniform. This train/test split method ensures all connected ads are either in the training or test set, not both, thus avoding biasing the evaluation process by sharing similar ads across dataset splits.

In the OAD task, each instance consists of a pair of advertisements, and the binary labels are determined by the edges of the Relatedness Graph. Each pair of ads with an edge that links them constitutes a positive sample. The complement of those pairs comprises the negatively labeled examples. In this context, there are two aspects we found worth highlighting. First, the domain from which negative samples can be drawn, i.e., the complement of the edges, is most likely several orders of magnitude larger than the positive samples because the Relatedness Graph is sparse. Considering that balanced datasets are Machine Learning-friendly, one must pick a subset of the negative samples of a sensible size. In our case, we randomly sub-sampled a negative set to match the size of the positive set of samples. Second, despite exact duplicates being discarded in previous steps, similar ads still abound within connected components. Since a similarity-based task~(like OAD) would make little sense if the texts are clearly very similar in appearance, it is wise to ignore lookalike examples. To measure similarity, we propose to use the complement of the string edit distance~(Levenshtein distance) normalized by the length of the maximum length input, which can be defined as:
\begin{equation}
\text{similarity}(s_1, s_2) = 1-\frac{\text{Levenshtein}(s_1,s_2)}{\max(|s_1|, |s_2|)},
\end{equation}
where $s_1,s_2$ are the two strings for which we want to establish their similarity.

This way, any pair of ads that are more similar than a given threshold is ignored. Based on a study of similarity we conducted with our research collaborators, we determined that a threshold of 0.5 serves our purposes well.

Creating pseudo-labels for the HTRP task is more sophisticated. Recall our ultimate goal is to find new insights into the use of the language in OEAs that are correlated with HT activity. Therefore, that rules out the possibility of using indicators based on keywords and phrases found in the text. Such labeling would highly bias our results towards finding precisely the language constructions used to label it in the first place. Therefore, we turn to heuristics that rely on information not directly provided in the description of the ads.

Each connected component of the graph is analyzed separately to create the labels. For each component, we aggregate the unique locations and hard identifiers of all the ads in the component. Then, all the ads in the component are labeled positive if and only if:

\begin{itemize}
    \item there are two locations separated by a distance greater than a given threshold, or
    \item the number of hard identifiers of a given category exceeds a given threshold for that category.
\end{itemize}

Concretely, we set the distance threshold to 300 miles in geodesic distance. The threshold is set based on previous studies of human trafficking circuits derived from OEAs data~\shortcite{ibanez2014detection,ibanez2016detecting}. In those papers, identified circuits connect cities as close as Oregon, PO and Seattle, WA (126 mi), the Hawaiian islands (farthest points are 306 mi away), and cities within the state of California with an average pairwise distance slightly over 300 mi~(Los Angeles, San Francisco, and San Diego). The 300-mile threshold was a sensitive choice based on that evidence.

We used the Google Geocoding API~\footnote{\url{https://developers.google.com/maps/documentation/geocoding/overview}} to retrieve the geo-coordinates of each named location. A total of 27,507 location-related strings were queried. The geo-coordinates are used to compute the geodesic distances. We keep the first U.S. location in the list of all the possible results obtained from the API call. This procedure significantly reduces false positives due to the distance heuristic. For example, searching for the string \textit{alexandria} results in a response with a list of different locations because of the under-specified query. The first of those corresponds to the Alexandria Governorate in Egypt, and the second to the city of Alexandria, VA. In this case, we ignore the first element and preserve the U.S. location.

\begin{figure}[H]
    \centering
    \includegraphics[width=0.8\columnwidth]{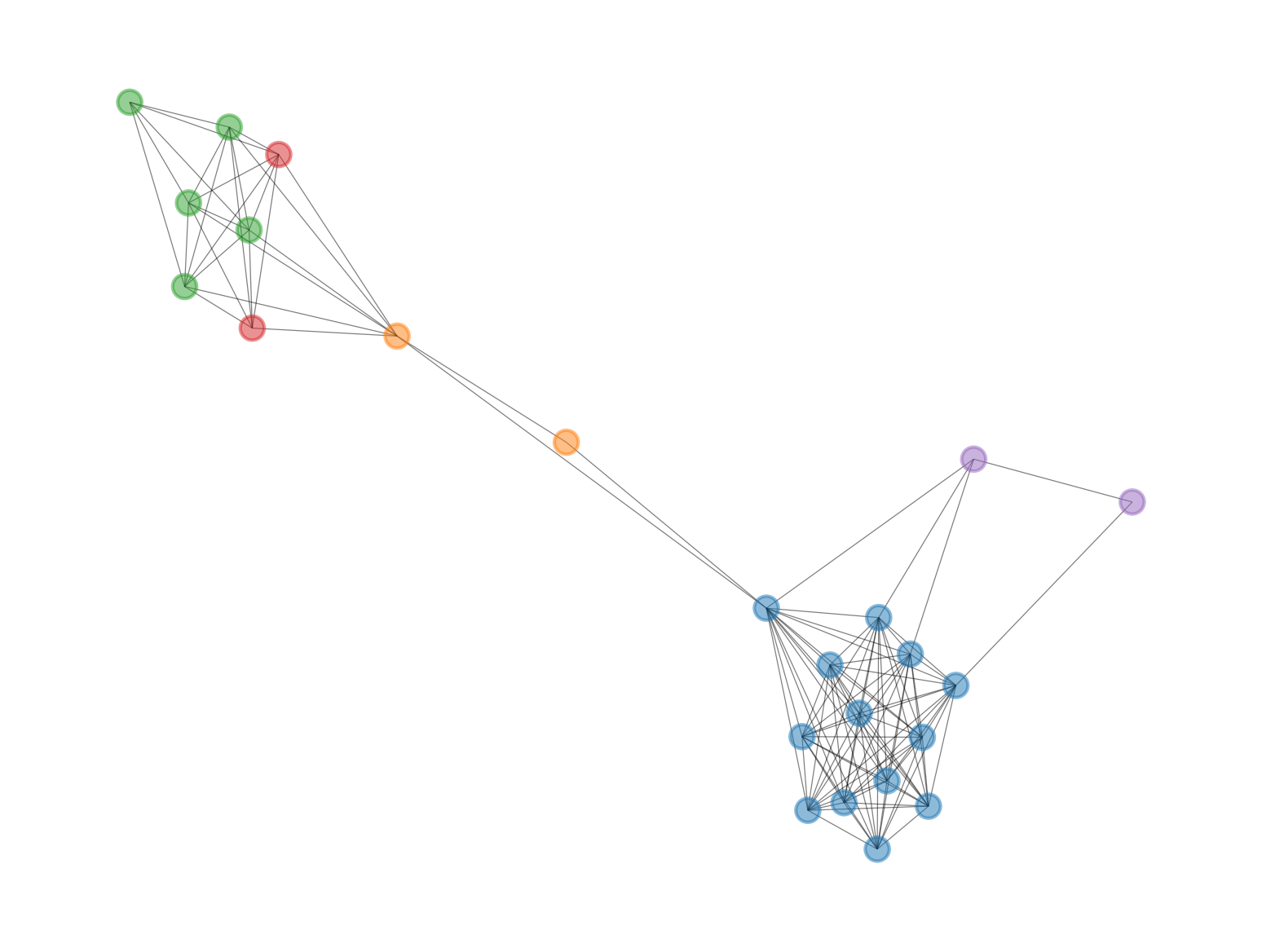}
    \caption{A connected component labeled positive due to several phone numbers found. Each color represents a different phone number encountered.}
    \label{fig:cc_identifiers}
\end{figure}

Regarding the hard identifiers, if a component had 2 or more distinct phone numbers or emails, it was labeled as positive. For the rest of the identifiers, the threshold was set to 3 to account for the lower accuracy of the NER pipeline. Figure \ref{fig:cc_identifiers} shows an example of a connected component labeled positive due to the existence of several distinct phone numbers.

Finally, we applied the same principle as in OAD of avoiding ads with similar description text. We only included an ad, either a positively or negatively labeled one, if it is not similar to any ad included before. The rationale in this case is avoiding feeding the model with similar examples. The similarity threshold used was the same, 0.5.

Table \ref{tab:tasks_splits} summarizes the size of the dataset splits in each task. Notice how the HTRP task's splits are imbalanced, with the positive class accounting only for roughly 16\% of the total. Class imbalance can significantly impact the ability of a classifier to properly learn to separate the classes by biasing it towards predicting correctly the predominant classes only. Appropriate adjustment of the learning process and careful selection of the metrics is thus vital. ~\shortcite{rivas2020deep}.

\begin{table}[]
    \centering
    \caption{Description of the OAD and HTRP tasks datasets splits.}
    \begin{tabular}{|c|c|c|c|c|}
        \hline
          \textbf{Task} & \textbf{Positives} & \textbf{Negatives} & \textbf{Total} & \textbf{Pos. Ratio}\\
          \hline
          \multicolumn{5}{|c|}{\textbf{Train split}} \\
         \hline
         OAD  & 1,154,687 & 1,154,687 & 2,309,374 & 0.5 \\
         HTRP & 41,897 & 227,005 &  268,902 & 0.156\\
         \hline
         \multicolumn{5}{|c|}{\textbf{Test split}} \\
         \hline
         OAD  & 265,964 & 265,964 & 531,928 & 0.5 \\
         HTRP & 11,008 & 57,730 & 68,738 & 0.16 \\
         \hline
    \end{tabular}
    
    \label{tab:tasks_splits}
\end{table}

Additionally, we measured the effect of discarding duplicates due to similarity. The discarded items are counted in terms of the number of edges for the OAD task and the number of nodes for the HTRP task. As observed, approximately $1/3$ of the edges could not make the positive class in the OAD task, and roughly the same share of nodes was not included as a data point in the HTRP task.

\begin{table}[]
    \centering
    \caption{Discarded duplicates due to text similarity in the OAD and HTRP tasks datasets splits.}
    \begin{tabular}{|c|c|c|}
        \hline
          \textbf{Task}  & \textbf{Discarded} &  \textbf{Ratio}\\
          \hline
          \multicolumn{3}{|c|}{\textbf{Train split}} \\
         \hline
         OAD  & 650,092 & 0.36\\
         HTRP & 143,810 & 0.349\\
         \hline
         \multicolumn{3}{|c|}{\textbf{Test split}} \\
         \hline
         OAD  & 135,455 & 0.337\\
         HTRP & 34,415 & 0.334\\
         \hline
    \end{tabular}
    
    \label{tab:tasks_splits}
\end{table}

\begin{figure}
    \centering
    \includegraphics[width=0.65\columnwidth]{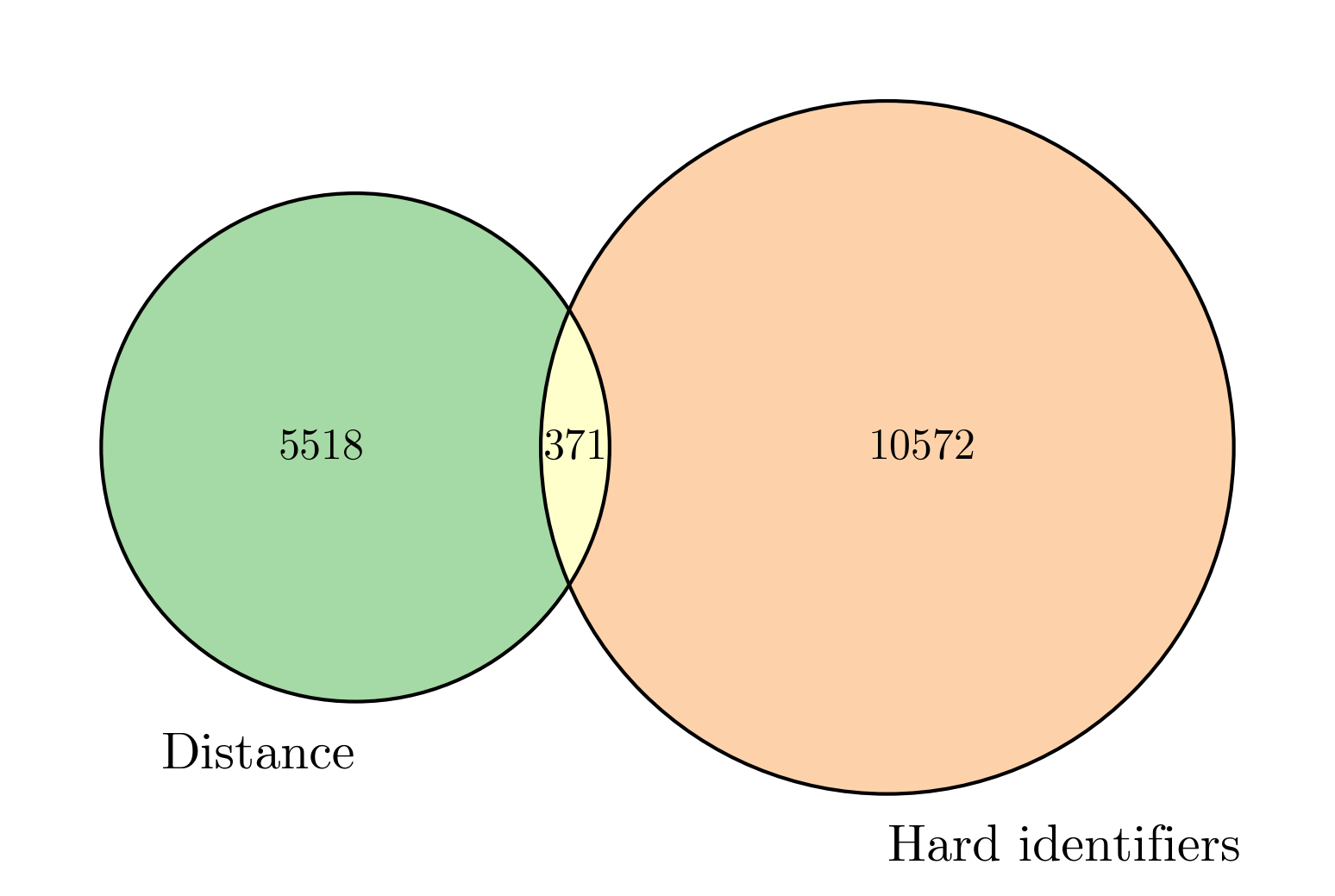}
    \caption{Positive examples of the HTRP task detected by each heuristic.}
    \label{fig:venn_htrp}
\end{figure}

Next, observe in Figure \ref{fig:venn_htrp} the influence of each heuristic applied when pseudo-labeling the HTRP task. We can see from the figure that due to the hard identifiers heuristic, twice as many components are labeled as compared to the distance heuristic. Notably, this could bias model predictions toward advertisements listing hard identifiers in their text description. Additionally, the overlap of both sources is relatively small, which was advantageous because it resulted in more suspected components. However, a plausible interpretation adverse to our method is that the inconsistency between the two may indicate the poor quality of the heuristics as predictors of human trafficking risk.

\section{Limitations}

Throughout the dataset creation process, we identified some limitations of our method we discuss separately in the next paragraphs.

Creating the Relatedness Graph is subject to multiple sources of error. For example, the data is generally sparse, and some records have missing information. Additionally, edges are drawn based on extracted hard identifiers. Therefore, the error in the NER pipeline is propagated to the dataset. Finally, we identified differences in the data coming from different sources. For example, the posts provided by DeliverFund were preprocessed by stripping leading spaces and line breaks, whereas our scraped data was not. Secondly, the DeliverFund data is missing picture hashes, making that subset of the data less connected in the Relatedness Graph due to missing potential links.

When creating the OAD task labels, positive edges between dissimilar posts are used. The choice of similarity metric and threshold impacts the resulting dataset. We chose a sensible similarity metric and threshold value but did not comprehensively evaluate all possible scenarios. This limitation applies to the HTRP task labeling as well. In this context, the way dissimilar posts were selected is greedy and sub-optimal, and it is influenced by the order in which the nodes in the graph are processed.

Moreover, there is noise associated with the geocoding process. Location strings obtained with the NER pipeline are prone to error. Furthermore, some locations are underspecified and ambiguous. Finally, the results are also sensitive to the distance threshold selected. All these are sources of variation and error in the HTRP task due to the distance heuristic. Moreover, manually set thresholds also affect the hard-identifiers heuristic.

Finally, we hypothesized the way we generate the positive and negative labels, particularly in the HTRP task, may induce some bias in the dataset. This may happen because posts with abundant locations or hard identifier mentions are more likely to be connected to other posts, making them more prone to belonging to suspected connected components.

Recall from the \textit{Named Entity Recognition} Section that the description text of an ad is processed to mask occurrences of extracted entities. Consequently, we inspected the prevalence of masked elements within the HTRP task data. Concretely, we counted the number of locations and hard identifiers to contrast the differences between the positive and the negative classes. For a fair comparison, the counts are normalized by the number of samples. Table \ref{tab:bias_mask} shows the results.

\begin{table}[]
    \centering
    \caption{Differences in number of masked elements in the HTRP task data.}
    \begin{tabular}{|c|c|c|c|c|}
        \hline
        & \multicolumn{2}{c|}{\textbf{Train split}} & \multicolumn{2}{c|}{\textbf{Test split}} \\
        \hline
        \textbf{Mask} & Positive & Negative & Positive & Negative \\
        \hline
        Phone Number & 1.003 & 0.442 & 0.969 & 0.43 \\
        Name/Nickname & 0.444 & 0.293 & 0.443 & 0.286 \\
        Location & 0.527 & 0.182 & 0.467 & 0.174 \\
        Onlyfans & 0.091 & 0.019 & 0.093 & 0.022 \\
        Snapchat & 0.133 & 0.022 & 0.097 & 0.021 \\
        Username & 0.037 & 0.013 & 0.033 & 0.013 \\
        Instagram & 0.051 & 0.015 & 0.05 & 0.014 \\
        Twitter & 0.037 & 0.005 & 0.03 & 0.006 \\
        Email & 0.023 & 0.004 & 0.027 & 0.004 \\
        \hline
    \end{tabular}
    
    \label{tab:bias_mask}
\end{table}

We observe differences that could be indicative of bias. A Wilcoxon rank-signed test~\shortcite{wilcoxon1945individual} indeed confirms that there are significant differences between the positive and the negative classes' samples with 99\% confidence. The p-value for the test in both train and test splits is $0.004$. Conversely, running the same test but comparing the train and test splits does not yield enough evidence to reject the null hypothesis under the same confidence~(p-value $0.05$), suggesting train and test distributions are similar as far as masked items are concerned. A paired t-test is less reliable in this scenario due to the non-normality of the data. Nevertheless, we measured it, and the results obtained are similar to those of the Wilcoxon test. These results highlight a limitation of our approach.

\section{Conclusion}

This chapter described a methodology for creating a pseudo-labeled dataset with Organized Activity Detection~(OAD) and Human Trafficking Risk Prediction~(HTRP) binary classification tasks inspired by our findings in the literature. We also showed a particular instance of that method in our case study with data from a sex advertisements website. The core artifact in the methodology is a Relatedness Graph created by connecting posts via hard identifiers extracted from the ad metadata or through a Named Entity Recognition~(NER) pipeline based on state-of-the-art Transformers.

First, the data collection and processing steps are described. Then, a section expands on the NER training and prediction. Finally, a detailed description of the construction process of the Relatedness Graph is provided, as well as the pseudo-labeling process. The edges in the graph constitute positive examples in the OAD task, and heuristics applied to the connected components allow computing labels for the HTRP task.

As part of the data processing step, we developed a NER system to identify different entity types in escort ads. The unique challenges OEAs pose, such as the intentional obfuscation of text to evade automated detection systems, necessitate a robust and adaptive NER system capable of high precision and recall. Our results indicate that the Longformer model, with byte-level BPE tokenization, exhibits superior precision, recall, and $F_1$ score. This model's ability to handle long contexts without truncation and its lack of unknown tokens make it well-suited for the complexities inherent in OEAs.

Overall, the methodology described is relevant to the research community for several reasons. First, the process is well-defined and reproducible. It is also flexible, for instance, to incorporate different data sources, extract and utilize varied information from the ad descriptions, and introduce novel heuristics or tasks for Machine Learning models to learn from. Finally, its application is not tightly coupled to the human trafficking domain. The main steps and rationale described in this chapter could be translated into fighting organized criminal activity in other domains, e.g., drugs and stolen car parts.
%
\chapter{Using Attribution to Discover New Suspicious Language Patterns in Online Escort Advertisements}\label{chpt:attr}

Interpretability methods in the area of Machine Learning help to shed light on the usually unintelligible predictions of complex models like deep neural networks. In the context of human trafficking, understanding the grounds of model predictions acquires even more significance for law enforcement officers who need understandable evidence they can pursue if they intend to build a case against suspected offenders.

Developing models that help predict human trafficking risk in the context of Online Escort Advertisements is a cumbersome task. One of the reasons for this is the lack of reliable, publicly available data, as argued in previous chapters. That is why we turn our focus, instead of reliably predicting human trafficking risk, to finding useful insights in the OEAs language that could represent indicators of human trafficking activity. We do so by training Machine Learning models on tasks conceptually linked to human trafficking with the aim that models learn to associate language patterns with underlying criminal activity—and then inspecting the structures in the language that contribute the most to the model's predictions. Those findings are contrasted with existing knowledge in the field.

This chapter explains our methodology and experimentation. First, we discuss the Transformer architecture in the context of the interpretability of deep learning models. Then, we describe our methodology in concrete steps. Next, the experimental setup is presented, including model training and results, followed by a discussion of the findings and the chapter conclusion.

\section{Preliminaries}

\subsection{Transformers}

The Transformer architecture~\cite{vaswani2017attention} constituted a major breakthrough in Machine Learning, initially in the Natural Language Processing field~\shortcite{devlin2019bert, radford2019language}, and later in Computer Vision as well~\shortcite{dosovitskiy2021image, li2020does}.

The key innovation that distinguishes Transformers from previous architectures is the Self-Attention mechanism. This operation enables the model to weigh the importance of different elements in a sequence when making predictions. Self-attention allows the model to consider the relationship between all elements in the input sequence simultaneously, rather than relying on fixed-size context windows as CNNs~\shortcite{lecun1989backpropagation} or needing to accumulate knowledge of past elements in a sequence into a fixed-size representation as relevant RNN architectures like Long-Short Term Memory~(LSTM)~\shortcite{hochreiter1997long} and Gated-Recurrent Unit~(GRU)~\shortcite{cho2014learning}. This enables Transformers to capture dependencies across long distances, making them highly effective for a wide variety of tasks~\shortcite{li2022survey,khan2022transformers}.

\begin{figure}
    \centering
    \includegraphics[width=.9\textwidth]{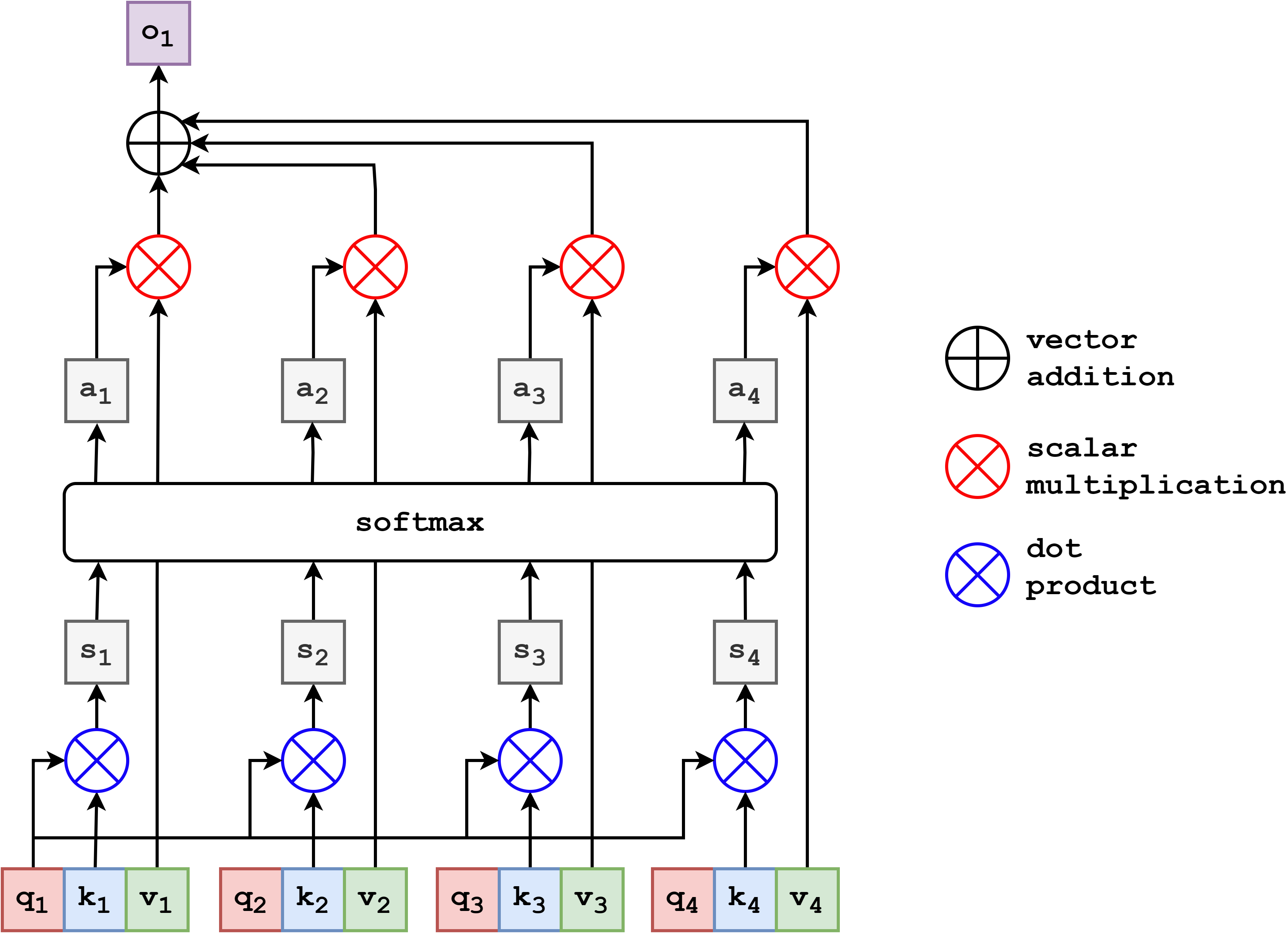}
    \caption{Computation of the scaled dot-product attention for the first position of an input sequence of size 4. Each element in the input sequence is a triplet of a query~($q$), key~($k$), and value~($v$) vectors.}
    \label{fig:attention}
\end{figure}

The Self-Attention operation is a particular case of the Scaled Dot-Product Attention~\cite{vaswani2017attention}. In Scaled Dot-Product Attention, each element in a sequence is represented by three vectors named \textbf{query}, \textbf{key}, and \textbf{value}. The output for a given position is a weighted sum of the value vectors of all the elements in the input, where weights are assigned based on how much the query vector of the given position ``wants to attend'' to each corresponding key vector~\cite{vaswani2017attention}. Figure \ref{fig:attention} depicts this process graphically.

In practice, this operation is computed efficiently for the entire input sequence using the following expression~\cite{vaswani2017attention}:
\begin{equation}
    \text{Attention}(Q,K,V) = \text{softmax}\Big(\frac{QK^T}{\sqrt{d_k}}\Big)V\text{ .}
\end{equation}

Notice that Figure \ref{fig:attention} ignores, for simplicity, the normalization factor $\sqrt{d_k}$ applied to the dot product between the queries and the keys. The value $d_k$ is the size of the key vectors~(which must be the same as the query vectors').

Self-Attention is the special case of Scaled Dot-Product Attention in which the matrices $Q$,$K$, and $V$ come from the same source. In the case of the Transformer, they come from the output of the previous layer at that position. Concretely, the model is parametrized by matrices $W^Q$, $W^K$, and $W^V$ such that if the output of the previous layer was the sequence of vectors represented by matrix $X$ then:
\begin{equation}
    \begin{array}{cc}
         Q & = XW^Q  \text{,}\\
         K & = XW^K \text{,}\\ 
         V & = XW^V \text{ .}\\ 
    \end{array}
\end{equation}

Similarly to CNNs having different filters applied to an input in the same layer, Transformers apply several attention heads to the same input, and concatenate the results at the end. This procedure, which is shown to improve the performance of the Transformer architecture, is called Multi-Headed Attention~\shortcite{vaswani2017attention}.

\subsection{Integrated Gradients Attribution Analysis}

Understanding the way Machine Learning methods make decisions is important from several perspectives, including assessing the reliability of systems, verifying their adherence to ethical and legal standards, and developing deeper knowledge of how models work~\shortcite{zhang2021survey}. Despite the breakthroughs that neural networks have had in recent times, one important deterrent is the lack of explainability of such models in nature~\shortcite{kadra2023breaking}. Several methods have been proposed to overcome this issue, as highlighted by \shortciteA{zhang2021survey} in their survey.

There are different types of interpretability methods. Particularly, attribution methods attempt to quantify the contributions of input features to the output predictions. Two widely known attribution methods are Local Interpretable Model-Agnostic Explanations~(LIME)~\shortcite{ribeiro2016explaining} and SHapley Additive exPlanations~(SHAP)~\shortcite{lunderberg2017unified}. Advantageously, these two methods work for any Machine Learning model. On the other hand, they are computationally expensive if the feature spaces are big and not accurate on noisy data.

A method more tailored to explaining neural networks~(and differentiable models in general) is that of Integrative Gradients~(IG)~\shortcite{sundararajan2017axiomatic}. 
Integrated Gradients is an axiomatic model interpretability technique that assigns significance scores to individual input features. It accomplishes this by estimating the integral of gradients related to the model's output with respect to the inputs. The integral is computed along the straight-line path from a given baseline to the input values. This concept is expanded next.

Having an arbitrary differentiable function $F$, a first take to using gradients for attribution is to define the contribution~(attribution) of a specific feature $x_i$ in an input $x$ to be proportional to both the value $x_i$ and the gradient of $F(x)$ with respect to $x_i$. Indeed, the gradient captures the notion of how a small change in the target feature affect the output value. However, in the Machine Learning context, a well-fit model $F$ is less sensitive to small changes in the input features, making the gradient of $F(x)$ to be nearly zero in a small neighborhood around point $x$.

\begin{figure}
    \centering
    \includegraphics[width=.7\columnwidth]{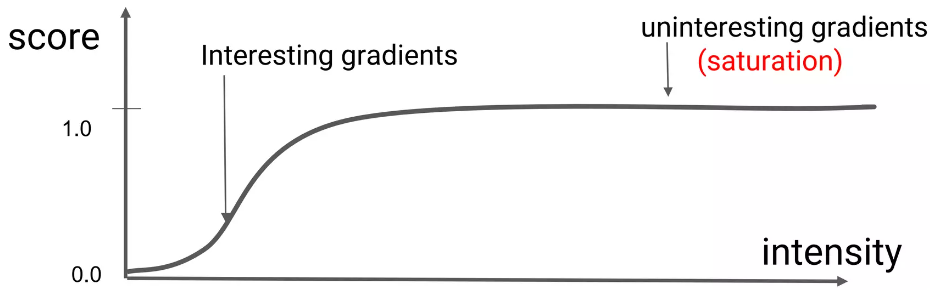}
    \caption{Change in output magnitude~(score) as a feature value goes from a baseline value~(zero intensity), to the full-extent value~(higher intensity)}
    \label{fig:ig_saturation}
\end{figure}

To cope with this apparent issue, IG introduces the concept of a baseline value for the feature. To compute the attribution, the method measures how the contribution of the feature changes as its value moves in a straight-line path from the baseline to the original feature value. An excellent illustration of this phenomenon is shown in Figure \ref{fig:ig_saturation}, which is an adaptation of an image from the tutorial by \shortciteA{gade2019explainable}.

Formally, let $F: \mathrm{R}^{n} \rightarrow[0,1]$ be a function that represents a differentiable model, and let $x\in\mathrm{R}^{n}$ be an input vector, and $x'\in\mathrm{R}^{n}$ be the baseline input. The integrated gradient for the $i$-th dimension of $x$ if defined as~\shortcite{sundararajan2017axiomatic}:
\begin{equation}
    \text{IntegratedGrads}_{i}(x)=\left(x_{i}-x_{i}^{\prime}\right) \times \int_{\alpha=0}^{1} \frac{\partial F\left(x^{\prime}+\alpha \times\left(x-x^{\prime}\right)\right)}{\partial x_{i}} d \alpha\text{ .}
\end{equation}

By moving from $\alpha=0$ to $1$, the formula allows the score to capture the range of interesting gradients and to accurately quantify the degree of that contribution in the range when the model output is more sensitive to changes in the target feature values.

With this method and given a binary text classification problem, a practitioner could, for instance, compute attribution scores for a given input and for each class to determine which input tokens contribute more to a positive classification and a negative one, respectively. A token score is computed by aggregating the scores of all the features of the token word embedding. This idea is formally expressed in \ref{eq:tok_attr}.

\section{Methodology}

The methodology we propose for identifying new language patterns that can signal human trafficking in OEAs is outlined as follows.

\begin{enumerate}
    \item Train a Transformer-based model on the HTRP and the OAD tasks of the pseudo-labeled Human Trafficking dataset proposed in Chapter \ref{chpt:data}.
    \item For each task, run an IG attribution analysis on the pseudo-labeled data.
    \item Select top-scoring $n$-grams according to their aggregated attribution scores.
    \item Study, by a domain expert, the proposed $n$-grams and determine if new knowledge has been produced or existing knowledge has been validated.
\end{enumerate}

The main assumption behind the proposed method is that a Transformer-based model is capable of capturing language patterns that contribute to predicting an output that is either directly human trafficking prediction or is related to it. According to our assumption, Step (1) could be conducted more generally by training on any problem whose output is correlated to human trafficking activity. Running an IG attribution analysis over the labeled data results in tokens being assigned scores as a function of how much they contribute to certain outputs. Similarly, one could give a score to $n$-grams instead of tokens simply by aggregating the scores of the token components of the $n$-gram. Then, $n$-grams contributing strongly to a human trafficking-related output could somehow be relevant to predicting human trafficking activity. Experiments suggest that this is a high recall-low precision process because of the several noise sources introduced throughout,e.g., pseudo-labeling, model training, and tokenization.

\section{Experiments}

\subsection{Training Setup}

We explored two Transformer models, BERT and RoBERTa. BERT was selected because of its widespread use and compatibility with libraries for interpretability purposes. RoBERTa, in turn, was one of the best-performing models in the Named Entity Recognition system. Hence, we hypothesize it performs better than BERT in this scenario as well. We verify this hypothesis through model testing. The size of these models was adequate for running training and evaluation on the hardware available in a manageable time frame.

The training setup for both tasks is as follows. In both tasks, the models were trained in a Google Cloud server with 8 Tesla V100 of 16GB each, funded by the National Science Foundation. A cross-entropy loss was optimized using AdamW~\shortcite{loshchilov2017decoupled}, with an initial learning rate of $5\cdot 10^{-5}$~(library's default). We trained for 3 epochs, as further training yielded no improvements in the validation set, and used a batch size of 16 samples per device because it was the maximum we could fit in our hardware. In this setting, training the HTRP task takes only a few hours, whereas OAD training lasts around three days. The training routines and the models and tokenizers implementations are from the Transformers Python library~\shortcite{wolfetal2020transformers}.

During training, the train split data was randomly divided into training and validation sets. The division was stratified, and 95\% of samples were assigned to the training set and 5\% to the validation set. The best checkpoint, according to validation performance, was used for reporting the test results.

\subsection{Attribution Scores}

After training the models, attribution scores via Integrated Gradients~(IG) were computed using the Captum Python library~\shortcite{kokhlikyan2020captum}. The scores were computed using a variation of the IG method called Layer Integrated Gradients, which computes the attributions with respect to the input/output of a layer instead of the input features. In our specific context, for the purpose of attributing decisions to the input words, we employ a scoring mechanism that relies on attributions derived from the features of the word embedding layer's output. The process is as follows.

Let $S=[t_1,t_2,\dots,t_N]$ be a sequence of $N$ tokens, the word embedding layer in the Transformer converts each token into a vector of $D$ dimensions, i.e., it produces a sequence $S_e=[e_1,e_2,\dots,e_N]$, where $e_i\in \mathbb{R}^D$ for all $i$. For every $i$, we then obtain an attribution scores vector $a_i\in \mathbb{R}^D$. The final attribution score of token $t_i$ is the normalized sum of the components of $a_i$, defined as follows:
\begin{equation}
    \text{attr}_i = \frac{\sum_{1\leq j \leq D} a_{i,j}}{||a_i||_2}\text{ .}
\end{equation}\label{eq:tok_attr}

Finally, to compute relevant $n$-grams, we iterated the target collection, and for each occurrence of an $n$-gram, the sum of the attribution scores of its components determines the $n$-gram attribution. The final relevance score of the $n$-gram is obtained by averaging the attribution scores of the individual occurrences. For each $n$-gram, we also keep track of its number of occurrences and the standard deviation of the instances' scores. For example, if an $n$-gram occurs three times in a target corpus, and the occurrences are given attributions $0.1$,$-0.8$, and $1.2$, respectively, then we compute the following entries for that $n$-gram:

\begin{itemize}
    \item Average attribution score: $0.2$
    \item Standard deviation of the attributions: $0.82$
    \item Number of occurrences: 3
\end{itemize}

\subsection{Results}

Tables \ref{tab:htrp_results} and \ref{tab:oad_results} show the results in training, validation, and testing of the models in each task, respectively. Results are reported in terms of balanced accuracy, precision, recall, $F_1$ score, and the area under the Receiver Operating Characteristic curve (AUC ROC). Importantly, the output of the models is two logits. The balanced accuracy, precision, recall, and $F_1$ score metrics are calculated using only the positive logit after applying softmax and by setting the binary classification threshold that maximizes the $F_1$ score in the validation set. The AUC ROC metric was equally calculated using the positive class probability.

\begin{table}[]
    \centering
    \caption{Results of the BERT and RoBERTa models in the HTRP task.}
    \begin{subtable}{\textwidth}
       \centering
       \begin{tabular}{|c|>{\centering\arraybackslash}p{0.135\textwidth}|>{\centering\arraybackslash}p{0.135\textwidth}|>{\centering\arraybackslash}p{0.135\textwidth}|>{\centering\arraybackslash}p{0.135\textwidth}|>{\centering\arraybackslash}p{0.135\textwidth}|}
        \hline
         \textbf{Model} &  \textbf{Bal. Acc.} &  \textbf{Precision} & \textbf{Recall} & \textbf{$\mathbf{F_1}$ score} & \textbf{AUC} \\
         \hline
            BERT & 0.64 $\pm$ 0.03& 0.58 $\pm$ 0.02 & 0.34 $\pm$ 0.07 & 0.42 $\pm$ 0.05 & 0.83 $\pm$ 0.02 \\
            RoBERTa & 0.67 $\pm$ 0.02 & 0.56 $\pm$ 0.02 & 0.40 $\pm$ 0.04 & 0.47 $\pm$ 0.02 & 0.85 $\pm$ 0.00 \\
         \hline
    \end{tabular}
    \caption{Training}
    \label{tab:htrp_results_train}
   \end{subtable}

       \begin{subtable}{\textwidth}
       \centering
       \begin{tabular}{|c|>{\centering\arraybackslash}p{0.135\textwidth}|>{\centering\arraybackslash}p{0.135\textwidth}|>{\centering\arraybackslash}p{0.135\textwidth}|>{\centering\arraybackslash}p{0.135\textwidth}|>{\centering\arraybackslash}p{0.135\textwidth}|}
        \hline
         \textbf{Model} &  \textbf{Bal. Acc.} &  \textbf{Precision} & \textbf{Recall} & \textbf{$\mathbf{F_1}$ score} & \textbf{AUC} \\
         \hline
            BERT  & 0.64 $\pm$ 0.03 & 0.57 $\pm$ 0.02 & 0.32 $\pm$ 0.08 & 0.40 $\pm$ 0.06 & 0.83 $\pm$ 0.02 \\
            RoBERTa & 0.67 $\pm$ 0.02 & 0.56 $\pm$ 0.01 & 0.39 $\pm$ 0.04 & 0.46 $\pm$ 0.02 &  0.85 $\pm$ 0.00 \\
         \hline
    \end{tabular}
    \caption{Validation}
    \label{tab:htrp_results_val}
   \end{subtable}

       \begin{subtable}{\textwidth}
       \centering
       \begin{tabular}{|c|>{\centering\arraybackslash}p{0.135\textwidth}|>{\centering\arraybackslash}p{0.135\textwidth}|>{\centering\arraybackslash}p{0.135\textwidth}|>{\centering\arraybackslash}p{0.135\textwidth}|>{\centering\arraybackslash}p{0.135\textwidth}|}
        \hline
         \textbf{Model} &  \textbf{Bal. Acc.} &  \textbf{Precision} & \textbf{Recall} & \textbf{$\mathbf{F_1}$ score} & \textbf{AUC} \\
         \hline
            BERT  & 0.63 $\pm$ 0.03 & 0.56 $\pm$ 0.02 & 0.31 $\pm$ 0.07 & 0.39 $\pm$ 0.05 & 0.81 $\pm$ 0.02 \\
            RoBERTa & 0.66 $\pm$ 0.01 & 0.55 $\pm$ 0.02 & 0.37 $\pm$ 0.04 & 0.44 $\pm$ 0.02 & 0.83 $\pm$ 0.00 \\
         \hline
    \end{tabular}
    \caption{Testing}
    \label{tab:htrp_results_test}
   \end{subtable}
      \label{tab:htrp_results}
\end{table}

\begin{table}[]
    \centering
    \caption{Results of the BERT and RoBERTa models in the OAD task.}
    \begin{subtable}{\textwidth}
       \centering
       \begin{tabular}{|c|>{\centering\arraybackslash}p{0.135\textwidth}|>{\centering\arraybackslash}p{0.135\textwidth}|>{\centering\arraybackslash}p{0.135\textwidth}|>{\centering\arraybackslash}p{0.135\textwidth}|>{\centering\arraybackslash}p{0.135\textwidth}|}
        \hline
         \textbf{Model} &  \textbf{Bal. Acc.} &  \textbf{Precision} & \textbf{Recall} & \textbf{$\mathbf{F_1}$ score} & \textbf{AUC} \\
         \hline
            BERT & 0.99 $\pm$ 0.01 & 0.99 $\pm$ 0.01 & 0.98 $\pm$ 0.01 & 0.99 $\pm$ 0.01 & 0.99 $\pm$ 0.00 \\
            RoBERTa & 0.98 $\pm$ 0.01 & 0.99 $\pm$ 0.00 & 0.98 $\pm$ 0.01 & 0.98 $\pm$ 0.01 & 0.99 $\pm$ 0.00  \\
         \hline
    \end{tabular}
    \caption{Training}
    \label{tab:oad_results_train}
   \end{subtable}

       \begin{subtable}{\textwidth}
       \centering
       \begin{tabular}{|c|c|c|c|c|c|}
        \hline
         \textbf{Model} &  \textbf{Bal. Acc.} &  \textbf{Precision} & \textbf{Recall} & \textbf{$\mathbf{F_1}$ score} & \textbf{AUC} \\
         \hline
            BERT  & 0.98 $\pm$ 0.01 & 0.98 $\pm$ 0.00 & 0.97 $\pm$ 0.01 & 0.98 $\pm$ 0.01 & 0.99 $\pm$ 0.00 \\
            RoBERTa & 0.97 $\pm$ 0.01 & 0.98 $\pm$ 0.00 & 0.97 $\pm$ 0.01 & 0.97 $\pm$ 0.01 & 0.99 $\pm$ 0.00  \\
         \hline
    \end{tabular}
    \caption{Validation}
    \label{tab:oad_results_val}
   \end{subtable}

       \begin{subtable}{\textwidth}
       \centering
       \begin{tabular}{|c|c|c|c|c|c|}
        \hline
         \textbf{Model} &  \textbf{Bal. Acc.} &  \textbf{Precision} & \textbf{Recall} & \textbf{$\mathbf{F_1}$ score} & \textbf{AUC} \\
         \hline
            BERT  & 0.92 $\pm$ 0.01 & 0.98 $\pm$ 0.00 & 0.85 $\pm$ 0.02 & 0.91 $\pm$ 0.01 & 0.98 $\pm$ 0.00 \\
            RoBERTa & 0.92 $\pm$ 0.00 & 0.98 $\pm$ 0.00 & 0.86 $\pm$ 0.00 & 0.92 $\pm$ 0.00 & 0.98 $\pm$ 0.00  \\
         \hline
    \end{tabular}
    \caption{Testing}
    \label{tab:oad_results_test}
   \end{subtable}
      \label{tab:oad_results}
\end{table}

In Table \ref{tab:htrp_results}, we observe how the RoBERTa model performs better than the BERT model. Notice the $F_1$ score is relatively low, which speaks to the complexity of the task presented. It could also be an indication that the task is ill-posed and ambiguous. If we contrast those with the results in the OAD task~(Table \ref{tab:oad_results}), the latter are much better. Both models perform considerably well, almost perfectly in the training dataset, and the results generalize especially well to the testing set. Notably, the drop in the recall values from training/validation to testing called our attention. We suspect that, since the training/testing split was done by splitting connected components, there are similarity patterns that appear in the testing components that do not appear in the training ones. Thus hindering the model's ability to properly generalize to those.

After obtaining a trained model HTRP, we computed the attribution scores of the $n$-grams with respect to the positive class prediction score. We experimented with several various $n$ and with both the training and testing data for comparison purposes.

\begin{figure}[t]
    \centering
    \includegraphics[width=.9\textwidth]{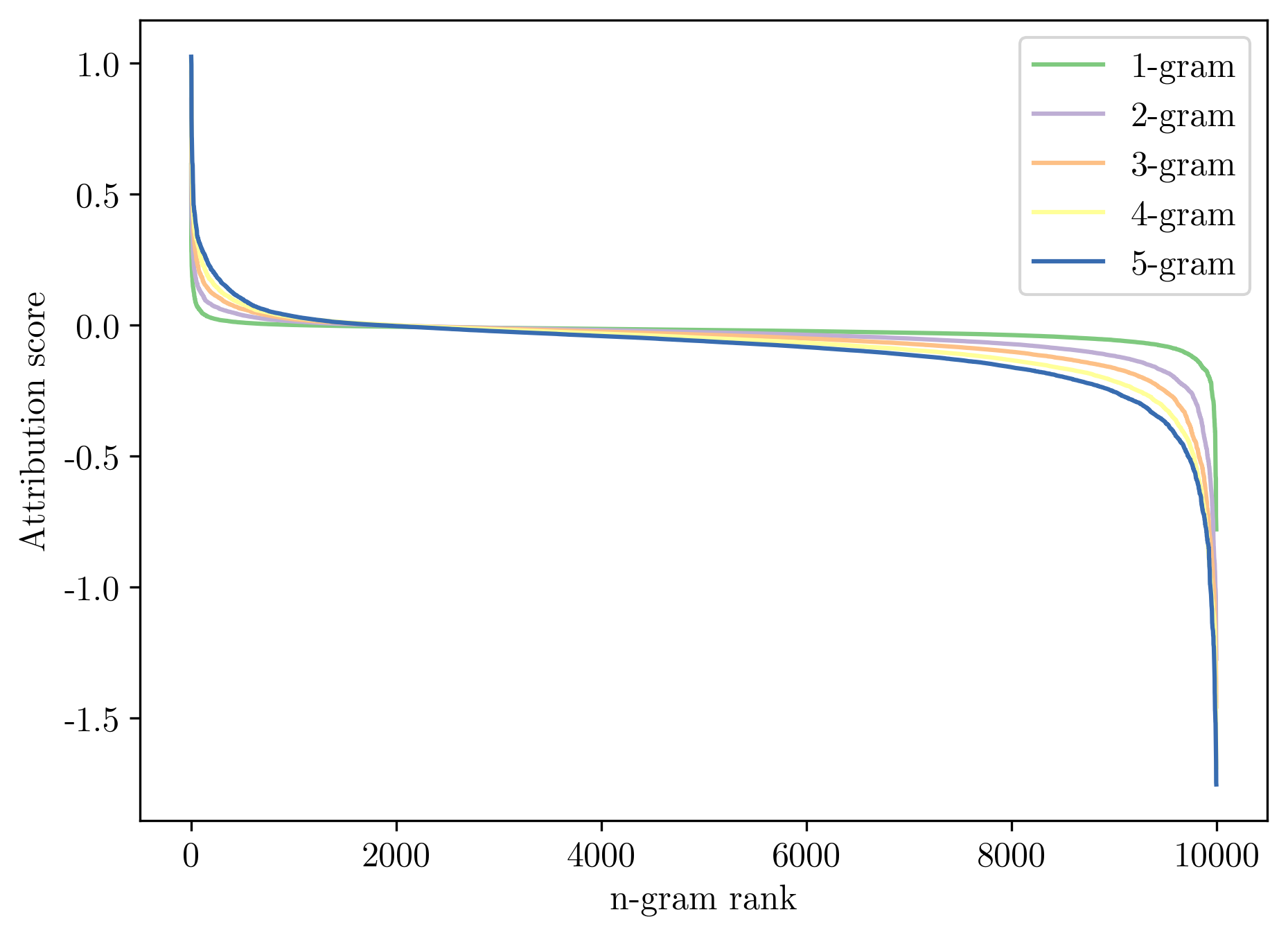}
    \caption{Sorted mean attribution score of 10,000 randomly sampled $n$-grams for different values of $n$ in the testing data.}
    \label{fig:ngram_cmp}
\end{figure}

First, observe in Figure \ref{fig:ngram_cmp} the distribution of the scores for different values of $n$ computed in the testing set\footnote{Training attributions behave similarly.}. As expected, the behavior for different values of $n$ is similar. The scores of longer $n$-grams are higher because the attribution of an $n$-gram is computed as the sum of the attributions of its components.

\begin{figure}[t!]
    \centering
    \includegraphics[width=\textwidth]{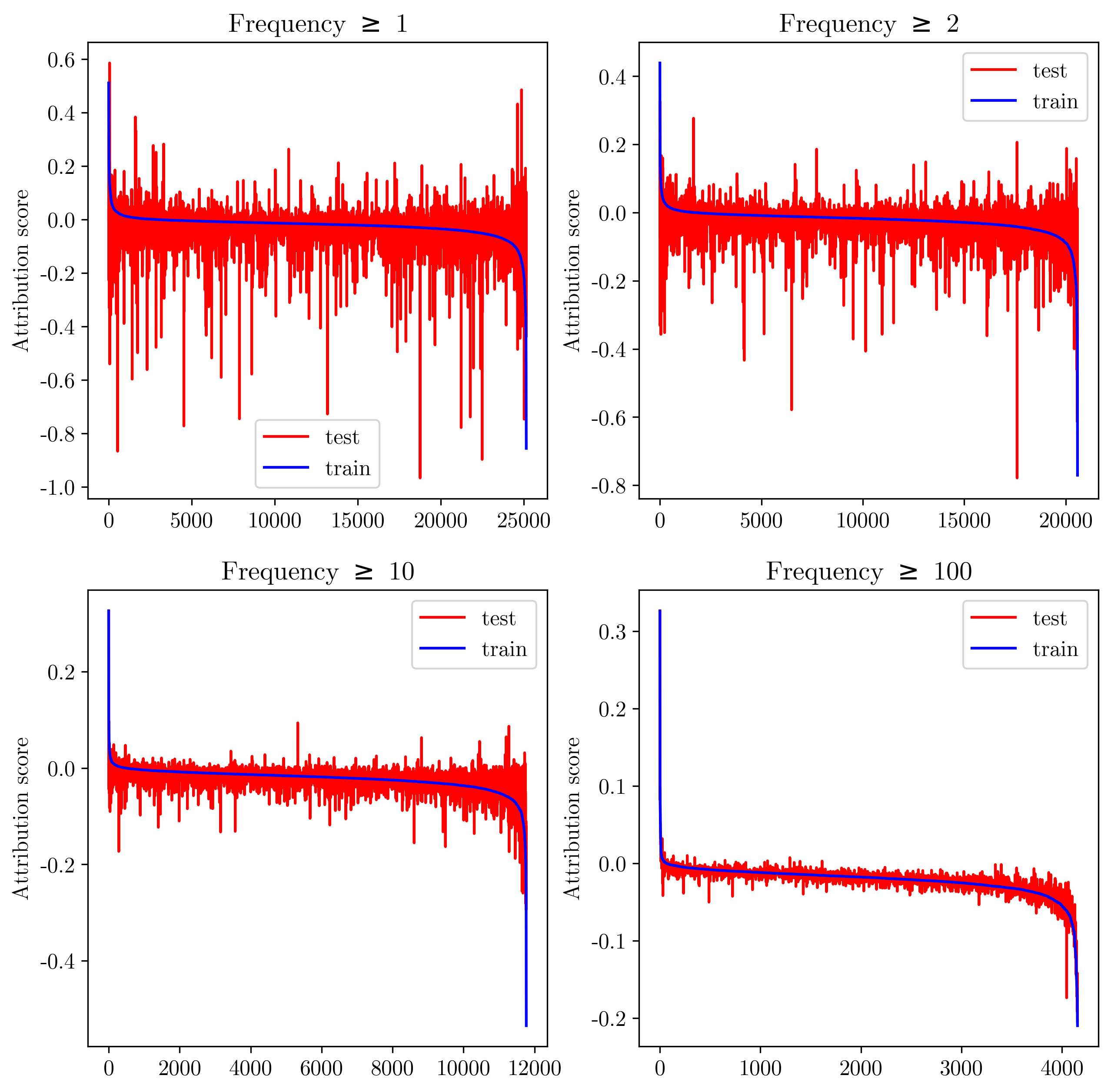}
    \caption{Comparison of the behavior of the attribution scores of shared tokens in the training and testing datasets. The x-coordinate of a point in the plots represents a shared token between the train and test datasets.}
    \label{fig:1gram_cmp}
\end{figure}

Next, we evaluate the differences between the behavior of the $n$-grams attribution scores in the train and test sets. To show the comparison, we computed a curve for the 1-grams~(tokens) of the training set in a similar fashion to those of Figure \ref{fig:ngram_cmp}. Then, we plotted the curve for the testing set but using the tokens in the intersection of both sets. Moreover, the curve of the test set is plotted so that values in the x-axis represent the same token. This allows us to faithfully compare the changes in attributions for the tokens in testing with respect to training. Additionally, we contrast the behavior considering tokens with different minimum frequencies. Figure \ref{fig:1gram_cmp} illustrates the results.

From Figure \ref{fig:1gram_cmp}, the plot for the testing set appears to be noisy. The fluctuation indicates a high variability of the attributions in testing with respect to the behavior in the training set. However, more frequently occurring tokens exhibit far less variability. These results suggest relying more on tokens occurring more frequently. Also, we expect a similar behavior with longer $n$-grams.

\begin{figure}[t]
    \centering
    \includegraphics[width=.7\textwidth]{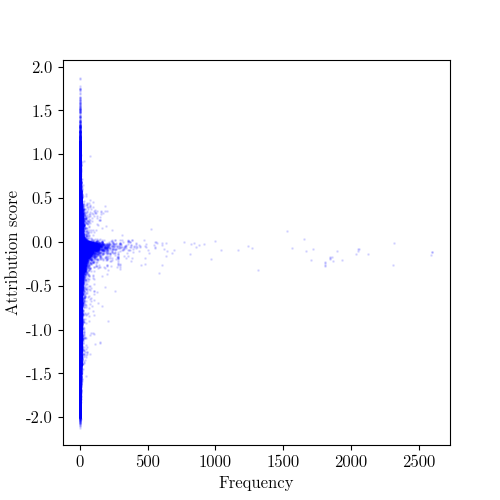}
    \caption{Plot of the attribution score versus the frequency of $5$-grams in the testing data.}
    \label{fig:freq_attr}
\end{figure}

Conversely, the most interesting behavior, as far as our data is concerned, occurs in the $n$-grams with lower frequencies. Indeed, as observed in Figure \ref{fig:freq_attr}, where each point $(x,y)$ represents an $n$-gram with frequency $x$ and attribution score $y$, the $n$-grams with extreme attribution scores occur less frequently.

\begin{figure}[t]
    \centering
    \includegraphics[width=\textwidth]{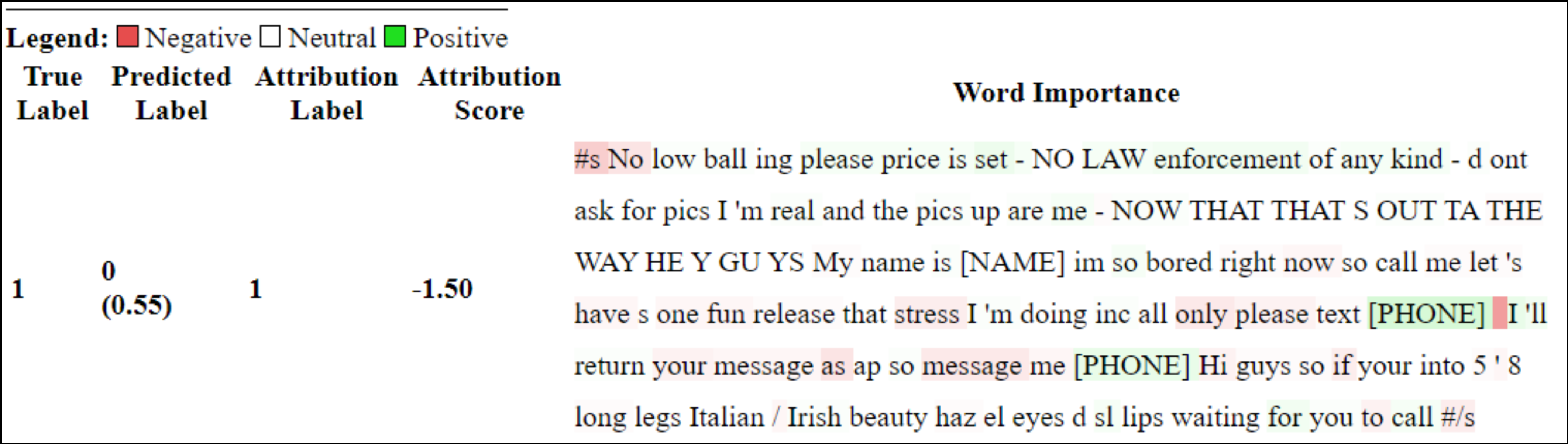}
    \caption{Attribution report on an input advertisement.}
    \label{fig:attr_example}
\end{figure}

Next, observe in Figure \ref{fig:attr_example} a visual example of the type of information that results from running the attribution analysis on a specific example. Observe in the image that the report contains information about the real and predicted labels, the logit upon which attribution was calculated~(Attribution label), and the overall attribution score for that example.

We have presented this kind of visualization to our collaborators with expertise in information systems and criminology, and we obtained validation of the utility of the information presented to analyze the characteristics of the language in the advertisement.

\begin{figure}[t]
    \centering
    \includegraphics[width=0.9\textwidth]{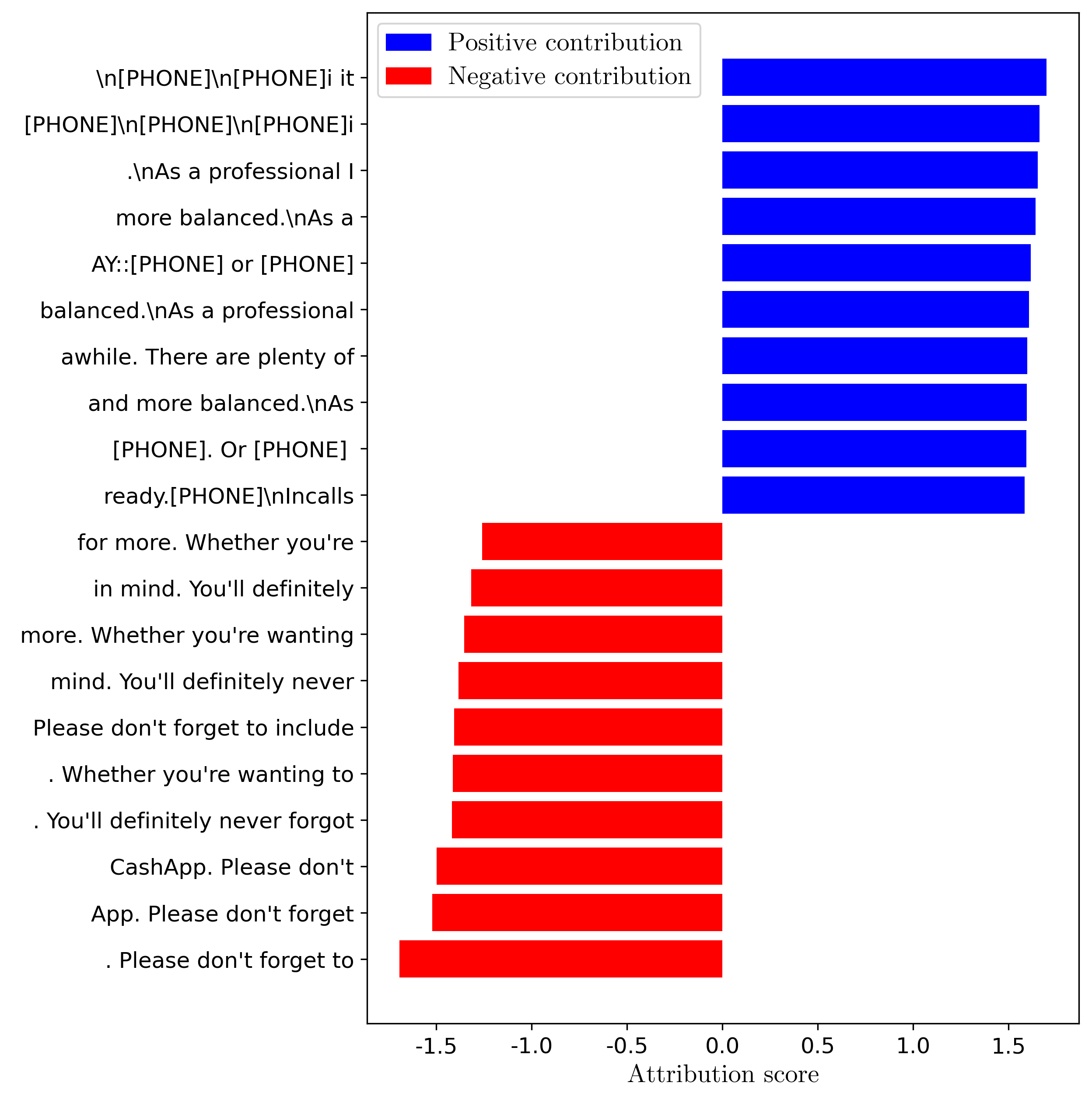}
    \caption{Top 10 $6$-grams with the highest attributions and bottom 10 with the lowest.}
    \label{fig:6gram_cmp_top10}
\end{figure}

Subsequently, we analyze the top-scoring $n$-grams. Figure \ref{fig:6gram_cmp_top10} shows a list of the top 10 $6$-grams with the most positive contribution to the positive-class logit~(blue) and also the top 10 that have the most negative contribution~(red). Importantly, these examples come from the test dataset, and we only focus our analysis on examples that were correctly predicted as positive. 

Notably, all $6$-grams displayed in Figure \ref{fig:6gram_cmp_top10} occur only once in the sub-sample of the dataset. Consequently, and following the insights given by Figure \ref{fig:freq_attr}, we plotted the same graphic, but only with $6$-grams occurring more than 10 times. In this case, the black line at the end of each bar informs us about the standard deviation from the mean attribution score of the respective $6$-gram.

\begin{figure}[t]
    \centering
    \includegraphics[width=0.9\textwidth]{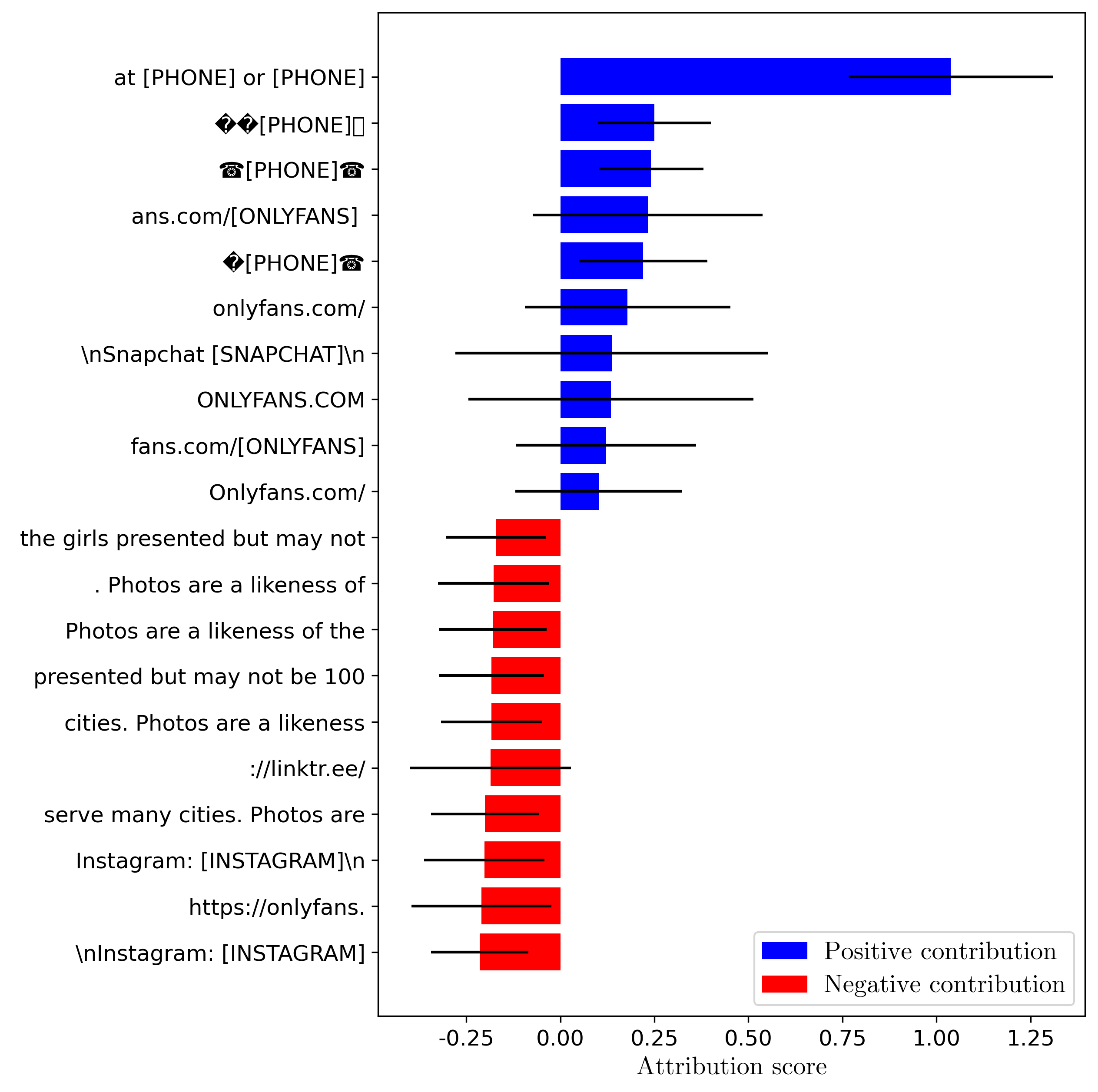}
    \caption{Top 10 $6$-grams with the highest attributions and bottom 10 with the lowest, out of the $6$-grams occurring at least 10 times.}
    \label{fig:6gram_cmp_top10_thr10}
\end{figure}

The results we observe from these tables, as well as from further experimentation we did, inform us that the model appears to be biased toward giving a positive prediction to inputs with the token \texttt{[PHONE]}. This is less surprising considering the analysis dataset biases presented in Chapter \ref{chpt:data}. This issue clearly limits the ability of the knowledge that can be obtained from the calculation of the attribution of each word in the input.

As we further analyzed the posts for HTRP, we noticed, for example, that amongst the ads in which the model is most certain about their prediction as carrying a positive risk of human trafficking, we find the repeated pattern ``[PHONE] or [PHONE]'', or some closely related variant. This pattern is the result of the authors giving two different phone numbers as contact information, thus getting flagged by our pseudo-labeling as a positive sample. Even though multiple phone numbers could indicate the presence of organized activity, the fact that the model makes its most certain predictions on posts exhibiting this pattern can be explained by and acts as further evidence of the aforementioned bias.

Interestingly, among the top 1,000 most positive entries, the average length for a post is $\approx 717$ characters, whereas, across the posts where the model is most certain they are not human trafficking, the average length is $\approx 222$ characters. Indeed, short, poorly described posts tend to be labeled as negative because they offer no personally identifiable information, and no connections are established between them and others.

\begin{figure}[t]
    \centering
    \includegraphics[width=\textwidth]{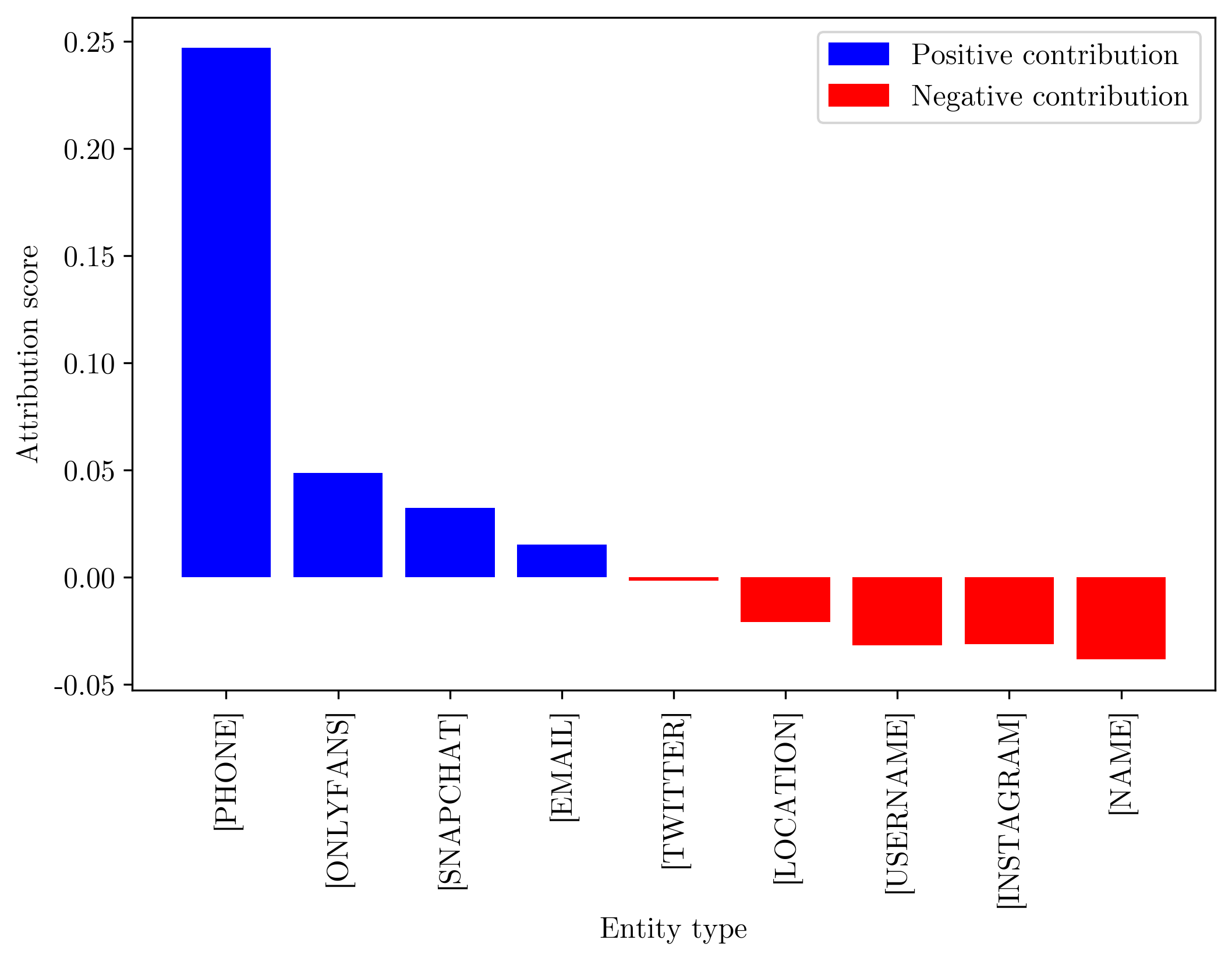}
    \caption{Comparison of the attribution scores of the special mask tokens that represent extracted entities.}
    \label{fig:mask_cmp}
\end{figure}

Finally, observe in Figure \ref{fig:mask_cmp} the average attributions of the masked tokens in the testing set for HTRP. As observed, the \texttt{[PHONE]} token alone makes a very strong contribution to a positive prediction. Notably, the presence of others masked tokens detriment the positive score. Indeed, the top three tokens with the most significant negative contribution were not used to pseudo-label the dataset, only for masking purposes. Consequently, we could consider these negative attribution scores to be unbiased by the pseudo-labeling process. 


We propose ways to confirm and mitigate the bias issue that arose from following our methodology. First, we deem it necessary to design a labeling task in which expert collaborators are asked to label an advertisement based on whether they have reasons to believe that it is human trafficking or not. This would allow us to measure, in the first place, the quality of the pseudo-labels created from the Relatedness Graph and, second, whether human experts reproduce the same biases as our model.

Moreover, we consider several alternatives to adapt the dataset generation process to mitigate the occurrence of the bias observed. First, and since it occurs primarily with phone numbers, one could consider augmenting the threshold used for the hard identifiers heuristic. Recall that when a connected component has two or more distinct phone numbers, that component is labeled positive. We changed it to three and observed the results. This slightly improves the balance between the positive and negative classes, but not significantly. Thus, we do not believe it is going to be enough.

A different alternative is to dispense with the hard identifiers heuristic altogether for pseudo-labeling. In this case, we observed the relative presence of hard identifiers between positive and negative class levels. However, the difference in the relative occurrence of locations between the two classes increases substantially, which is expected given that with such a modification, the labeling relies on the distance heuristic solely.

A second alternative to mitigating the bias is to remove the occurrence of those tokens completely. In preliminary work, we were not even masking the information of hard identifiers. We observed the models were obsessed with these tokens. Hence, our alternative solution is to mask them. Consequently, we speculate that removing these tokens' instances will force the models to pay attention to different, more informative parts of the inputs.

\section{Conclusion}

In this chapter, we presented a novel methodology that leverages the Integrated Gradients~(IG) interpretability method in order to inspect the predictions made by Transformer models fine-tuned in artificial datasets related to human trafficking. We were able to show that Transformer models are adequate for fitting these complex tasks. We assessed that a RoBERTa model performs slightly better than a BERT model in this context, backing our hypothesis that using tokenizers like Byte-Pair Encoding that allows tokens to go to the level of bytes is beneficial in a domain with very noisy language.

However, applying IG was unfavorable as far as the interpretation of the results is concerned. Through the proposed methodology, we realized that the fine-tuned model in the task of Human Trafficking Risk Prediction appears to be biased towards the positive class whenever the mask token corresponding to a phone number appears in the input text. We have strong reasons to believe that the dataset generation process induces this bias, so we propose ways to attempt to correct it.

While our current findings may not align with initial expectations, it is essential to emphasize that our research remains an active pursuit dedicated to unraveling complex challenges and addressing unresolved issues.

%
\chapter{Conclusion}\label{chpt:conclusion}

In this thesis, we delved into the complex realm of consumer-to-consumer~(C2C) marketplaces and the ever-evolving challenges posed by illicit activities, particularly in the context of human trafficking. As we explored the interplay between the thriving online C2C platforms and the persistent issue of human exploitation, it became evident that innovative, technology-driven solutions are imperative for combating such criminal activities effectively.

We initiated our investigation by comprehensively reviewing existing literature pertaining to human trafficking, especially within the context of online escort websites. This review not only highlighted the scarcity of curated, publicly available data but also underscored the critical role of Natural Language Processing (NLP) in understanding and combating this pervasive issue. Leveraging this understanding, we formulated a robust methodology capable of extracting insights from the complex and often obfuscated language utilized by offenders in online advertisements. Our approach integrated a meticulous process for developing pseudo-labeled datasets, underpinned by minimal human supervision, facilitating the training and evaluation of state-of-the-art NLP models.

Through rigorous experimentation and analysis, we harnessed the power of advanced NLP models, employing them to predict human trafficking activity and detect connected advertisements within the data. Moreover, we devised a methodology for interpreting the predictions of these models, thereby extracting explainable insights that could guide and inform law enforcement efforts.

Our work presents several significant contributions, including a Systematic Literature Review~(SLR) of computational literature on human trafficking, a flexible methodology for dataset development with minimal supervision, an evaluation of cutting-edge Transformer models for Named Entity Recognition within the challenging context of online escort advertisements, and a robust approach to extracting insights from neural network predictions. These contributions not only serve as a foundation for further research but also hold the potential to aid law enforcement agencies in their ongoing battle against human trafficking in the digital realm.

In the face of the ever-evolving challenges posed by online criminal activities, our research underscores the importance of continued innovation and interdisciplinary collaboration, laying the groundwork for more effective and targeted approaches to combating human exploitation in the digital sphere. As we conclude this thesis, we advocate for sustained efforts to harness the power of technology and research, empowering us to proactively tackle the intricate and multifaceted issue of human trafficking in the digital age.

\subsection{Future Directions}

Throughout the development of this investigation, we identified several avenues that we did not explore and that we believe are worth pursuing. We expand on those in the following paragraphs.

Regarding our systematic inspection of the literature, the scope of our review was limited, and it could be expanded to cover other interesting areas like multi-modal approaches to the same problems, usage of data outside the scope of online escort advertisements~(OEAs), and a deeper investigation of other problems different from Human Trafficking Risk Prediction~(HTRP) and Organized Activity Detection~(OAD).

Our pseudo-labeled dataset generation methodology described in Chapter \ref{chpt:data} also offers multiple opportunities for further investigation. For example, different, more varied and comprehensive sources of data could be used to create the collection of OEAs and observe how that impacts the overall results and characteristics of the datasets. If different platforms are incorporated, one might consider the interesting problem of cross-linking, to connect references to the same entities across different websites. Furthermore, other Machine Learning tasks associated with human trafficking could be defined apart from HTRP and OAD. Additionally, the heuristics we propose can be improved and new ones could be proposed. Finally, a number of limitations of our approach were listed. Future research could work to address the limitations we identified or others.

Regarding the Named Entity Recognition system we proposed, our work could be extended with the execution of a more thorough hyperparameter tuning. Additionally, the results of some entity types were significantly worse than others. Future research could aim at bridging the gap between them. We targeted our NER pipeline only to location and identifiable information, but we identified other valuable information that is significant in the context of human trafficking analysis like physical characteristics of the victims, services offered, price information, and advertisers' constraints. All these could be incorporated as new entity types.

Chapter \ref{chpt:attr} leaves the most open questions and avenues. Integrated Gradients proved to be a sensible method for interpreting results in Transformer-based models in the context of OEAs. However, others could be explored. For instance, a recent paper~\shortcite{hao2021self} proposed a self-attention attribution method to interpret the information inside Transformer attention heads.

Furthermore, we only trained models in the individual tasks of HTRP and OAD. However, future investigators could attempt to train them in a joint fashion. Also, we selected pre-trained models and directly fine-tuned them in our specific tasks. Instead, models could be further pre-trained in their original objectives but using OEAs, provided that there is sufficiently large data.

\section*{Ethics Statement}
This study relies solely on internally scraped and curated datasets and transformer-based models, and does not involve humans as subjects; however, the original data contains personally identifiable information that prevents the investigators from releasing the dataset to the general public. The methodology is supported by internal review board (IRB) approval at Baylor University. 

Further, while we have vetted our model regarding ethical considerations, we acknowledge that it may inherit biases in the original transformer-based NER embeddings. We emphasize the need for further research to mitigate these biases and are committed to methodological transparency.

\section*{Acknowledgements}
This work is supported by the National Science Foundation under Grant Nos. 2039678, 2136961, and 2210091. 

The views expressed herein are solely those of the author and do not necessarily reflect those of the National Science Foundation.


\clearpage
\vspace*{4.25in}
 \begin{center}
      APPENDICES
 \end{center}
 \pagestyle{plain}
  \pagebreak

\newpage
\appendix
\renewcommand\thesection{\Alph{chapter}.\arabic{section}}
\chapter{Methods for Seed Papers} \label{ap:seed_papers}

\section{Search Protocol} \label{ap:ml}

The study was populated by searching for the terms: ``escort ads''. ``human trafficking'' and ``machine learning''. A time filter was applied to limit results to the range 2012-2022. The indexers used were Google Scholar, Jstor and Ebsco, ACM DL, IEEE Xplore, Springer Link, and Science Direct.

\section{Inclusion and Exclusion Criteria}

The inclusion and exclusion criteria used are the following:

\begin{itemize}
    \item Detection of escort ads linked to sex trafficking
    \item Uses machine learning technology (e.g., natural language processing).
    \item English language paper published with an academic outlet (e.g., peer-review article or conference proceeding).
    \item Uses data from escort ads to train the program.
    \item Project involves training a customized or created program (not testing an off-the-shelf system).
    \item Published between 2012 – 2022 (current efforts given evolving communications landscape and legal proceedings).
\end{itemize}

\section{Initial Sample}

Table \ref{tab:indexers} summarizes the search results by indexer.

\begin{table}[H]
    \centering
    \begin{tabular}{|>{\centering\arraybackslash}m{0.18\textwidth}|>{\centering\arraybackslash}m{0.16\textwidth} |>{\centering\arraybackslash}m{0.16\textwidth}|>{\centering\arraybackslash}m{0.16\textwidth}|>{\centering\arraybackslash}m{0.19\textwidth}|}
        \hline
        \textbf{Indexer} & \textbf{Date Queried} & \textbf{Found \mbox{Documents}} & \textbf{Fit \mbox{Criteria}} & \textbf{Contribution to Initial \mbox{Sample}}\\
        \hline
         Google Scholar & 06/20/2022 & 155 & 47 & 47 \\
         Jstor and Ebsco & 06/20/2022 & 0 & 0 & 0 \\
         ACM DL & 07/18/2022 & 56 & 4 & 2 \\
         IEEE Xplore & 07/18/2022 & 25 & 6 & 1 \\
         Springer Link & 07/18/2022 & 50 & 2 & 0 \\
         Science Direct & 07/18/2022 & 7 & 1 & 0 \\
         \hline
         Total & - & 293 & 60 & 50 \\
         \hline
    \end{tabular}
    \caption{Search results in several indexers used.}
    \label{tab:indexers}
\end{table}

The column \textbf{Contribution to Initial Sample} counts the number of non-duplicate works. In this case, 10 documents were duplicated in Google Scholar: ACM DL (2), IEEE Xplore (5), Springer Link (2), and Science Direct (1). The occurrences in Google Scholar are counted as the ones contributing to the initial sample.

Upon closer inspection, many of the documents that passed the initial screening protocol did not describe research that fit within the scope of the review, e.g., machine learning was involved, but it was not an NLP model. Of the 50 documents examined, 29 papers were excluded (see Table \ref{tab:excluded}). The final sample included 21 original studies.

\begin{table}[H]
    \centering
    \caption{Search results in several indexers used.}
    \begin{tabular}{|p{0.58\textwidth}|>{\centering\arraybackslash}m{0.2\textwidth} |>{\centering\arraybackslash}m{0.15\textwidth}|}
        \hline
        \textbf{Reason} & \textbf{Documents Excluded} & \textbf{Percent} \\
        \hline
         Not original (multiple articles for the same study, i.e., dissertation and paper) & 7 & 24.14\\
         \hline
         Incomplete study (research note about work in progress or training protocols not reported) & 3 & 10.34\\
         \hline
         Not an NLP model to detect trafficking & 17 & 58.62\\
         \hline
         Data source did not include C2C escort ads & 2 & 6.90\\
         \hline
         \textbf{Total Excluded} & 29 & 100.00 \\
         \hline
    \end{tabular}
    
    \label{tab:excluded}
\end{table}

\chapter{Comprehensive Results of the Literature Review.} \label{ap:comprehensive}

\setcounter{rowno}{0}

\begin{longtable}{|>{\stepcounter{rowno}\therowno.}p{.05\columnwidth}|p{.40\columnwidth}|p{.27\columnwidth}|p{.18\columnwidth}|}
    \caption{Final list of relevant studies.}\label{tab:fulllist} \\\hline
    
    \multicolumn{1}{|r|}{} & \textbf{Title} & \textbf{Short Name} & \textbf{Citation} \\\hline
    \endfirsthead
    
    \multicolumn{3}{c}%
    {{\bfseries \tablename\ \thetable{} -- continued from previous page}} \\\hline \multicolumn{1}{|r|}{} & \textbf{Title} & \textbf{Short Name} & \textbf{Citation} \\\hline
    \endhead
    
    \hline \multicolumn{4}{r}{{Continued on next page}} \\ 
    \endfoot
    
    \hline
    \endlastfoot
        & Active Search of Connections for Case Bulding and Combating Human Trafficking & ACTIVESEARCH & \citeA{rabbany2018active} \\\hline
        
        & Backpage and Bitcoin: Uncovering Human Traffickers & BACKPAGEBITCOIN & \citeA{portnoff2017backpage} \\\hline

        & Building and Using a Knowledge Graph to Combat Human Trafficking & BUILDINGKG & \citeA{szekely2015building} \\\hline

        & Detecting Sex Trafficking Circuits in the U.S. Through Analysis of Online Escort Advertisements & CIRCUITSUS & \citeA{ibanez2016detecting} \\\hline

        & Context-specific Language Modeling for Human Trafficking Detection from Online Advertisements & CONTEXTLM & \citeA{esfahani2019context} \\\hline

        & Extracting Co-occurences of Emojis and Words as Important Features for Human Trafficking Detection Models & COOCCUREMOJI & \citeA{wiriyakun2022extracting} \\\hline

        & Coupled Clustering of Time-Series and Networks & COUPLEDCLUSTER & \citeA{liu2019coupled} \\\hline
        
        & Detecting Covert Sex Trafficking Networks in Virtual Markets & COVERTSEX & \citeA{ibanez2016detecting} \\\hline

        & Cracking Sex Trafficking Data Analysis, Pattern Recognition, and Path Prediction & CRACKINGSEX & \citeA{keskin2021cracking} \\\hline

        & Data Integration from Open Internet Sources to Combat Sex Trafficking of Minors & DATAINTMINORS & \citeA{wang2012data} \\\hline
        
        & Data Integration from Open Internet Sources and Network Detection to Combat Underage Sex Trafficking & DATAINTUNDER & \citeA{silva2014data} \\\hline

        & DeltaShield Information Theory for Human Trafficking Detection & DELTASHIELD & \citeA{vajiac2023deltashield} \\\hline

        & Detection and Characterization of Human Trafficking Networks Using Unsupervised Scalable Text Template Matching & DETECTCHAR & \citeA{li2018detection} \\\hline

        & Detection of Domestic Human Trafficking Indicators and Movement Trends Using Content Available on Open Internet Sources & DETECTDOMESTIC & \citeA{ibanez2014detection} \\\hline

        & Detection of Organized Activity in Online Escort Advertisements & DETECTIONORG & \citeA{kulshrestha2021detection} \\\hline

        & Don't Want to Get Caught Don't Say It: The Use of Emojis in Online Human Sex Trafficking Ads & DONTWANT & \citeA{whitney2018dont} \\\hline

        & Ensemble Sentiment Analysis to Identify Human Trafficking in Web Data & ENSEMBLESENT & \citeA{mensikova2018ensemble} \\\hline

        & An Entity Resolution Approach to Isolate Instances of Human Trafficking Online & ENTITYRES & \citeA{nagpal2017entity} \\\hline

        & Feature Selection for Human Trafficking Detection Models & FEATURESELECT & \citeA{wiriyakun2021feature} \\\hline

        & FlagIt: A System for Minimally Supervised Human Trafficking Indicator Mining & FLAGIT & \citeA{kejriwal2017flagit} \\\hline

        & Detecting Food Safety Risks and Human Trafficking Using Interpretable Machine Learning Methods & FOODSAFETY & \citeA{zhu2019detecting} \\\hline

        & Identification and Detection of Human Trafficking Using Language Models & IDDETECTLM & \citeA{zhu2019identification} \\\hline

        & Identifying Sex Trafficking in Adult Services Websites: an Exploratory Study with a British Police Force & IDENTBRITISH & \citeA{l2021identifying} \\\hline

        & Identifying Human Trafficking Indicators in the UK Online Sex Market & IDENTIFYINGHTUK & \citeA{giommoni2021identifying} \\\hline

        & Knowledge Graphs for Social Good: An Entity-Centric Search Engine for the Human Trafficking Domain & KGFORSOCIAL & \citeA{kejriwal2022knowledge} \\\hline

        & A Knowledge Management Approach to Identify Victims of Human Sex Trafficking & KNOWMANAGE & \citeA{hultgren2018knowledge} \\\hline
        
        & Leveraging Publicly Available Data to Discern Patterns of Human-Trafficking Activity  & LEVERAGING &  \citeA{dubrawski2015leveraging} \\\hline

        & Always Lurking: Understanding and Mitigating Bias in Online Human Trafficking Detection & LURKING & \citeA{hundman2018always} \\\hline

        & A Multi-Input Machine Learning Approach to Classifying Sex Trafficking from Online Escort Advertisements & MULTIINPUT & \citeA{summers2023multi} \\\hline

        & Network-theoretic Information Extraction Quality Assessment in the Human Trafficking Domain & NETWORKTHEOHT & \citeA{kejriwal2019network} \\\hline

        & Network-theoretic Modeling of Complex Activity Using UK Online Sex Advertisements & NETWORKTHEOUK & \citeA{kejriwal2020network} \\\hline

        & A Non-parametric Learning Approach to Identify Online Human Trafficking & NONPARAMETRIC & \citeA{alvari2016nonparametric} \\\hline

        & Sex Trafficking Detection with Ordinal Regression Neural Networks & ORDINAL & \citeA{wang2019sex}\\\hline

        & Predictive Patterns of Sex Trafficking Online & PREDICTPATTERNS & \citeA{kennedy2012predictive} \\\hline

        & Semi-supervised Learning for Detecting Human Trafficking & SEMISUP & \citeA{alvari2017semi} \\\hline

        & Identifying Sources of Human Trafficking Within Online Escort Advertisements Written in Spanish & SPANISH & \citeA{rodriguez2021identifying} \\\hline

        & Combating Human Trafficking with Deep Multimodal Models & TRAFFICKING10K & \citeA{tong2017combating} \\\hline

        & TrafficVis Visualizing Organized Activity and Spatio-Temporal Patterns for Detecting and Labeling Human Trafficking & TRAFFICKVIS & \citeA{vajiac2022trafficvis} \\\hline
        
        & Unmasking Human Trafficking Risk in Commercial Sex Supply Chains with Machine Learning & UNMASKING & \citeA{ramchandani2021unmasking} \\\hline 
        
        & Using Knowledge Management to Assist in Identifying Human Sex Trafficking & USINGKMIDENT & \citeA{hultgren2016using} \\\hline
\end{longtable}

\newpage

\begin{longtable}{|p{.18\columnwidth}|p{.36\columnwidth}|p{.36\columnwidth}|}
    \caption{Comprehensive list of Machine Learning methods and related primary studies.}\label{tab:ml_method} \\\hline
    
    \textbf{Algorithm/ Method} & \textbf{Human Trafficking Risk Prediction} & \textbf{Organized Activity Detection} \\\hline
    \endfirsthead
    
    \multicolumn{3}{c}%
    {{\bfseries \tablename\ \thetable{} -- continued from previous page}} \\\hline \textbf{Algorithm/ Method} & \textbf{Human Trafficking Risk Prediction} & \textbf{Organized Activity Detection} \\\hline
    \endhead
    
    \hline \multicolumn{3}{r}{{Continued on next page}} \\ 
    \endfoot

    \hline
    \endlastfoot

        Supervised Learning & ORDINAL, LEVERAGING, LURKING, ENTITYRES, TRAFFICKING10K, FEATURESELECT, IDDETECTLM, SEMISUP, MULTIINPUT, ENSEMBLESENT, CONTEXTLM, COOCCUREMOJI & LEVERAGING, BACKPAGEBITCOIN, ENTITYRES, SPANISH\\\hline
        
        Semi-supervised Learning & FLAGIT, NONPARAMETRIC, SEMISUP & \\\hline
        
        Self-supervised Learning & FLAGIT, ORDINAL, UNMASKING, TRAFFICKING10K, CONTEXTLM & DETECTCHAR, SPANISH\\\hline
        
        Unsupervised Learning &  & LEVERAGING, LURKING, DETECTCHAR, COUPLEDCLUSTER, DETECTIONORG, SPANISH \\\hline
        
        Active Learning & UNMASKING & ACTIVESEARCH \\\hline


\end{longtable}

\begin{longtable}{|p{.18\columnwidth}|p{.36\columnwidth}|p{.36\columnwidth}|}
    \caption{Comprehensive list of Machine Learning featurization and modeling algorithms and related primary studies.}\label{tab:ml_algo} \\\hline
    
    \textbf{Algorithm} & \textbf{Human Trafficking Risk Prediction} & \textbf{Organized Activity Detection} \\\hline
    \endfirsthead
    
    \multicolumn{3}{c}%
    {{\bfseries \tablename\ \thetable{} -- continued from previous page}} \\\hline \textbf{Algorithm} & \textbf{Human Trafficking Risk Prediction} & \textbf{Organized Activity Detection} \\\hline
    \endhead
    
    \hline \multicolumn{3}{r}{{Continued on next page}} \\ 
    \endfoot

    \hline
    \endlastfoot

    \multicolumn{3}{|c|}{\textbf{Featurization}} \\\hline

        doc2vec & FLAGIT & \\\hline
        
        word2vec & ORDINAL, UNMASKING, TRAFFICKING10K & SPANISH \\\hline

        Term Frequency - Inverse Document Frequency & FOODSAFETY, IDDETECTLM, SEMISUP, COOCCUREMOJI & DETECTIONORGANIZED \\\hline
        
        FastText & FLAGIT, CONTEXTLM & \\\hline
        
        Latent Dirichlet Allocation & NONPARAMETRIC, CONTEXTLM & \\\hline
        
        GloVe &  & DETECTCHAR\\\hline
        
        SIF embedding &  & DETECTCHAR\\\hline
        
        Principal Component Analysis &  & LEVERAGING \\\hline

        Locality-Sensitive Hashing & & DETECTCHAR, BUILDINGKG \\\hline

        Singular Value Decomposition & & DETECTIONORG \\\hline
        
        \multicolumn{3}{|c|}{\textbf{Modeling}} \\\hline
        
        Logistic Regression & LURKING, FOODSAFETY, IDDETECTLM, SEMISUP, CONTEXTLM, COOCCUREMOJI & LEVERAGING, BACKPAGEBITCOIN, ENTITYRES\\\hline
        
        Random Forest & LEVERAGING, LURKING, ENTITYRES, FOODSAFETY, SEMISUP, COOCCUREMOJI & ENTITYRES\\\hline

        Classification/ Regression Trees & FOODSAFETY, FEATURESELECT & \\\hline
        
        Naive Bayes & FEATURESELECT, SEMISUP & ENTITYRES \\\hline

        Radial Basis Functions & NONPARAMETRIC, SEMISUP & \\\hline

        K-nearest Neighbors & NONPARAMETRIC, SEMISUP & \\\hline

        Ada Boost & SEMISUP & \\\hline
        
        Support Vector Machines & LURKING, SEMISUP, COOCCUREMOJI & \\\hline
        
        Recurrent Neural Network & ORDINAL, UNMASKING, TRAFFICKING10K, MULTIINPUT & SPANISH\\\hline
        
        Convolutional Neural Network & TRAFFICKING10K, MULTIINPUT & BUILDINGKG\\\hline

        Neural Network & FEATURESELECT, MULTIINPUT & \\\hline
        
        Transformer & UNMASKING, CONTEXTLM & \\\hline
        
        Clustering & & LURKING, DETECTCHAR, DETECTIONORG, SPANISH \\\hline

\end{longtable}

\begin{longtable}{|p{.18\columnwidth}|p{.36\columnwidth}|p{.36\columnwidth}|}
    \caption{Comprehensive list of Machine Learning features and related primary studies.}\label{tab:ml_feat} \\\hline
    
    \textbf{Feature} & \textbf{Human Trafficking Risk Prediction} & \textbf{Organized Activity Detection} \\\hline
    \endfirsthead
    
    \multicolumn{3}{c}%
    {{\bfseries \tablename\ \thetable{} -- continued from previous page}} \\\hline \textbf{Feature} & \textbf{Human Trafficking Risk Prediction} & \textbf{Organized Activity Detection} \\\hline
    \endhead
    
    \hline \multicolumn{3}{r}{{Continued on next page}} \\ 
    \endfoot

    \hline
    \endlastfoot

    Text Embeddings & FLAGIT, CONTEXTLM  &  DETECTCHAR, SPANISH \\\hline
    
    Word Embeddings & ORDINAL, UNMASKING, CONTEXTLM, TRAFFICKING10K  & DETECTCHAR, SPANISH \\\hline
    
    Bag of Words & LEVERAGING, LURKING, FEATURESELECT  & LEVERAGING, BACKPAGEBITCOIN, ENTITYRESOLUTION, FOODSAFETY, IDDETECTLM \\\hline
    
    Physical/ Operational & LEVERAGING, ENTITYRESOLUTION, SEMISUP & LEVERAGING, ENTITYRESOLUTION, ENSEMBLESENT \\\hline
    
    Text Aggregate & FOODSAFETY & BACKPAGEBITCOIN, ENTITYRESOLUTION, IDDETECTLM, SEMISUP, MULTIINPUT, ENSEMBLESENT \\\hline

    TF-IDF & FOODSAFETY, IDDETECTLM, SEMISUP, COOCCUREMOJI & DETECTIONORGANIZED \\\hline

    Perplexity & FOODSAFETY, IDDETECTLM & \\\hline

    LDA & NONPARAMETRIC, CONTEXTLM & \\\hline

    Raw Text & MULTIINPUT &  \\\hline


\end{longtable}

\begin{longtable}{|p{.21\textwidth}|p{.72\textwidth}|}
    \caption{Comprehensive list of human trafficking indicators and related primary studies.}\label{tab:ml_feat} \\\hline
    
    \textbf{Indicator} & \textbf{Primary Studies} \\\hline
    \endfirsthead
    
    \multicolumn{2}{c}%
    {{\bfseries \tablename\ \thetable{} -- continued from previous page}} \\\hline \textbf{Indicator} & \textbf{Primary Studies}
    \\\hline
    \endhead
    
    \hline \multicolumn{2}{r}{{Continued on next page}} \\ 
    \endfoot

    \hline
    \endlastfoot

    Multiple Victims & FLAGIT, IDENTIFYINGHTUK, COVERTSEX, NONPARAMETRIC, DETECTDOMESTIC, SEMISUP, PREDICTPATTERNS, KNOWMANAGE, IDENTBRITISH, USINGKMIDENT \\\hline
    
    Movility of Victims & FLAGIT, IDENTIFYINGHTUK, COVERTSEX, DETECTDOMESTIC, PREDICTPATTERNS, DONTWANT, KNOWMANAGE, CIRCUITSUS, IDENTBRITISH, USINGKMIDENT \\\hline
    
    Risky Activity & FLAGIT, IDENTIFYINGHTUK, IDENTBRITISH, USINGKMIDENT \\\hline
    
    Business & UNMASKING, COVERTSEX, NONPARAMETRIC, SEMISUP, PREDICTPATTERNS, IDENTBRITISH \\\hline

    Underage Victims & DATAINTUNDERAGE, IDENTIFYINGHTUK, NONPARAMETRIC, SEMISUP, PREDICTPATTERNS, DONTWANT, KNOWMANAGE, IDENTBRITISH, USINGKMIDENT \\\hline

    Shared Management & IDENTIFYINGHTUK, COVERTSEX, NONPARAMETRIC, DETECTDOMESTIC, SEMISUP, PREDICTPATTERNS, IDENTBRITISH, USINGKMIDENT \\\hline

    Restricted Movement & IDENTIFYINGHTUK, COVERTSEX, DETECTDOMESTIC, DONTWANT, KNOWMANAGE, IDENTBRITISH, USINGKMIDENT \\\hline

    Ethnicity/ Nationality & COVERTSEX, NONPARAMETRIC, DETECTDOMESTIC, SEMISUP, DONTWANT, KNOWMANAGE, IDENTBRITISH, USINGKMIDENT \\\hline


\end{longtable}

\begin{longtable}{|p{.21\columnwidth}|p{.72\columnwidth}|}
    \caption{Comprehensive list of advertisements connectors and related primary studies.}\label{tab:ml_feat} \\\hline
    
    \textbf{Connector} & \textbf{Primary Studies} \\\hline
    \endfirsthead
    
    \multicolumn{2}{c}%
    {{\bfseries \tablename\ \thetable{} -- continued from previous page}} \\\hline \textbf{Connector} & \textbf{Primary Studies}
    \\\hline
    \endhead
    
    \hline \multicolumn{2}{r}{{Continued on next page}} \\ 
    \endfoot

    \hline
    \endlastfoot

    Phone Number & LEVERAGING, LURKING, BACKPAGEBITCOIN, DETECTCHAR, CRACKINGSEX, UNMASKING, DATAINTMINORS, DATAINTUNDERAGE, IDENTIFYINGHTUK, COVERTSEX, ENTITYRESOLUTION, FOODSAFETY, NETWORKTHEOHT, NETWORKTHEOUK, COUPLEDCLUSTER, DETECTDOMESTIC, BUILDINGKG, ACTIVESEARCH, DETECTIONORGANIZED, TRAFFICKVIS, KGFORSOCIAL, CIRCUITSUS, SPANISH \\\hline
    
    Social Media & LEVERAGING, UNMASKING, TRAFFICKVIS \\\hline
    
    URL & LEVERAGING, UNMASKING, ENTITYRESOLUTION, ACTIVESEARCH \\\hline
    
    Email & LEVERAGING, BACKPAGEBITCOIN, UNMASKING, BUILDINGKG, ACTIVESEARCH, TRAFFICKVIS, KGFORSOCIAL \\\hline
    
    Text & LEVERAGING, LURKING, BACKPAGEBITCOIN, DETECTCHAR, CRACKINGSEX, DATAINTUNDERAGE, IDENTIFYINGHTUK, ENTITYRESOLUTION, BUILDINGKG, ACTIVESEARCH, DELTASHIELD, DETECTIONORGANIZED, TRAFFICKVIS, SPANISH \\\hline
    
    Physical/ Operational & LEVERAGING, DATAINTUNDERAGE, NETWORKTHEOHT, COUPLEDCLUSTER, ACTIVESEARCH, CIRCUITSUS \\\hline
    
    Image & LURKING, CRACKINGSEX, DATAINTMINORS, DATAINTUNDERAGE, COVERTSEX, ENTITYRESOLUTION, BUILDINGKG, ACTIVESEARCH, TRAFFICKVIS \\\hline
    
    Name/ Alias & DATAINTMINORS, DATAINTUNDERAGE, COVERTSEX, NETWORKTHEOHT, ACTIVESEARCH \\\hline


\end{longtable}


\newpage
\raggedright
\bibliographystyle{apacite} 
\urlstyle{same}
\interlinepenalty=10000
\bibliography{main}

\end{document}